\documentclass[lettersize,journal]{IEEEtran}

\usepackage{amsmath,amsfonts,amssymb}
\usepackage{algorithmic}
\usepackage[ruled,linesnumbered]{algorithm2e}
\usepackage{array}
\usepackage[caption=false,font=sf,labelfont=sf,textfont=sf]{subfig}
\usepackage{textcomp}
\usepackage{stfloats}
\usepackage{url}
\usepackage{verbatim}
\usepackage{graphicx}
\usepackage{balance}

\usepackage{amsthm}

\usepackage{cite}
\usepackage{booktabs}
\usepackage{bm}
\usepackage{makecell}
\usepackage{tabularx}
\usepackage{multirow}
\usepackage{xcolor}
\usepackage{listings}
\usepackage{pifont}
\usepackage{arydshln}
\usepackage{xspace}

\usepackage[pagebackref,breaklinks,colorlinks,allcolors=blue]{hyperref}

\usepackage[capitalize]{cleveref}
\crefname{paragraph}{Section}{Sections}
\Crefname{paragraph}{Section}{Sections}

\def\BibTeX{{\rm B\kern-.05em{\sc i\kern-.025em b}\kern-.08em
    T\kern-.1667em\lower.7ex\hbox{E}\kern-.125emX}}

\newcommand{\cmark}{\ding{51}}
\newcommand{\xmark}{\ding{55}}

\newcommand{\tocite}[1]{\textcolor{blue}{[TO CITE]}}
\newcommand{\ourmethod}{{DiverseDiT++}\xspace}
\newcommand{\etal}{\textit{et al.}}

\newcommand{\myparagraph}[1]{\subsubsection{#1}}

\DeclareRobustCommand*{\IEEEauthorrefmark}[1]{\raisebox{0pt}[0pt][0pt]{\textsuperscript{\normalfont\footnotesize #1}}}

\begin{document}

\title{DiverseDiT++: Quantifying, Analyzing, and Promoting Representation Diversity in Diffusion Transformers}

\author{
    \IEEEauthorblockN{Binglei~Li\IEEEauthorrefmark{1}\textsuperscript{,}\IEEEauthorrefmark{2}} \and
    \IEEEauthorblockN{Mengping~Yang\IEEEauthorrefmark{1}\textsuperscript{,}\IEEEauthorrefmark{3}\textsuperscript{$\dagger$}} \and
    \IEEEauthorblockN{Zhiyu~Tan\IEEEauthorrefmark{1}\textsuperscript{,}\IEEEauthorrefmark{3}} \and
    \IEEEauthorblockN{Xiaomeng~Yang\IEEEauthorrefmark{3}} \and
    \IEEEauthorblockN{Zhizhong~Huang\IEEEauthorrefmark{1}} \\
    \IEEEauthorblockN{Junping~Zhang\IEEEauthorrefmark{1}\textsuperscript{$\ddagger$}} \and
    \IEEEauthorblockN{Hao~Li\IEEEauthorrefmark{1}\textsuperscript{,}\IEEEauthorrefmark{2}\textsuperscript{,}\IEEEauthorrefmark{3}\textsuperscript{$\ddagger$}} \\
    \IEEEauthorblockA{\IEEEauthorrefmark{1} Fudan University, Shanghai, China} \\
  \IEEEauthorblockA{\IEEEauthorrefmark{2} Shanghai Innovation Institute, Shanghai, China} \\
  \IEEEauthorblockA{\IEEEauthorrefmark{3} Shanghai Academy of AI for Science, Shanghai, China} \\
  \IEEEauthorblockA{\textsuperscript{†} Project Lead \qquad \textsuperscript{$\ddagger$} Corresponding Authors}
}

\maketitle

\begin{abstract}

Recent advances in Diffusion Transformers (DiTs) have enabled remarkable progress in visual synthesis, benefiting from their superior scalability.
To facilitate DiTs' capability of capturing meaningful internal representations, recent works such as REPA incorporate external pretrained encoders for representation alignment.
However, the underlying mechanisms governing representation learning within DiTs remain poorly understood in the community.
To this end, this paper first presents a systematic analysis of the representation dynamics of DiTs via quantifying the diversity of block-wise representations.
Specifically, we introduce a novel metric, termed the Weighted Diversity Score (WDS), to measure the representational discrepancies across different blocks.
Through extensive investigations on the evolution and influence of internal representations under various settings, we reveal that \textit{representation diversity across blocks} is a critical factor for effective representation learning in DiTs.
More importantly, WDS exhibits a strong correlation with synthesis quality across diverse settings, model scales, and training stages  (Pearson's $r=-0.869$ with $\log(\text{FID})$), suggesting its potential as an indicator to reflect model performance and a principled guide for model optimization.
Based on this key finding, we propose \ourmethod, a novel framework that explicitly promotes diverse representation learning.
Concretely, our method incorporates long residual connections to diversify input representations across blocks and a representation diversity loss to encourage blocks to learn distinct features.
Extensive experiments on ImageNet $256\times256$ and $512\times512$ demonstrate that our \ourmethod yields consistent performance gains and convergence acceleration when applied to different backbones with various sizes, even when tested on the challenging one-step generation setting.
Beyond image generation, \ourmethod also generalizes to protein inverse folding and 3D molecule generation, further validating representation diversity as a general principle for effective diffusion model training.
Overall, our work provides valuable insights into the representation learning dynamics of DiTs
and offers a practical framework for enhancing their representation learning.

\end{abstract}

\begin{IEEEkeywords}
Diffusion Transformers, Representation Learning, Representation Diversity, Image Generation, Protein Inverse Folding, Molecule Generation.
\end{IEEEkeywords}

\section{Introduction}
\label{sec:intro}

\begin{figure*}[!t]
    \centering
    \includegraphics[width=\linewidth]{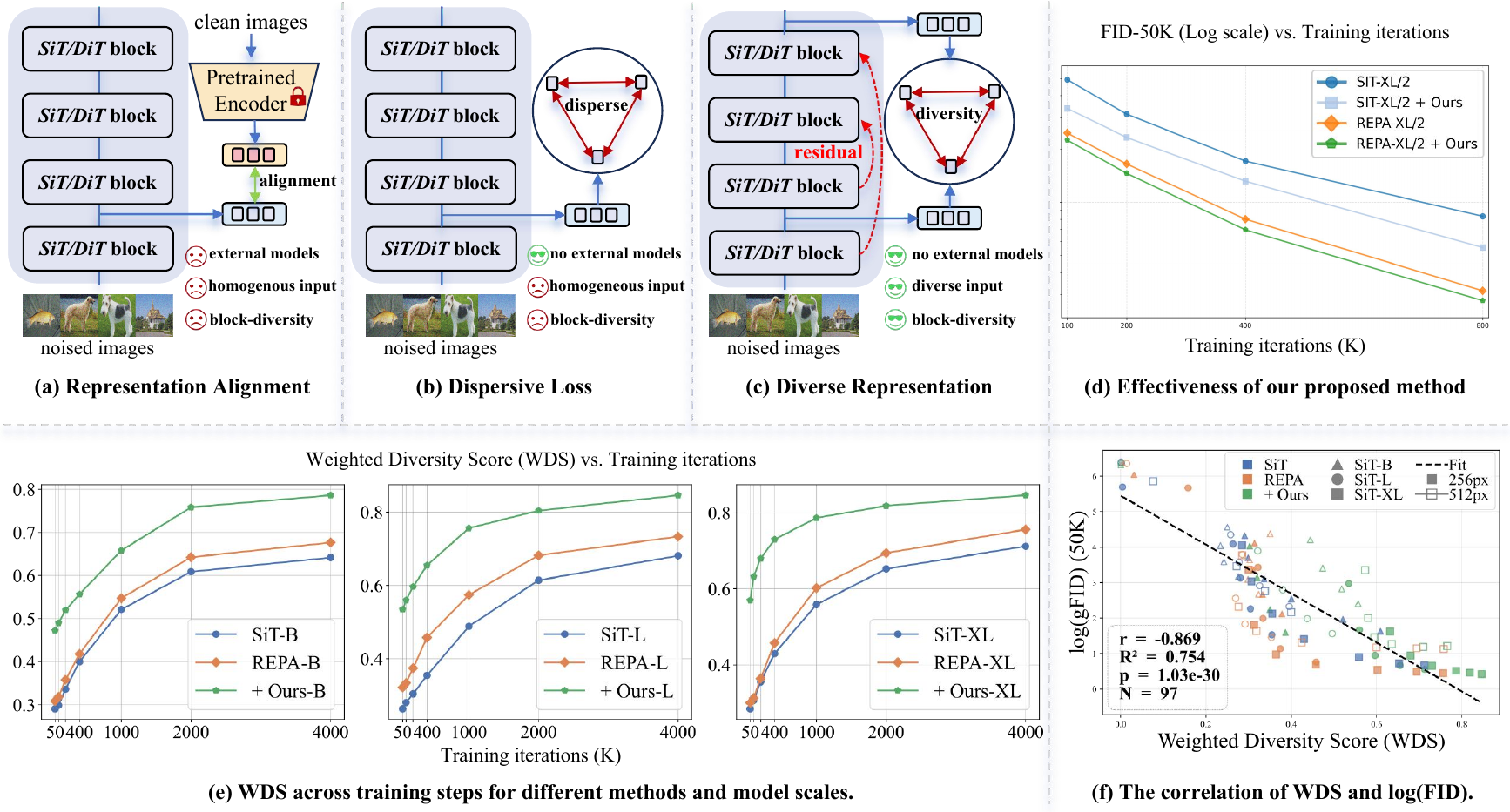}
    \caption{
    \textbf{Comparison between Representation Alignment~\cite{repa}, Dispersive Loss~\cite{wang2025diffuse} and our DiverseDiT++ in learning representations, and quantitative validation of the Weighted Diversity Score (WDS)}.
    (a) REPA~\cite{repa} employs external encoders as guidance and different blocks' inputs are homogeneous.
    (b) Disp~\cite{wang2025diffuse} encourages internal representations to spread out but still with homogeneous input and without block-wise diversity.
    (c) We propose long residual connections to enhance input diversity and diversity loss to encourage diverse feature representations across blocks.
    (d) On ImageNet $256\times256$, our method consistently improves training efficiency and effectiveness when applied to both SiT and REPA.
    (e) WDS consistently increases during training for all methods and model scales. \ourmethod achieves the highest WDS, indicating the greatest representational diversity among blocks.
    (f) WDS exhibits a strong linear correlation with $\log(\text{FID})$ ($r = -0.869$), confirming that feature diversity reliably predicts generation quality across all settings.
    }
     \label{fig:teaser}
\end{figure*}

\IEEEPARstart{D}{iffusion} models~\cite{sohl2015deep}~\cite{ddpm}, particularly diffusion transformers (DiT)~\cite{dit}~\cite{bao2023all}, have significantly advanced visual synthesis due to their superior scalability in modeling complex data distributions.
These advances have enabled thrilling progress across a wide range of generative tasks, including text-to-image~\cite{xie2025sana}~\cite{rombach2022high}, text-to-video generation~\cite{yang2024cogvideox}~\cite{gupta2024photorealistic}, \emph{etc}.
Recent studies further suggest that top-performing diffusion models capture more discriminative internal representations~\cite{mittal2023diffusion}~\cite{chen2024deconstructing}~\cite{xiang2023denoising}, yielding an implicit connection between diffusion models and representation learning.

Following this perspective, recent methods have attempted to improve DiT training by explicitly guiding their internal representations.
For instance, REPA~\cite{repa} (\cref{fig:teaser} (a)) aligns latent noisy representations with pre-trained visual encoder features to facilitate representation learning.
Subsequent work REPA-E~\cite{leng2025repa} extends such alignment via joint end-to-end training with VAE tuning and REG~\cite{wu2025representation} entangles low-level visual latents and high-level class tokens for two-level alignment.
However, these methods rely on powerful external foundation models, whose training requires massive resources for training.
Other approaches seek to enhance representations without external guidance.
SRA~\cite{jiang2025no} aligns representations between a student and an EMA teacher model, while Wang \etal~\cite{wang2025diffuse} propose a dispersive loss to separate internal representations (\cref{fig:teaser} (b)).
Despite their considerable advancements, the underlying mechanisms governing representation learning in DiTs remain insufficiently understood.
Key questions persist:
How do DiT models learn meaningful representations, and why are external alignment techniques effective?
This lack of fundamental understanding hinders the development of more principled and efficient training paradigms.

To address this gap, we first conduct a systematic analysis of the representation dynamics in DiTs by quantifying and visualizing block-wise representation diversity (\cref{sec:revisting_repa}).
Specifically, we develop \emph{Weighted Diversity Score} (WDS) as a quantitative metric and plot the Centered Kernel Alignment (CKA) similarities~\cite{kornblith2019similarity} to measure representational discrepancies across different DiT blocks.
With these tools, we investigate both the evolution and the influence of internal representations under various settings, including different training stages, alignment strategies, denoising timesteps, model scales, and encoder choices.
Through these analyses, we reveal several key findings:
1) Representation discrepancy across blocks naturally increases during training;
2) Aligning a single block with a pre-trained model significantly increases its discrepancy from other blocks;
3) Aligning more blocks or using multiple external encoders does not necessarily improve performance, suggesting that excessive alignment might harm overall representational diversity.
4) The diversity patterns remain consistent across denoising timesteps, confirming that representation diversity is an intrinsic structural property rather than an artifact of specific noise steps.
Moreover, WDS exhibits a strong correlation with synthesis quality across diverse settings, model scales, and training stages (Pearson's $r=-0.869$ with FID), demonstrating its potential utility as a indicator of model performance and a principled guide for model optimization.
These observations provide a new perspective on DiT representation learning and offer a plausible explanation for the effectiveness of existing techniques like REPA: the rationale behind effective representation learning in DiTs lies in \textit{improving the representation diversity} across different blocks.

Capitalizing on the above insights, we propose \ourmethod, a novel and effective framework for promoting diverse representation learning in DiTs.
Specifically, \ourmethod introduces two simple yet powerful components.
First, we incorporate long residual connections to diversify block inputs, preventing representational homogenization.
Second, we introduce a representation diversity loss that explicitly penalizes similarity between features from different blocks. This encourages each block to specialize and capture unique, complementary aspects of the data.
Furthermore, we design an adaptive weighting strategy for the diversity loss via gradient conflict analysis, suppressing the loss when diversity saturates and decaying it over training.
Together, these components promote diverse representation learning through both diverse inputs and inter-block diversity constraints, without requiring external guidance models (\cref{fig:teaser} (c)).

We conduct extensive experiments across diverse generative tasks and settings to comprehensively identify the effectiveness of \ourmethod.
Regarding image generation, we evaluate \ourmethod on ImageNet 256$\times$256 and 512$\times$512 across different backbones, model scales and training settings.
The results demonstrate consistent improvements in the training convergence and synthesis quality across baselines with or without external guidance and across model scales (\cref{fig:teaser} (d)), including the challenging one-step setting~\cite{geng2025mean}.
Moreover, \ourmethod is complementary to existing alternatives such as Disp~\cite{wang2025diffuse} and SRA~\cite{jiang2025no}, yielding additional performance improvement when integrated with them.
In order to further testify the generalization capability of our method, we extend \ourmethod to scientific generation tasks, \emph{i.e.,} protein inverse folding and 3D molecule generation tasks.
Our method yields consistent performance gains on these tasks, indicating that promoting representation diversity is a general principle for enhancing diffusion model training across data modalities and application domains.

We summarize our primary contributions as follows:
1) We systematically investigate the representation learning dynamics of DiTs and reveal that \textit{representation diversity across blocks} naturally emerges during training and serves as a key factor for effective learning. To our knowledge, this is the first work to elucidate this relationship and provide mechanistic insight into why existing alignment techniques are effective.
2) We introduce the Weighted Diversity Score (WDS), a metric to quantitatively capture inter-block diversity and strongly correlates with FID across model scales and methods (\cref{fig:teaser}(f)), providing a principled metric for understanding and diagnosing DiT training.
3) We propose \ourmethod, a practical framework that explicitly promotes diverse representation learning through long residual connections and a representation diversity loss with adaptive weighting. These components diversify block-wise inputs and encourage different blocks to learn distinct and complementary representations without relying on external guidance models.
4) Extensive experiments across visual and scientific generation tasks demonstrate the effectiveness and generality of \ourmethod. On image generation, \ourmethod consistently accelerates convergence and improves synthesis quality across scales and baselines, including one-step generation, while being complementary to REPA, Disp, and SRA. Additionally, \ourmethod further generalizes to protein inverse folding and 3D molecule generation, demonstrating the broad applicability of representation diversity across domains.

We note that this paper substantially extends the preliminary version published in CVPR 2026~\cite{diversedit_yang2026} with the following major improvements.
First, we deepen the diversity analysis with a new finding of diversity pattern towards denoising timesteps, supported by additional quantitative results and visualizations.
Second, we propose the Weighted Diversity Score, a principled metric that quantifies inter-block representational diversity as a distance-weighted complement of pairwise CKA similarities.
Third, we validate WDS through a large-scale correlation study spanning $97$ experiment settings across model scales, methods, and resolutions, revealing a strong relationship with $\log(\text{FID})$ and confirming WDS as a reliable diagnostic for DiT training.
Fourth, we enrich the methodology with three formal theoretical motivations for the diversity loss, and a more thorough analysis of the adaptive weighting mechanism.
Fifth, we extend \ourmethod to scientific generation tasks, including protein folding and molecule generation, showing that promoting representation diversity is a universal principle across data modalities and model architectures.
Finally, we provide more comprehensive experimental evaluation, including long-schedule training, additional checkpoints, WDS values for other SiT/DiT-based methods, and extensive new ablation studies.
In sum, \ourmethod contributes to a deeper understanding of representation learning in DiTs and offers a practical strategy for boosting their performance.

\section{Related Work}
\label{sec:related}

\noindent \textbf{Deep Representation Learning.}
Central to representation learning~\cite{6472238} is to learn rich and meaningful representations for downstream tasks.
This field has evolved through several key paradigms, including discriminative, generative, and multimodal approaches~\cite{10579040}~\cite{10529603}~\cite{9086055}.
Discriminative methods, exemplified by contrastive learning-based methods including BYOL~\cite{grill2020bootstrap}, DINO~\cite{oquab2023dinov2}, and MoCo~ \cite{chen2021empirical}, capture discriminative signals between images to learn strong representations.
The generative variant learns the underlying data distribution via reconstructing input images, representative works including auto-encoder methods VAE~\cite{kingma2013auto}, MAE~\cite{he2022masked}, and masked image modeling~\cite{xie2022simmim}.
Similarly, diffusion models also learn informative features as inherent denoising autoencoders~\cite{mittal2023diffusion}~\cite{yang2023diffusion}~\cite{zhang2022unsupervised}.
To enable cross-model understanding and retrieval, multimodal methods~\cite{CLIP_radford2021learning}~\cite{li2022blip}~\cite{zhai2023sigmoid}~\cite{tschannen2025siglip} align textual and visual signals in a shared representation space.
The quality of learned representations is commonly evaluated via linear probing accuracy on ImageNet~\cite{imgnet}, and retrieval metrics~\cite{sajjadi2018assessing}~\cite{precrecall}.
Moreover, CKA~\cite{kornblith2019similarity}~\cite{davari2022reliability} is widely used for quantifying neural network representations.
For generative models, FID~\cite{fid} and its variant~\cite{fd_dinov2_stein2023}~\cite{sfid} are widely adopted to jointly evaluate generation fidelity and the semantic richness of the underlying features.
Wang \etal~\cite{wang2025diffuse} utilized diversity metrics
to assess whether a model learns diverse representations.
Despite these advancements, it remains unclear what representations should be learned for diffusion generative models, and principled metrics for quantitatively evaluating the diversity of learned features in such models are still lacking.

\noindent \textbf{Improving Diffusion Models with Representation Learning.}
Diffusion probabilistic models~\cite{sohl2015deep}~\cite{ddpm}~\cite{ddim_song2020}~\cite{diffusion_survey_10081412}, which generate images via iteratively denoising Gaussian noises, have become the dominating paradigm for image~\cite{rombach2022high}~\cite{chen2023pixart} and video generation~\cite{yang2024cogvideox}~\cite{hunyuan_video_2024}, driven by improved training stability with flow matching~\cite{flowmatching_lipman2022}~\cite{RectifiedFlow_liuflow} and exceptional model scalability from conventional UNet-based models~\cite{rombach2022high}~\cite{dhariwal2021diffusion}~\cite{blattmann2023stable} to the Transformer-based architectures~\cite{dit}~\cite{bao2023all}~\cite{sit}.
Improved representation learning in diffusion models advances both synthesis quality and downstream tasks~\cite{li2023dreamteacher}~\cite{xiang2023denoising}~\cite{11367476}.
REPA~\cite{repa} aligns diffusion representations with pretrained encoders, extended by REPA-E~\cite{leng2025repa} for end-to-end VAE training and SARA~\cite{chen2025sara} for structural alignment.
SoftREPA~\cite{softrepa} extends alignment to text embeddings, REG~\cite{wu2025representation} entangles image latents with class tokens for discriminative representations, and REED~\cite{reed} learns cross-modal representations via time-weighted optimization.
Without external guidance, SRA~\cite{jiang2025no} uses lower-noise later-layer representations to supervise higher-noise earlier layers, and Wang \etal~\cite{wang2025diffuse} regularize features via DispLoss to encourage informative representations.
While concurrent work iREPA~\cite{irepa_singh2025matters} pointed out that structure is the key for improving representation learning in DiTs, it remains unclear how meaningful representations are learned and what representations within models are more suitable.
We thus perform a systematic investigation on this and develop \ourmethod to learn diverse and effective representations, providing a brand new perspective for delivering stronger diffusion models.

\noindent \textbf{Diffusion Models for Scientific Domains.}
Beyond visual synthesis, diffusion-based generative models have shown strong performance across diverse scientific domains.
For protein inverse folding, discrete diffusion and flow models~\cite{multiflow_campbell2024generative}~\cite{proteinmpnn} generate protein sequences conditioned on backbone 3D structures, with ProteinMPNN~\cite{proteinmpnn} as the de facto baseline.
For molecular generation, $E(3)$-equivariant approaches model 3D molecular structures, including MiDi~\cite{midi}, EQGAT-diff~\cite{eqgatdiff}, and SemlaFlow~\cite{irwin2025semlaflow}.
Recent works~\cite{reed} show that pretrained representations from domain-specific foundation models (\emph{e.g.,} AlphaFold3~\cite{af3} for proteins, Unimol~\cite{zhou2023unimol} for molecules) can significantly accelerate training and improve generation quality.
However, most existing approaches obtain improved representations through external pretrained models.
Our framework instead enhances the representations learned within the diffusion model itself, without requiring additional encoders, making it complementary to these methods.

\section{How Representations Are Learned?}
\label{sec:revisting_repa}

\begin{figure*}[!t]
    \centering
    \includegraphics[width=\linewidth]{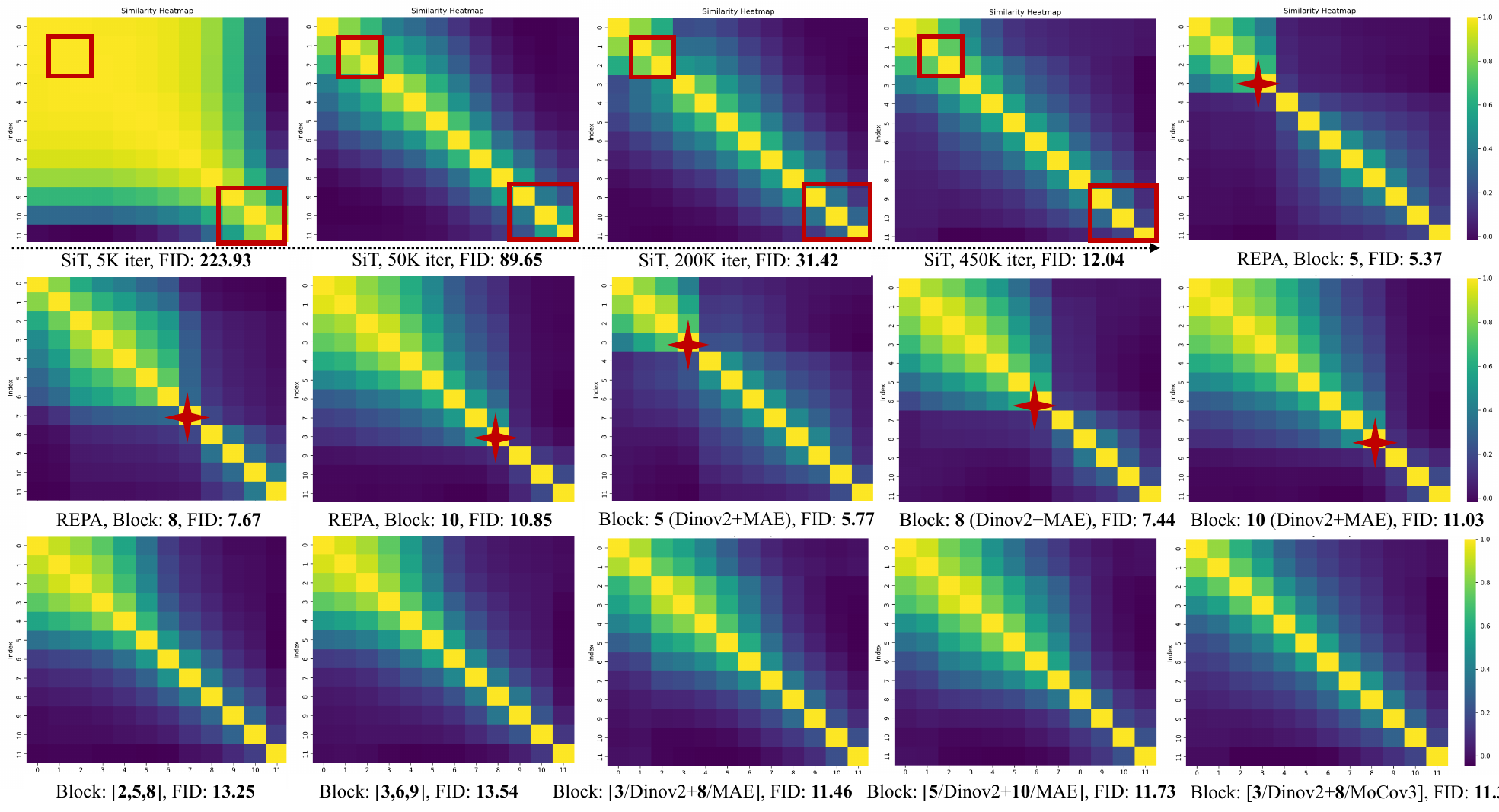}
    \caption{\textbf{CKA representation similarities of models trained on various settings.}
    We can observe that 1) the discrepancies between different blocks increases as training progresses;
    2) aligning specific blocks significantly increases the dissimilarity between the corresponding block and other blocks;
    3) aligning on more blocks with different pretrained encoders brings marginal performance improvements.
    Detailed quantitative results are provided in \cref{tab:analysis_results}.
    }
    \label{fig:analysis}
\end{figure*}

\begin{table}[t]
    \centering
    \scriptsize
    \vskip -0.1in
    \caption{\textbf{Detailed quantitative results of our systematic analysis}.
    All implementation details strictly follow the default settings of SiT and REPA for our investigation.
    All baselines are reported using vanilla-REPA \cite{repa} for training.
    Encoder abbreviations: \textit{\textbf{D}}=DINOv2-B, \textit{\textbf{M}}=MAE, \textit{\textbf{MC}}=MoCoV3.
    }
    \vskip -0.1in
    \setlength{\tabcolsep}{0.6mm}{
    \begin{tabular}{lccccccccc}
        \toprule
        \textbf{Model} & \textbf{Align?}  & \textbf{Encoder}. & \textbf{Align Depth}. &\textbf{Iter}. & \textbf{FID}$_\downarrow$ & \textbf{sFID}$_\downarrow$ & \textbf{IS}$_\uparrow$ & \textbf{Prec.}$_\uparrow$ & \textbf{Rec.}$_\uparrow$ \\
        \midrule
        \multirow{5}{*}{\textbf{SiT-B}}
           & \xmark & \xmark  & \xmark & 50k  & 89.65 & 12.58 & 18.28 & 0.34 & 0.43 \\
           & \xmark & \xmark  & \xmark & 100k & 40.46 & 6.15 & 36.26 & 0.52 & 0.49 \\
           & \xmark & \xmark  & \xmark & 200k & 31.42 & 5.87 & 58.08 & 0.61 & 0.53 \\
           & \xmark & \xmark  & \xmark & 400k & 12.69 & 5.29 & 106.21 & 0.71 & 0.54 \\
           & \xmark & \xmark  & \xmark & 450k & 12.04 & 5.24 & 110.19 & 0.71 & 0.54 \\
        \midrule
        \multirow{15}{*}{\textbf{REPA-B}}
           & \cmark & \textit{\textbf{D}}       & 5       & 450k & 5.37    & 5.35  & 175.07   & 0.75    & 0.58 \\
           & \cmark & \textit{\textbf{D}}       & 8       & 450k & 7.67    & 5.60  & 150.87   & 0.72    & 0.58 \\
           & \cmark & \textit{\textbf{D}}       & 10      & 450k & 10.85   & 6.12  & 128.34   & 0.70    & 0.58 \\
           & \cmark & \textit{\textbf{D}}       & \{2,5,8\} & 450k & 13.25   & 5.27  & 105.64   & 0.70    & 0.55 \\
           & \cmark & \textit{\textbf{D}}       & \{3,6,9\} & 450k & 13.54   & 5.50  & 104.88   & 0.69    & 0.56 \\
           & \cmark & \textit{\textbf{M}}            & 5       & 450k & 10.15   & 5.11  & 123.24   & 0.72    & 0.55 \\
           & \cmark & \textit{\textbf{M}}            & 8       & 450k & 11.40   & 5.22  & 115.20   & 0.72    & 0.55 \\
           & \cmark & \textit{\textbf{M}}            & 10      & 450k & 12.11   & 5.30  & 111.01   & 0.71    & 0.55 \\
           & \cmark & \textit{\textbf{D}}+\textit{\textbf{M}}    & 5       & 450k & 5.77    & 5.10  & 166.15   & 0.76    & 0.57 \\
           & \cmark & \textit{\textbf{D}}+\textit{\textbf{M}}    & 8       & 450k & 7.44    & 5.37  & 150.12   & 0.73    & 0.57 \\
           & \cmark & \textit{\textbf{D}}+\textit{\textbf{M}}    & 10      & 450k & 11.03   & 6.17  & 124.89   & 0.70    & 0.55 \\
           & \cmark & \textit{\textbf{D}}+\textit{\textbf{M}}    & \{3,8\}   & 450k & 11.46   & 5.20  & 113.95   & 0.72    & 0.55 \\
           & \cmark & \textit{\textbf{D}}+\textit{\textbf{M}}    & \{5,10\}  & 450k & 11.73   & 5.22  & 111.77   & 0.71    & 0.54 \\
           & \cmark & \textit{\textbf{D}}+\textit{\textbf{MC}} & \{3,8\}   & 450k & 11.36   & 5.24  & 115.75   & 0.71    & 0.55 \\
           & \cmark & \textit{\textbf{D}}+\textit{\textbf{MC}} & \{5,10\}  & 450k & 12.71   & 6.08  & 104.46   & 0.66    & 0.56 \\
        \bottomrule
    \end{tabular}
    }
    \label{tab:analysis_results}
\end{table}

\subsection{Preliminaries}
\label{sec:preliminaries}

\noindent \textbf{Scalable Interpolant Transformers (SiT).}
Our work is based on SiT, which unifies flow~\cite{flowmatching_lipman2022} and diffusion~\cite{ddpm} models that transform Gaussian noise $\epsilon$ into samples $\mathbf{x}_*$:
\begin{equation}
    \mathbf{x}_t = \alpha_t \mathbf{x}_* + \sigma_t \epsilon,
\end{equation}
where $\alpha_t$ decreases and $\sigma_t$ increases with time $t$.
Flow-based models interpolate between noise and data while diffusion models define a stochastic differential equation to approach Gaussian distribution as $t\rightarrow \infty$.
Sampling is performed via a reverse SDE for diffusion or a probability flow ODE for flow models: $\dot{\mathbf{x}}_t = \mathbf{v}(\mathbf{x}_t, t)$, the velocity field $\mathbf{v}(\mathbf{x}_t, t)$ can be formulated with conditional expectation:
\begin{equation}
    \mathbf{v}\!(\mathbf{x}\!,\!t)\! \!=\! \mathbb{E}[\dot{\mathbf{x}}_t \!\mid\! \!\mathbf{x}_t\! \!=\! \mathbf{x}] \!\!=\!\! \dot{\alpha}_t \mathbb{E}[\mathbf{x}_* \!\mid\! \!\mathbf{x}_t\! \!=\! \!\mathbf{x}] \!\!+\! \dot{\sigma}_t \mathbb{E}[\epsilon \!\!\mid\!\! \mathbf{x}_t\! \!=\! \!\mathbf{x}].
\end{equation}
The velocity for the velocity field $\mathbf{v}(\mathbf{x}_t, t)$ is derived by a model $\mathbf{v_\theta}(\mathbf{x}_t, t)$ trained to minimize:
\begin{equation}
    \mathbb{E}_{\mathbf{x}_*, \epsilon, t} \left[ \| \mathbf{v}_\theta(\mathbf{x}_t, t) - \dot{\alpha}_t \mathbf{x}_* - \dot{\sigma}_t \epsilon \|^2 \right].
\end{equation}
Once trained, we can synthesize samples from random noises by the reverse SDE via computing the velocity field:
\begin{equation}
d\mathbf{x}_t \!=\! \mathbf{v}(\mathbf{x}_t, t) dt \!-\! \frac{1}{2} w_t \mathbf{s}(\mathbf{x}_t, t) dt + \sqrt{w_t} d\overline{\mathbf{w}}_t,
\end{equation}
where score $\mathbf{s}(\mathbf{x}_t, t)$ is obtained via conditional expectation:
\begin{equation}
\mathbf{s}(\mathbf{x}_t, t) \!=\! -{\sigma}_t^{-1} \mathbb{E}[\epsilon \!\mid\! \mathbf{x}_t \!=\! \mathbf{x}] \!=\! \sigma_t^{-1} \frac{\alpha_t \mathbf{v}(\mathbf{x}, t) - \dot{\alpha}_t \mathbf{x}}{\alpha_t \dot{\sigma}_t - \dot{\alpha}_t \sigma_t}.
\end{equation}

\noindent \textbf{Representation Alignment (REPA).}
To leverage external models to aid representation learning for DiTs, REPA proposes to perform patch-wise projection alignment between the model's intermediate hidden states $h$ with features $\mathbf{y}_*$ derived from pretrained visual encoders:
\begin{equation}
\mathcal{L}_{\text{REPA}}(\!\theta,\! \phi)\! \!:=\! -\mathbb{E}_{\mathbf{x}_*, \epsilon, t} \!\left[\! \frac{1}{N} \sum_{n=1}^{N} \text{sim}(\mathbf{y}_*^{[n]}, h_\phi(\mathbf{h}_t^{[n]})) \!\right],
\end{equation}
where $\mathbf{x}_*$ denotes clean images, $h_\phi$ is MLP projectors and $\text{sim}(\cdot,\cdot)$ denotes similarity function.

\noindent \textbf{Centered Kernel Alignment (CKA).}
CKA is a widely used similarity index for quantifying neural network representations~\cite{kornblith2019similarity}~\cite{davari2022reliability}.
Accordingly, we adopt CKA to calculate the similarities of representations across DiT blocks for our analysis.
Formally, CKA is normalized from Hilbert-Schmidt Independence Criterion (HSIC)~\cite{HSIC} to be invariant to orthogonal
transformation and isotropic scaling:
\begin{align}
    \label{eq:cka}
    \mathrm{CKA(X,Y)}=\frac{\mathrm{HSIC}(\mathrm{x},\mathrm{y})}{\sqrt{\mathrm{HSIC}(\mathrm{x},\mathrm{x}) \mathrm{HSIC}(\mathrm{y},\mathrm{y})}}.
\end{align}
HSIC identifies whether two distributions ($\mathrm{X,Y}$) are independent: $\mathrm{HSIC}(K,L) \!=\! \frac{1}{(n-1)^2}\operatorname{Tr}(K H L H)$, $K_{i j}\!=\!k\left(\mathrm{x}_i, \mathrm{x}_j\right)$ { and } $L_{i j}\!=\!l\left(\mathrm{y}_i, \mathrm{y}_j\right)$, where $k$ and $l$ are kernels.

\subsection{Key Observations}
\label{sec:observations}
With CKA as the representation similarity index, we systematically investigate how representations are learned and how they respond to external alignment in three settings.
1) SiT training stage analysis: we track the evolution of internal representations and quantify the change in similarity between different blocks as the model learns.
2) REPA block-specific alignment: we identify the effect of aligning pretrained visual features on different blocks to assess how external knowledge alters the representation of specific blocks.
3) REPA multiple block guidance from multiple encoders: we explore the impact of applying external guidance to multiple blocks with multiple encoders to probe whether guiding multiple blocks leads to improvement.
All implementation details strictly follow the settings of SiT-B/2 and REPA-B/2 on Imagenet $256\times256$ for $450K$ iterations.
We use DINOv2-B~\cite{oquab2023dinov2}, MAE-L~\cite{mae} and MoCov3~\cite{chen2021empirical} for REPA alignment.
The visualized results are shown in \cref{fig:analysis} and detailed quantitative results are given in \cref{tab:analysis_results}, from which we can observe several findings.

\noindent \textbf{(1). Representation diversity across different blocks increases during training:}
The similarity heatmaps become progressively more diagonal as training progresses (5K$\rightarrow$450K), indicating increasing inter-block dissimilarity.
Intuitively, different blocks specialize, develop more distinct and complementary representations.
Such observation aligns with the broader understanding that deep models learn hierarchical representations.

\noindent \textbf{(2). External alignment enhances block differentiation:}
The REPA heatmaps exhibit more distinct patterns around the red mark, indicating that aligning specific blocks significantly increases the dissimilarity between the representations of the targeted block and other blocks.
Additionally, consistent with REPA~\cite{repa}, aligning earlier blocks (\emph{i.e.,} Block $5$, Block $8$) yields better performance than aligning later blocks (Block $10$).
In other words, REPA encourages each block to learn more distinct and complementary features, leading to a more diverse and more effective representation.
More importantly, these observations explain REPA's effectiveness: by enforcing specialization, it prevents representational collapse and encourages the network to explore a wider range of features.
This specialization-driven perspective may also explain why aligning with larger models (DINOv2-L, -g) brings only marginal improvements over DIVOv2-B in original REPA.

\noindent \textbf{(3). Aligning on more blocks with more external models does not necessarily improve performance:}
Multi-block guidance (e.g., Block:$\{2,5,8\}$, $\{3,6,9\}$) fails to replicate the improvements seen with single-block alignment.
In some cases, the FID score is even slightly worse (Block $\{2,5,8\}$), suggesting more blocks counterintuitively reduce overall diversity.
We attribute this to the conflicting constraints that prevent individual blocks from effectively specializing.
Furthermore, aligning multiple blocks with different external encoders (\{$5$/Dinov2+$10$/MAE\}, \emph{i.e.}, aligning DinoV2 features on block $5$ and MAE features on block $10$) also provides limited benefit and shows limited representation diversity.
Such observation further reflects that the representation diversity across blocks is a crucial factor for high-quality synthesis.

\begin{figure}[t]
    \centering
    \includegraphics[width=0.9\linewidth]{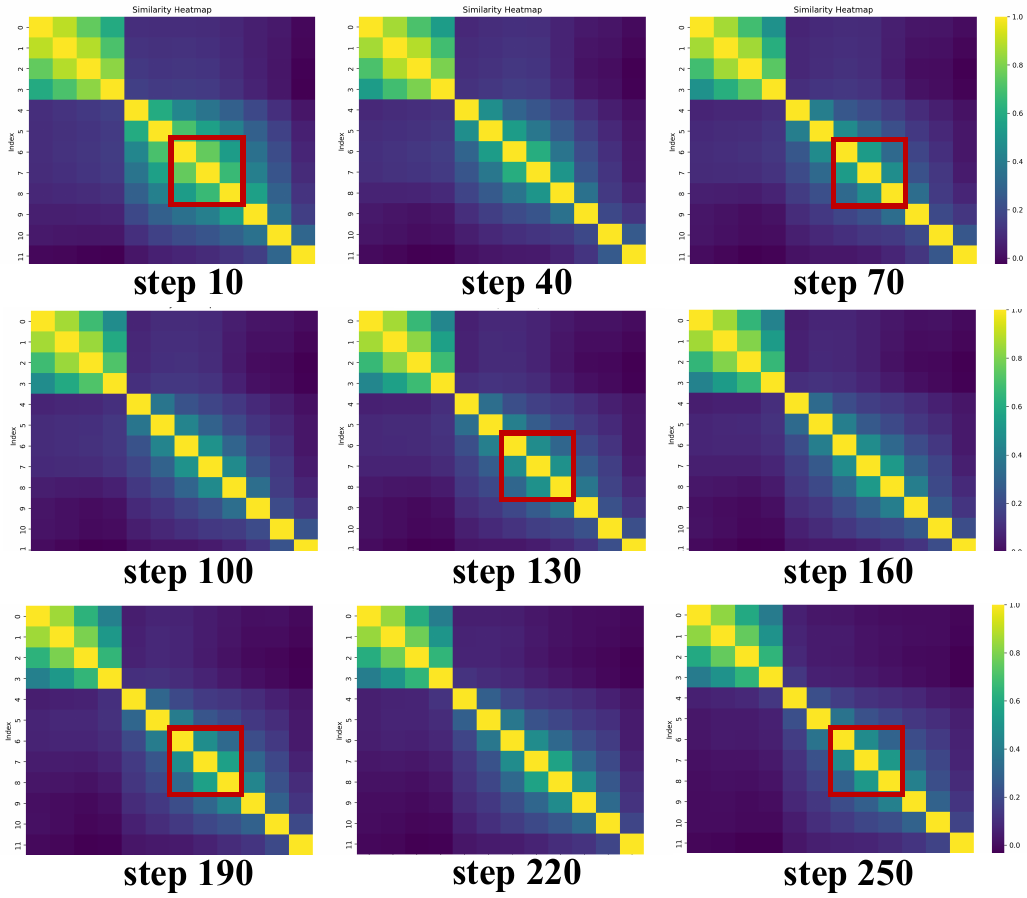}
    \vskip -0.1in
    \caption{\textbf{CKA representation similarities across different timesteps.}
    The representational discrepancies across different timesteps show similar correlation patterns, indicating that block diversity originates from internal representations rather than denoising timesteps.
    }
    \label{fig:cka_different_time}
\end{figure}

\noindent \textbf{(4). Representation diversity originates from internal representations rather than denoising timesteps:}
To further investigate the source of representational discrepancies, we analyze the CKA similarities across different denoising timesteps in~\cref{fig:cka_different_time}.
Block similarity patterns exhibit remarkably similar behaviors across timesteps, indicating that the block diversity is intrinsic rather than timestep-dependent.
Notably, as the inference step increases, the representational discrepancy tends to grow slightly, aligning with the intuition that cleaner inputs allow blocks to play more distinct roles.

Generally, our analysis provides a comprehensive understanding of representation dynamics and reveals that inter-block discrepancy is key to representation learning in DiTs.
These findings explain existing methods and highlight the critical role of block specialization.
These findings motivate us to design more effective methods to enhance representation diversity for performance improvement and effective training.

\subsection{Weighted Diversity Score}
\label{sec:wds}
The CKA heatmap analysis in~\cref{sec:observations} provides qualitative evidence that representation diversity correlates with generation quality.
However, heatmaps cannot be directly compared across models, scales, or training stages.
To enable rigorous quantitative validation, we propose the \textit{Weighted Diversity Score (WDS)}, a scalar metric summarizing the overall feature diversity across all transformer blocks.

\myparagraph{Definition}
Given a model with $L$ transformer blocks, let $\mathbf{H}_i$ denote the feature matrix of the $i$-th block computed on a batch of samples.
WDS is defined as the distance-weighted complement of pairwise CKA similarities:
\begin{equation}
    \text{WDS} = 1 - \frac{\sum_{i<j} |j-i| \cdot \text{CKA}(\mathbf{H}_i, \mathbf{H}_j)}{\sum_{i<j} |j-i|},
    \label{eq:wds}
\end{equation}
where $|j-i|$ is the layer distance between block $i$ and block $j$.
WDS $\in [0, 1]$, where $0$ indicates complete representational collapse and $1$ indicates fully orthogonal representations across all blocks.
WDS is averaged over $5$ uniformly sampled timesteps on $50,000$ samples in the manuscript.

\begin{figure}[t]
    \centering
    \includegraphics[width=\linewidth]{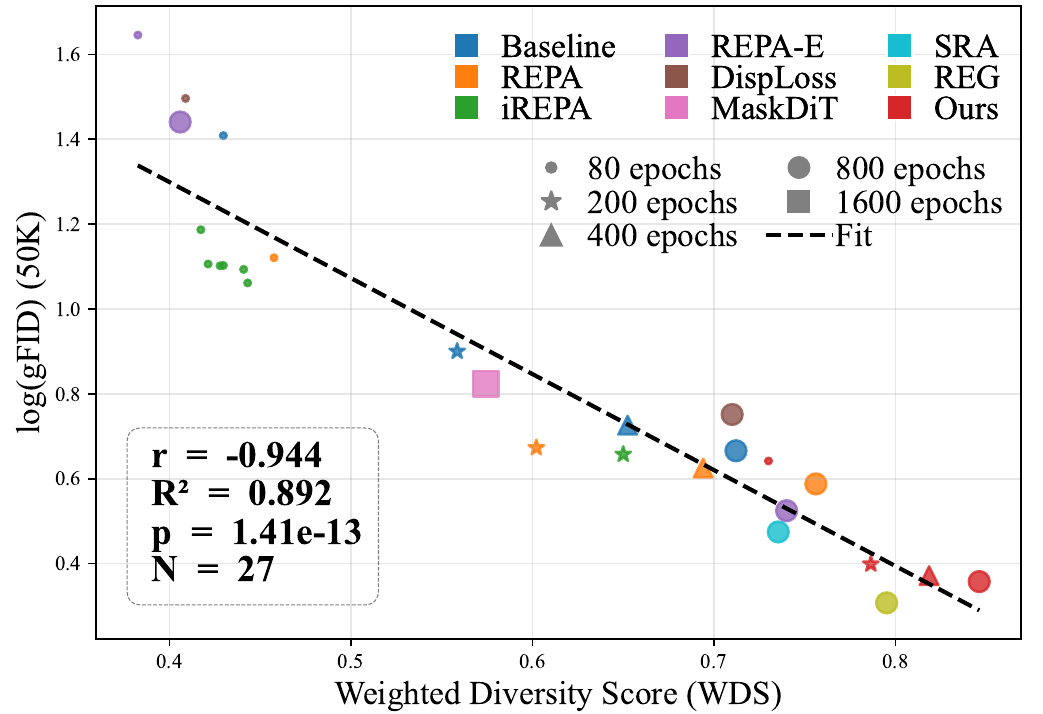}
    \vskip -0.1in
    \caption{\textbf{WDS and $\log(\text{FID})$ correlation for SiT-XL at $255\times256$ resolution.}
    }
    \label{fig:wds_logfid_sit_xl_256}
\end{figure}

\myparagraph{Training dynamics}
To examine how representation diversity evolves during training, we track WDS over training for different methods and model scales in \cref{fig:teaser}(e).
WDS consistently increases during training for all configurations, consistent with \cref{sec:observations}.
Notably, \ourmethod consistently achieves higher WDS than the others at every training step.
Furthermore, larger models attain higher WDS values, suggesting that greater model capacity facilitates representational specialization.
These trends confirm that WDS effectively captures the training dynamics of representation diversity.

\myparagraph{Correlation with generation quality}
To validate whether WDS reliably predicts generation performance, we conduct a correlation study across scales, methods, resolutions, and checkpoints, yielding $97$ data points.
Since FID values span a wide range with a right-skewed distribution~\cite{fid}, we apply the standard log-transformation and analyze the linear relationship between WDS and $\log(\text{FID})$.
As shown in \cref{fig:teaser}(f), WDS exhibits a strong linear correlation with $\log(\text{FID})$ with Pearson coefficient $r=-0.869$, demonstrating that higher block diversity consistently corresponds to lower FID.
Notably, this correlation holds across all model scales, resolutions, and training methods without any grouping or conditioning, confirming that WDS captures a universal structural property of diffusion transformers rather than a method-specific artifact.
Furthermore, WDS generalizes to other DiT/SiT-based methods.
Restricting the analysis to SiT-XL at $256 \times 256$ resolution yields an even stronger correlation of $r=-0.944$, as shown in~\cref{fig:wds_logfid_sit_xl_256}.
We provide the detailed WDS values for this setting in~\cref{tab:res256_512}.
Empirically, a WDS range of $0.75\sim 0.85$ appears to indicate relatively strong model diversity, offering a practical reference for future training and evaluation.

\section{Methodology}
\label{sec:method}

\begin{figure}[t]
    \begin{center}  \centerline{\includegraphics[width=0.8\linewidth]{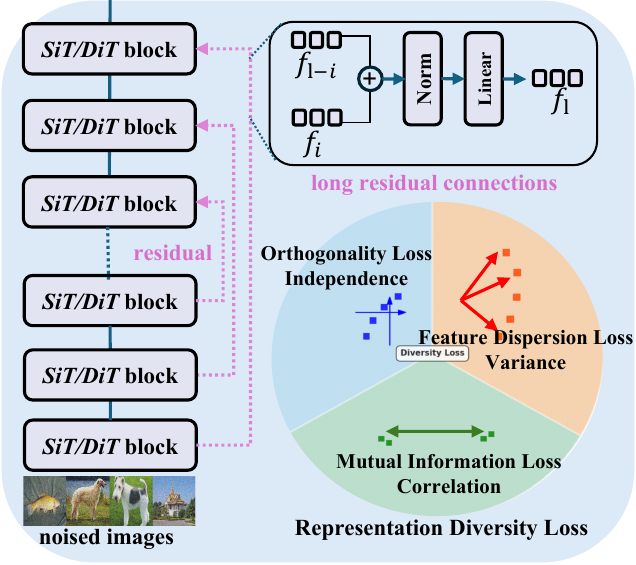}}
    \caption{\textbf{Detailed diagram of our \ourmethod.}
    Our method incorporates long residual connections to diversify input representations across blocks and a representation diversity loss to encourage blocks to learn distinct features.
    }
    \label{fig:method_detail}
    \end{center}
    \vskip -0.2in
\end{figure}

Motivated by the analysis in \cref{sec:revisting_repa}, we introduce \ourmethod, which promotes representation diversity via two components: long residual connections that diversify block inputs, and a representation diversity loss that explicitly encourages inter-block feature specialization.
\cref{fig:method_detail} provides a detailed illustration of our proposed architecture.

\subsection{Long Residual Connections}
Motivated by the findings in \cref{sec:revisting_repa}, we argue that the diversity of inputs for each block plays a crucial role in shaping the learned representations.
However, conventional diffusion transformers often suffer from a lack of input diversity because each block's input is typically homogeneous and is derived solely from the output of the preceding layer.
To address this, we selectively inject early-layer outputs into later layers via long residual connections, promoting feature reuse and preventing representational collapse.
Formally, given a model with $L$ DiT blocks, we connect the $i$-th block's output to the ($L-i$)-th block via:
\begin{equation}
    {f}_l \!=\! \mathcal{R}_{\text{res}}^i({f}_i, {f}_{l-1}) \!=\! \text{Linear}(\text{Norm}({f}_i \oplus {f}_{l-1})),
\end{equation}
where $i \in [0,\ldots,\lfloor L/2 \rfloor - 1]$, $l = L - i$ is the index of the target block that receives the skip connection.
${f}_i \in \mathbb{R}^{N \times T \times D}$ is the representation of the $i$-th block, $\oplus$ denotes concatenating the representation of two blocks ${f}_i$ and ${f}_{l-1}$, which is further processed by a layer normalization and a linear layer.
By injecting skip connections, we break the chain of homogeneous inputs and encourage the network to learn more varied and informative representations from different sources.

\subsection{Representation Diversity Loss}
To further encourage specialization and promote diversity in the learned representations, we introduce a representation diversity loss with three motivations:
1) introducing an explicit inductive bias that encourages block-wise diversity to model the underlying observed distribution;
2) improving the representational orthogonality across blocks, thus reducing the redundancy and mutual correlation of different blocks;
3) promoting the coverage of the representation space, enabling blocks to specialize in complementary structures.

Specifically, our representation diversity comprises three key components:
an orthogonality loss, a mutual-information minimization loss, and a feature dispersion loss.
Notably, to reduce computational cost, we only consider a subset of all possible pairs from $L$ blocks: $\mathcal{P} \subseteq \{(i, j) : i < j,\; i, j \in \{1,\ldots,L\}\}$, with $|\mathcal{P}|=10$ by default.
For each block, we define token-wise mean feature along the $N$ and $T$ dimensions as:
\begin{equation}
    \bm{\mu}_l = \frac{1}{NT} \sum_{n=1}^{N} \sum_{t=1}^{T} f_l[n, t, :] \in \mathbb{R}^{D}.
\end{equation}
Then, we compute the \textit{orthogonality loss} by penalizing high cosine similarity between block-wise mean representations:
\begin{equation}
    \mathcal{L}_{\mathrm{orth}}  \!=\! \frac{1}{|\mathcal{P}|}\! \!\sum_{(i, j)\! \!\in \mathcal{P}}\! \cos(\bm{\mu}_i, \bm{\mu}_j) \!=\! \frac{1}{|\mathcal{P}|} \!\sum_{(i, j)\! \in \mathcal{P}} \frac{\bm{\mu}_i^{\top} \, \bm{\mu}_j}{\lVert \bm{\mu}_i \rVert_2 \lVert \bm{\mu}_j \rVert_2}.
\end{equation}

Next, we minimize mutual information between block representations to ensure statistical independence.
However, directly computing mutual information is computationally intractable for high-dimensional features.
Therefore, we use a computationally efficient proxy based on the average cosine similarity of normalized feature vectors to approximate mutual information.
Specifically, we define flattened, $\ell_2$-normalized token representations along the $N$ and $T$ dimensions as:
\begin{equation}
    \hat{\bm{f}}_{l, n, t} = \frac{f_l[n, t, :]}{\lVert f_l[n, t, :] \rVert_2} \in \mathbb{R}^{D} .
\end{equation}
Then, we compute the \textit{proxy mutual-information loss} as:
\begin{equation}
    \mathcal{L}_{\mathrm{MI}} = \frac{1}{|\mathcal{P}|} \sum_{(i, j) \in \mathcal{P}} \frac{1}{NT} \sum_{n=1}^{N} \sum_{t=1}^{T} \hat{\bm{f}}_{i, n, t}^{\top} \hat{\bm{f}}_{j, n, t} .
\end{equation}
This avoids costly covariance computation while minimizing the correlation between representations.
Further, we employ a feature dispersion loss to encourage diverse channel usage by maximizing the variance of feature activations.
The representations of each block are flattened to $\tilde{f}_l \in \mathbb{R}^{(NT) \times D}$ and normalized along the sample axis to obtain $\widehat{\tilde{f}}_l$.
Then we compute the averaged activation per dimension:
\begin{equation}
    {a} = \frac{1}{|\mathcal{P}|} \sum_{(i,j) \in \mathcal{P}} \frac{1}{2}\!\left[\operatorname{mean}_{n, t}( \widehat{\tilde{f}}_i[n, t, :]) + \operatorname{mean}_{n, t}( \widehat{\tilde{f}}_j[n, t, :])\right],
\end{equation}
${a}$ is then normalized to ${a}'$ by ${a}' = {a} / \max_k a_k$, and its variance is maximized to obtain the \textit{feature dispersion loss}:
\begin{equation}
    \mathcal{L}_{\mathrm{disp}} = -\frac{1}{D} \sum_{k=1}^{D} (a'_k - \bar{a}')^2, \bar{a}' = \frac{1}{D} \sum_{k=1}^{D} a'_k .
\end{equation}
Finally, the \textit{overall representation diversity loss} aggregates the above three components as:
\begin{equation}
    \mathcal{L}_{\text{div}} = \lambda_{\text{orth}}\mathcal{L}_{\text{orth}} + \lambda_{\text{MI}}\mathcal{L}_{\text{MI}} + \lambda_{\text{disp}}\mathcal{L}_{\text{disp}},
\end{equation}
where $\lambda_{\text{orth}}, \lambda_{\text{MI}}, \lambda_{\text{disp}}$ control the relative weight of each loss, we set them as $0.33$ by default without any tuning.

\subsection{Optimization Strategy}

In practice, we find that when $\mathcal{L}_{\text{div}}$ is optimized too small (\emph{e.g.,} close to 0), the model tends to diverge and becomes unable to effectively model the underlying data distribution.
We formalize this via gradient conflict analysis.
Define $G := \langle \nabla_\theta \mathcal{L}_{\text{diff}},\, \nabla_\theta \mathcal{L}_{\text{div}} \rangle$ as the inner product of the two gradient vectors.
The joint update $\nabla_\theta \mathcal{L}_{\text{diff}} + w\nabla_\theta \mathcal{L}_{\text{div}}$ is beneficial to the diffusion objective if and only if:
\begin{equation}
    \|\nabla_\theta \mathcal{L}_{\text{diff}}\|^2 + w \cdot G > 0.
    \label{eq:grad_conflict}
\end{equation}
When $G < 0$ (gradient conflict), this requires $w < \|\nabla_\theta \mathcal{L}_{\text{diff}}\|^2 / |G|$, i.e., the diversity weight must be reduced as conflict intensifies.
We empirically observe that $G$ turns negative once $\mathcal{L}_{\text{div}} \leq \epsilon_{\min}$ (diversity saturated) and progressively declines as training advances and block specialization matures~(\cref{sec:revisting_repa}).
We thus develop an adaptive weight $w$ for the overall $\mathcal{L}_{\text{div}}$:
\begin{equation}
w = \mathrm{clip}\!\left(\frac{\mathcal{L}_{\text{div}} - \epsilon_{\min}}{\epsilon_{\max} - \epsilon_{\min}},\; 0,\; 1\right) \cdot \frac{1 + \cos(\pi t / T)}{2},
\label{eq:adaptive_weight_hard}
\end{equation}
where $\epsilon_{\min} = 0.1$, $\epsilon_{\max} = 0.5$, $t$ denotes the current training step, and $T$ the total steps.
This formulation encodes two conditions motivated by \Cref{eq:grad_conflict}:
1) the \emph{clipping term} enforces a value-based condition, which sets $w\!=\!0$ once $\mathcal{L}_{\text{div}} \leq \epsilon_{\min}$, eliminating gradient conflict when diversity is already saturated.
2) the \emph{cosine decay} enforces a time-based condition, which progressively reduces $w$ as block specialization matures, allowing the model to focus on fine-grained generation details in later training.

\subsection{Generalization to Other Architectures}

Our two components apply to any network of $L$ stacked blocks, provided each block
emits a feature tensor $\mathbf{z}_l \in \mathbb{R}^{B \times M \times D}$,
where $M$ denotes the number of tokens, atoms, or residues.
Under this abstraction, the skip connection and diversity loss take a single unified form:
\begin{equation}
    \mathbf{z}_l = \mathcal{R}_{\text{res}}^i(\mathbf{z}_i, \mathbf{z}_{l-1}),
    \qquad
    \mathcal{L} = \mathcal{L}_{\mathrm{task}} + w(t)\,\mathcal{L}_{\mathrm{div}},
    \label{eq:general}
\end{equation}
where $\mathcal{L}_{\mathrm{div}}$ pools over the $M$ dimension and $\mathcal{L}_{\mathrm{task}}$
is the domain-specific training objective.
We instantiate \cref{eq:general} on two domains.
For \textit{molecular generation}, $\mathbf{z}_l$ is the node feature output of each ``EquivariantBlock'', and $\mathcal{L}_{\mathrm{task}}$ is the DDPM ELBO loss.
For \textit{protein inverse folding}, $\mathbf{z}_l$ is the residue feature output of each ``EncLayer'' block, and
$\mathcal{L}_{\mathrm{task}}$ is the flow matching cross-entropy loss.
In both cases, \ourmethod consistently improves generation diversity and quality, demonstrating that the proposed components are architecture-agnostic and broadly effective.

\section{Experiments}
\label{sec:exps}

\begin{table}[!t]
    \centering
    \scriptsize
    \caption{\textbf{Variation in model-scale on ImageNet without CFG}.
    Our proposed method brings consistent performance gains across all model scales when applied to both SiT and REPA, at both $256\times256$ and $512\times512$ resolutions.
    }
    \setlength{\tabcolsep}{2.8mm}{
    \begin{tabular}{lcccccc}
        \toprule
        \textbf{Model}  & \textbf{Iter}. & \textbf{FID}$_\downarrow$ & \textbf{sFID}$_\downarrow$ & \textbf{IS}$_\uparrow$ & \textbf{Prec.}$_\uparrow$ & \textbf{Rec.}$_\uparrow$ \\
        \midrule
        \multicolumn{7}{c}{\textbf{\textit{ImageNet 256$\times$256}}} \\
        \midrule
        SiT-B             & 400k  & 36.80          & 6.77          & 40.09          & 0.51          & \textbf{0.63} \\
        \textbf{+ (Ours)} & 400k  & \textbf{27.98} & \textbf{6.02} & \textbf{51.07} & \textbf{0.58} & \textbf{0.63} \\
        \hdashline
        REPA-B            & 400k  & 22.99          & 6.70          & 64.73          & 0.59          & \textbf{0.65} \\
        \textbf{+ (Ours)} & 400k  & \textbf{17.22} & \textbf{6.52} & \textbf{80.04} & \textbf{0.63} & \textbf{0.66} \\
        \midrule
        SiT-L             & 400k  & 18.77          & 5.27          & 71.44          & 0.64          & 0.63 \\
        \textbf{+ (Ours)} & 400k  & \textbf{16.01} & \textbf{5.03} & \textbf{79.89} & \textbf{0.66} & \textbf{0.64} \\
        \hdashline
        REPA-L            & 400k  & 9.57           & 5.34 & 113.32         & \textbf{0.69} & 0.66          \\
        \textbf{+ (Ours)} & 400k  & \textbf{8.42}  & \textbf{5.32}          & \textbf{124.21}& \textbf{0.69} & \textbf{0.68} \\
        \midrule
        SiT-XL            & 400k  & 17.43          & 5.11          & 76.00          & 0.64          & \textbf{0.64} \\
        \textbf{+ (Ours)} & 400k  & \textbf{12.38} & \textbf{4.81} & \textbf{95.47} & \textbf{0.68} & \textbf{0.64} \\
        \hdashline
        REPA-XL           & 400k  & 8.73           & 5.21          & 118.68         & 0.69          & 0.65 \\
        \textbf{+ (Ours)} & 400k  & \textbf{8.06}  & \textbf{5.01} & \textbf{123.46}& \textbf{0.70} & \textbf{0.66} \\
        \midrule
        \multicolumn{7}{c}{\textbf{\textit{ImageNet 512$\times$512}}} \\
        \midrule
        SiT-B             & 400k  & 43.46          & 7.53          & 36.80          & 0.60          & 0.64           \\
        \textbf{+ (Ours)} & 400k  & \textbf{32.83} & \textbf{6.76} & \textbf{46.41} & \textbf{0.67} & \textbf{0.65}  \\
        \hdashline
        REPA-B            & 400k  & 30.13          & 7.79          & 53.92          & 0.68          & \textbf{0.64}  \\
        \textbf{+ (Ours)} & 400k  & \textbf{23.77} & \textbf{7.73} & \textbf{65.14} & \textbf{0.70} & \textbf{0.64}           \\
        \midrule
        SiT-L             & 400k  & 22.75          & 5.78          & 64.05          & 0.73          & \textbf{0.63}  \\
        \textbf{+ (Ours)} & 400k  & \textbf{19.16} & \textbf{5.69} & \textbf{71.85} & \textbf{0.76} & 0.62           \\
        \hdashline
        REPA-L            & 400k  & 10.82          & 5.52          & 106.43         & 0.78          & 0.63           \\
        \textbf{+ (Ours)} & 400k  & \textbf{9.78}  & \textbf{5.46} & \textbf{114.25}& \textbf{0.78} & \textbf{0.65}  \\
        \midrule
        SiT-XL            & 400k  & 19.65          & 5.55          & 71.57          & 0.75          & 0.60  \\
        \textbf{+ (Ours)} & 400k  & \textbf{17.63} & \textbf{5.47} & \textbf{76.90} & \textbf{0.77} & \textbf{0.61}  \\
        \hdashline
        REPA-XL           & 400k  & 7.91           & 5.41          & 127.83         & \textbf{0.79} & \textbf{0.65}  \\
        \textbf{+ (Ours)} & 400k  & \textbf{7.17}  & \textbf{5.34} & \textbf{137.34}& \textbf{0.79}          & \textbf{0.65}           \\
        \bottomrule
    \end{tabular}
    }
    \label{tab:different_size}
\end{table}

\subsection{Experiment Setup}

\myparagraph{Implementation Details}
We incorporate \ourmethod into several popular baselines to evaluate its versatility and effectiveness including SiT~\cite{sit}, REPA~\cite{repa} and MeanFlow~\cite{geng2025mean}, leaving other details untouched.
To ensure fair comparison, we strictly follow the training configurations of SiT~\cite{sit}, REPA~\cite{repa} and MeanFlow~\cite{geng2025mean} for experimental evaluation on the ImageNet~\cite{deng2009imagenet} dataset with the resolution of $256\times256$ and $512\times512$, images and pre-computed VAE features are preprocessed following REPA~\cite{repa} with Stable Diffusion VAE~\cite{rombach2022high}.
We adapt the B/2, L/2, and XL/2 model configurations from SiT with a patch size of $2$ to evaluate the scalability.
For training, we use AdamW with a constant learning rate of $1e^{-4}$, ($\beta_1$, $\beta_2$) = ($0.9$, $0.999$) without decay, the batchsize is fixed to $256$, and we adopt mixed-precision (fp$16$) training with gradient clipping.
For performance evaluation, we strictly follow the ADM setup~\cite{dhariwal2021diffusion} and adopt several popular metrics: Fréchet Inception Distance (FID)~\cite{fid}, structural FID (sFID)~\cite{sfid}, Inception Score (IS)~\cite{is}, Precision (Prec.) and Recall (Rec.)~\cite{precrecall}, all calculated from 50K generated images against the official ADM reference batches.
Classifier-free guidance (CFG)~\cite{cfg} is not employed unless specified.
We use the Euler-Maruyama SDE sampler with a diffusion coefficient $\sigma_t$ and 250 sampling steps.
All experiments are conducted on 8 $\times$ 80GB H800 GPUs, with a training speed of approximately 5.8 steps/s for SiT-XL + Ours; generating 50K evaluation images takes about 1.38 hours (10.04 images/s).

\myparagraph{Pretrained vision foundation models}
For representation alignment, we use three pretrained visual encoders as external guidance: DINOv2-B~\cite{oquab2023dinov2}, MAE~\cite{he2022masked}, and MoCov3~\cite{chen2021empirical}.
DINOv2 employs a ViT architecture trained with self-supervised consistency objectives, capturing high-level semantic information.
MAE learns robust representations via masked patch reconstruction.
MoCov3 leverages contrastive learning to maximize similarity between different views of the same image while minimizing similarity across different images.
For all encoders, we perform projection with three MLP layers with SiLU activations following REPA~\cite{repa}.

\myparagraph{Hyperparameter sensitivity}
We apply the same hyperparameter settings when incorporating our method into all baselines (SiT, REPA, DispLoss, SRA, and MeanFlow).
Our method consistently improves performance across different backbones and sampler settings (multi-step and one-step), indicating robustness to hyperparameter choices despite introducing several new hyperparameters.

\myparagraph{Compared baselines}
For multi-step comparison, we compare our method against existing alternatives from four categories:
1) pixel-diffusion: ADM~\cite{dhariwal2021diffusion}, VDM++~\cite{kingma2023understanding}, CDM~\cite{ho2022cascaded};
2) latent-diffusion with UNet: LDM~\cite{rombach2022high};
3) diffusion transformers: SiT~\cite{sit}, DiT~\cite{dit}, SD-DiT~\cite{zhu2024sd}, MaskDiT~\cite{zheng2023fast}, MDTv2~\cite{gao2023mdtv2};
and 4) current state-of-the-art models: REPA~\cite{repa}, REG~\cite{wu2025representation}, E2E-REPA~\cite{leng2025repa}, SRA~\cite{jiang2025no}, DispLoss~\cite{wang2025diffuse}.
For one-step comparison, we compare our method with recent popular methods:
iCT-XL/2~\cite{ict}, Shortcut-XL/2~\cite{shortcut}, IMM-XL/2~\cite{inductive} and Meanflow~\cite{geng2025mean}.

\subsection{Main Results}
In this part, we first present comparison results of applying our methods on different baselines across various scales to investigate its effectiveness and scalability, and then compare our methods against existing SoTA models.

\begin{figure}[!t]
    \centering
    \includegraphics[width=\linewidth]{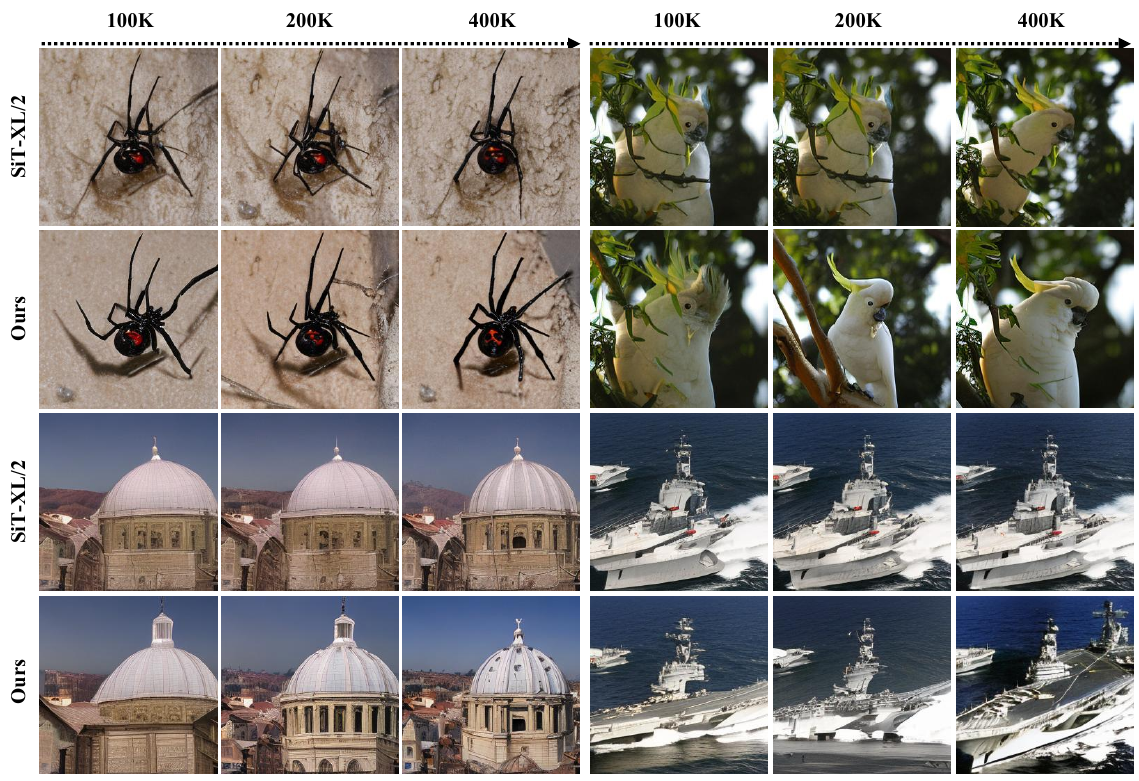}
    \caption{\textbf{Generated samples from different training iterations.}
    Images are sampled using the same seed, noise and class label.
    We use a classifier-free guidance scale of $4.0$ during sampling.
    }
    \label{fig:iter_qualitative}
\end{figure}

\begin{table}[]
\centering
\scriptsize
\caption{\textbf{Comparison results on ImageNet with CFG at $256\times256$ and $512\times512$ resolutions}.}
\vskip -0.1in
\setlength{\tabcolsep}{1.3mm}{
\begin{tabular}{llcccccc}
\toprule
\textbf{Method}     & \textbf{Epochs} & \textbf{WDS$_\uparrow$} & \textbf{FID$_\downarrow$}   & \textbf{sFID$_\downarrow$} & \textbf{IS$_\uparrow$} & \textbf{Pre.$_\uparrow$} & \textbf{Rec.$_\uparrow$} \\
\midrule
\multicolumn{8}{c}{\textbf{\textit{ImageNet $256\times256$}}} \\
\midrule
ADM-U~\cite{dhariwal2021diffusion}
& 400  & -     & 3.94     & 6.14       & 186.70    & 0.82    & 0.52  \\
VDM++~\cite{kingma2023understanding}
& 560  & -     & 2.40     & -          & 225.30    & -       & -     \\
CDM~\cite{ho2022cascaded}
& 2160 & -     & 4.88     & -          & 211.80    & -       & -     \\
LDM-4~\cite{rombach2022high}
& 200  & -     & 3.60     & -          & 247.70    & \textbf{0.87}    & 0.48  \\
SD-DiT~\cite{zhu2024sd}
& 480  & -     & 3.23     & -          & -         & -       & -     \\
MaskDiT~\cite{zheng2023fast}
& 1600 & 0.57     & 2.28     & 5.67       & 276.70    & 0.80    & 0.62  \\
MDTv2-XL/2~\cite{gao2023mdtv2}
& 1080 & 0.75     & 1.58     & 4.52       & \textbf{314.70}    & 0.79    & 0.65  \\
DiT-XL/2~\cite{dit}
& 1400 & 0.71     & 2.27     & 4.60       & 278.20    & 0.83    & 0.57  \\
SiT-XL/2~\cite{sit}
& 1400 & 0.71     & 2.06     & 4.50       & 270.30    & 0.82    & 0.59  \\
REPA~\cite{repa}
& 200  & 0.60     & 1.96     & 4.49       & 264.00    & 0.82    & 0.60  \\
REPA~\cite{repa}
& 800  & 0.76     & 1.80     & 4.50       & 284.00    & 0.81    & 0.61  \\
iREPA~\cite{irepa_singh2025matters}
& 200  & 0.65     & 1.93     & 4.59      & 268.80    & 0.80       & 0.60     \\
REG~\cite{wu2025representation}
& 800  & 0.80     & \textbf{1.36}     & 4.25       & 299.40    & 0.77    & 0.66  \\
E2E-REPA~\cite{leng2025repa}
& 800  & 0.74     & 1.69     & \textbf{4.17}       & 219.30    & 0.77    & \textbf{0.67}  \\
SRA~\cite{jiang2025no}
& 800 & 0.74      & 1.58     & 4.65       & 311.40    & 0.80    & 0.63  \\
DispLoss~\cite{wang2025diffuse}
& 800 & 0.71      & 2.12     & -          & 281.26    & -       & -     \\
DiverseDiT~\cite{diversedit_yang2026} & 200 & 0.78      & 1.52     & 4.23       & 282.72    & 0.81    & 0.66  \\
\midrule
\textbf{DiverseDiT++ (Ours)} & 80 & 0.73       & 1.90     & 4.40       & 278.18    & 0.81    & 0.66  \\
\textbf{DiverseDiT++ (Ours)} & 200 & 0.79      & 1.49     & 4.26       & 283.35    & 0.81    & 0.66  \\
\textbf{DiverseDiT++ (Ours)} & 400 & 0.82      & 1.45     & 4.25       & 292.56    & 0.81    & 0.65  \\
\textbf{DiverseDiT++ (Ours)} & 800 & \textbf{0.85}      & \textbf{1.43}     & \textbf{4.21}       & \textbf{296.43}  & 0.81    & 0.65  \\
\midrule
\multicolumn{8}{c}{\textbf{\textit{ImageNet $512\times512$}}} \\
\midrule
ADM-G~\cite{dhariwal2021diffusion}
& 400 & -        & 2.85     & 5.86       & 221.70    & {\bf0.84}    & 0.53  \\
VDM++~\cite{kingma2023understanding}
& -    & -      & 2.65     & -          & \bf{278.10} & -       & -     \\
MaskDiT~\cite{zheng2023fast}
& 800  & 0.47      & 2.50     & 5.10       & 256.30    & 0.83    & 0.56  \\
DiT-XL/2~\cite{dit}
& 600 & 0.42       & 3.04     & 5.02       & 240.80    & {\bf0.84}    & 0.54  \\
SiT-XL/2~\cite{sit}
& 600 & 0.47       & 2.62     & {\bf4.18}  & 252.20    & {\bf0.84}    & 0.57  \\
REPA~\cite{repa}
& 200 & 0.49       & 2.08     & 4.19       & 274.60    & 0.83    & 0.58  \\
DiverseDiT~\cite{diversedit_yang2026} & 200 & 0.59      & 1.99 & 4.19 & 267.12    & 0.83    & {\bf0.61}  \\
\midrule
\textbf{DiverseDiT++ (Ours)} & 80  & 0.42       & 2.20     & 4.31       & 243.23    & 0.82    & {\bf0.61}  \\
\textbf{DiverseDiT++ (Ours)} & 160 & 0.57        & 2.04     & 4.27       & 261.87    & 0.83    & {\bf0.61}  \\
\textbf{DiverseDiT++ (Ours)} & 200 & \textbf{0.61}       & {\bf1.97}& {\bf4.17}  & 268.34    & \textbf{0.84}    & {\bf0.61}  \\
\bottomrule
\end{tabular}
}
\label{tab:res256_512}
\vskip -0.1in
\end{table}

\myparagraph{Improving representation learning across various model scales}
\cref{tab:different_size} presents the quantitative results of applying our proposed techniques to SiT and REPA across various model scales on ImageNet $256\times256$ without CFG.
We could observe that incorporating our method consistently yields substantial improvements on all evaluation metrics across all model scales, demonstrating its effectiveness and generalization regardless of the underlying training paradigm.
Notably, our method achieves an FID of $17.22$ on the REPA-B setting with $400K$ iterations, which is better than that of SiT-L (\emph{i.e.,} $18.77$) with the same training iterations.
Similarly, the performance of applying our method on the REPA-L outperforms that of the REPA-XL, \emph{i.e.,} $8.42$ vs $8.73$ on FID and $124.21$ vs $118.68$ on IS.
Moreover, \cref{fig:iter_qualitative} shows the images generated by SiT-XL and our proposed method at different training iterations.
Our generated images exhibit more details, better structures, and fewer artifacts, demonstrating that our method leads to faster convergence and higher visual quality compared to the baseline models.
Furthermore, \cref{tab:different_size} also presents the results on ImageNet $512\times512$ across various model scales.
Consistent with the $256\times256$ results, our method yields substantial improvements across all scales.
For instance, when applied to SiT-B, our method reduces FID from $43.46$ to $32.83$, and for REPA-XL, the FID decreases from $7.91$ to $7.17$ with IS increasing from $127.83$ to $137.34$.
These results confirm the generalizability of our approach to higher resolutions.

\begin{figure*}[!t]
    \centering
    \includegraphics[width=.99\linewidth]{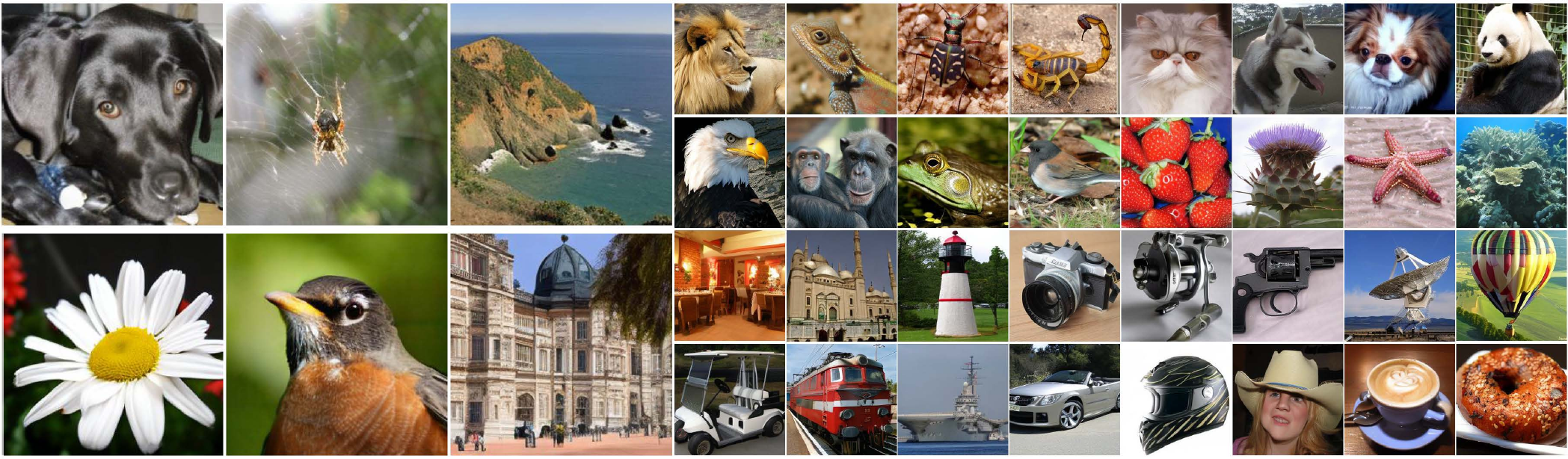}
    \caption{\textbf{Generated samples on ImageNet 256$\times$256 from our DiverseDiT++}.
    We use a classifier-free guidance scale of 4.0.
    }
    \label{fig:qualitative}
\end{figure*}

\myparagraph{Comparison with SoTA Models}
\cref{tab:res256_512} presents the comparison results with recent state-of-the-art (SoTA) methods using CFG on ImageNet $256\times256$.
As can be observed from the table, our method achieves competitive performance compared to SoTA models while requiring significantly fewer training epochs.
At $80$ epochs, our method attains an FID score of $1.90$, outperforming REPA trained for $200$ epochs ($1.96$) and surpassing the performance of several established methods trained for hundreds or even thousands of epochs.
For instance, the SiT-XL/2 model requires $1400$ epochs to reach an FID of $2.06$, while we achieve $1.49$ with only $200$ epochs.
While REG achieves a slightly better FID of $1.36$, it requires $800$ epochs, four times the training cost of ours.
Moreover, with extended training, our method continues to improve: achieving FID of $1.45$ at $400$ epochs and $1.43$ at $800$ epochs, demonstrating sustained gains with longer training schedules.
Furthermore, \cref{tab:res256_512} shows the comparison results on ImageNet $512\times512$.
Similarly, our \ourmethod achieves a comparable FID of $2.20$ with only $80$ epochs and obtains the best FID score of $1.98$ when trained for $200$ epochs.
The performance also improves steadily with more training (FID $2.04$ at $160$ epochs), confirming the sustained benefit of our approach.
The consistent strong performance across multiple metrics, coupled with the significantly reduced training time, shows the efficiency and effectiveness of our model in learning diverse and high-quality representations.
Additionally, we provide selected samples generated by our method in \cref{fig:qualitative}. The generated images demonstrate that \ourmethod produces images with excellent quality.

\myparagraph{Robustness across different alignment depths.}
To further demonstrate the robustness of our method, \cref{tab:different_depth} presents results with different alignment depths on ImageNet $256\times256$ without CFG.
Our method consistently improves performance regardless of the alignment depth configuration.
For example, on REPA-L with alignment depth $5$, our method reduces FID from $12.02$ to $10.03$, and with depth $8$, from $9.57$ to $8.42$.
Similarly, on REPA-XL, our method improves FID from 8.27 to 8.19 (depth $5$) and from $8.73$ to $8.06$ (depth $8$).
These results confirm that our approach is robust to the choice of alignment depth and consistently enhances performance across different configurations.

\begin{table}[!t]
    \centering
    \scriptsize
    \caption{\textbf{Variation in alignment depth on ImageNet 256$\times$256 without CFG}.
    Our method brings consistent gains across different alignment depths.
    }
    \setlength{\tabcolsep}{2.0mm}{
    \begin{tabular}{lccccccc}
        \toprule
        \textbf{Model} & \textbf{Depth} & \textbf{Iter}. & \textbf{FID}$_\downarrow$ & \textbf{sFID}$_\downarrow$ & \textbf{IS}$_\uparrow$ & \textbf{Prec.}$_\uparrow$ & \textbf{Rec.}$_\uparrow$ \\
        \midrule
        REPA-B            & 5 & 400k & 22.99 & 6.70 & 64.73  & 0.59 & 0.65 \\
        \textbf{+ (Ours)} & 5 & 400k & \textbf{17.22} & \textbf{6.52} & \textbf{80.04}  & \textbf{0.63} & \textbf{0.66} \\

        REPA-B            & 8 & 400k & 27.94 & 7.19 & 54.32  & 0.56 & 0.64 \\
        \textbf{+ (Ours)} & 8 & 400k & \textbf{23.24} & \textbf{6.83} & \textbf{62.57}  & \textbf{0.59} & \textbf{0.65} \\
        \midrule
        REPA-L            & 5 & 400k & 12.02 & 6.77 & 40.09  & 0.51 & 0.63 \\
        \textbf{+ (Ours)} & 5 & 400k & \textbf{10.03} & \textbf{5.49} & \textbf{107.31} & \textbf{0.68} & \textbf{0.64} \\
        REPA-L            & 8 & 400k & 9.57  & \textbf{5.34} & 113.42 & \textbf{0.69} & 0.66 \\
        \textbf{+ (Ours)} & 8 & 400k & \textbf{8.42}  & 5.32 & \textbf{124.21} & \textbf{0.69} & \textbf{0.68} \\
        \midrule
        REPA-XL           & 5 & 400k & 8.27  & 5.19 & 123.85 & 0.69 & \textbf{0.66} \\
        \textbf{+ (Ours)} & 5 & 400k & \textbf{8.19}  & \textbf{5.03} & \textbf{128.12} & \textbf{0.70} & \textbf{0.66} \\
        REPA-XL           & 8 & 400k & 8.73  & 5.21 & 118.68 & 0.69 & 0.65 \\
        \textbf{+ (Ours)} & 8 & 400k & \textbf{8.06}  & \textbf{5.01} & \textbf{123.46} & \textbf{0.70} & \textbf{0.66} \\
        \bottomrule
    \end{tabular}
    }
    \label{tab:different_depth}
\end{table}

\myparagraph{Improving representation learning for one-step generation}
We applied our proposed techniques to MeanFlow (MF)~\cite{geng2025mean} to assess the generalization ability.
\cref{tab:different_size_mf} presents the quantitative results across different model scales.
Similar to our previous findings, incorporating our method consistently improves the performance with different model sizes, \emph{e.g.,} our method improves the FID score of MF-B/2 from $9.44$ to $8.49$ and the IS from $152.55$ to $160.07$.
Additionally, \cref{tab:mf} shows a comparison of our method with other one-step generative models.
Notably, we achieve a new SoTA with an FID score of $2.94$ by applying our method to MeanFlow-XL/2.
These results identify the effectiveness of our method in improving representation learning for one-step generation.

\begin{table}[t]
    \centering
    \caption{\textbf{Variation in model scale on ImageNet 256$\times$256 for one-step generation without CFG}.
    }
    \scriptsize
    \setlength{\tabcolsep}{2.2mm}{
    \begin{tabular}{lccccccc}
        \toprule
        \textbf{Model}  & \textbf{Iter}. & \textbf{Step}. & \textbf{FID}$_\downarrow$ & \textbf{sFID}$_\downarrow$ & \textbf{IS}$_\uparrow$ & \textbf{Prec.}$_\uparrow$ & \textbf{Rec.}$_\uparrow$ \\
        \midrule
        MF-B/2            & 400k & 1 & 9.44          & 6.21          & 152.55          & 0.77          & 0.42          \\
        \textbf{+ (Ours)} & 400k & 1 & \textbf{8.49} & \textbf{5.90} & \textbf{160.07} & \textbf{0.78} & \textbf{0.45} \\
        \midrule
        MF-L/2            & 400k & 1 & 8.73          & 6.11          & 161.69          & 0.79          & 0.40          \\
        \textbf{+ (Ours)} & 400k & 1 & \textbf{7.17} & \textbf{5.85} & \textbf{197.83} & \textbf{0.81} & \textbf{0.42} \\
        \midrule
        MF-XL/2           & 400k & 1 & 5.94          & 6.10          & 213.13          & \textbf{0.83} & 0.41 \\
        \textbf{+ (Ours)} & 400k & 1 & \textbf{5.67} & \textbf{5.81} & \textbf{217.16} & \textbf{0.83} & \textbf{0.42} \\
        \bottomrule
    \end{tabular}
    }
    \label{tab:different_size_mf}
\end{table}

\begin{table}[t]
  \scriptsize
  \centering
  \caption{\textbf{Comparison results on ImageNet 256$\times$256 for one-step generation with CFG}.}
  \setlength{\tabcolsep}{3.8mm}{
  \begin{tabular}{lcccr}
    \toprule
    \textbf{Method} & \textbf{Params} & \textbf{Step} & \textbf{NFE} & \textbf{FID$_\downarrow$} \\ \midrule
    iCT-XL/2 \cite{ict}                           & 675M & 1 & 1 & 34.24 \\
    Shortcut-XL/2 \cite{shortcut}                 & 675M & 1 & 1 & 10.60 \\
    IMM-XL/2 \cite{inductive}                     & 675M & 1 & 2 & 7.77 \\
    MF-XL/2 \cite{geng2025mean}                   & 676M & 1 & 1 & 3.43 \\
    MF-XL/2 + DispLoss~\cite{wang2025diffuse}     & 676M & 1 & 1 & 3.21 \\
    MF-XL/2 + \textbf{ours}                       & 713M & 1 & 1 & \textbf{2.94} \\
    \bottomrule
  \end{tabular}
  }
\label{tab:mf}
\end{table}

\subsection{Extending \ourmethod to Other Domains}
\label{sec:exp_other_domains}
To validate the generalization capability of our proposed method, we apply \ourmethod to two additional domains with fundamentally different data modalities and model architectures: protein inverse folding and 3D molecule generation.

\begin{figure}[th]
    \centering
    \includegraphics[width=\linewidth]{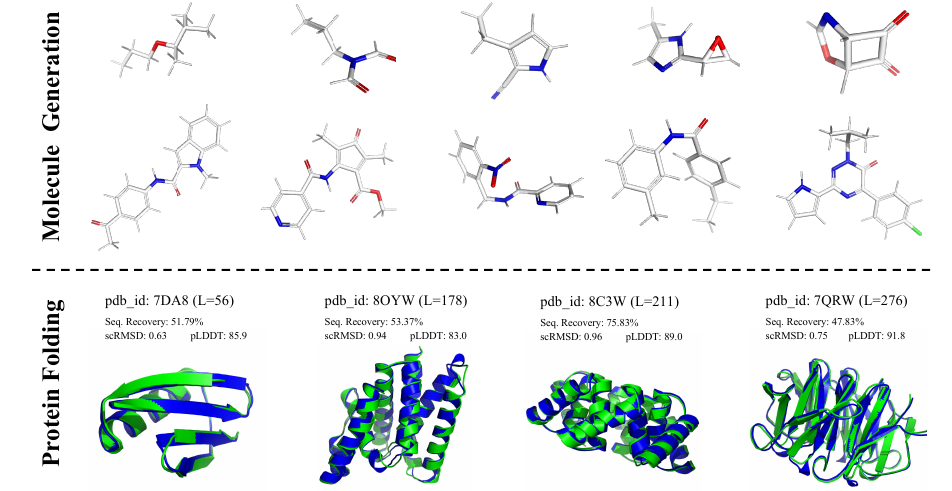}
    \caption{\textbf{Qualitative results on molecule generation and protein folding.}
    }
    \label{fig:qualitative_molecule_protein}
\end{figure}

\myparagraph{Protein Inverse Folding}
We apply our method to protein inverse folding, which generates amino acid sequences conditioned on backbone 3D structures.
Following~\cite{wang2025fine}, we train an inverse folding model using the Multiflow~\cite{multiflow_campbell2024generative} objective and the ProteinMPNN~\cite{proteinmpnn} architecture on the filtered PDB dataset, which filters for single-chain proteins with sequences under $256$ residues, yielding $8,460$ training and $1,145$ test instances.
We evaluate using sequence recovery rate (Seq. Rec.), self-consistency RMSD (scRMSD), and pLDDT from ESMFold~\cite{esmfold}, along with the proportions of scRMSD $< 2$\,\AA{} and pLDDT $> 80$.
As shown in \cref{tab:protein} and \cref{fig:qualitative_molecule_protein}, our method consistently outperforms both the baseline and REED~\cite{reed} across all metrics, achieving higher sequence recovery and better structural quality.
Specifically, DiscreteDiff + Ours achieves a sequence recovery rate of $38.30\%$, scRMSD of $1.81$.
These results indicate that promoting representation diversity effectively accelerates learning in the protein domain.

\begin{table}[!t]
\centering
\scriptsize
\caption{\textbf{Protein inverse folding results on PDB dataset.}
We report the mean across 3 random seeds.}
\label{tab:protein}
\setlength{\tabcolsep}{0.3mm}{
\begin{tabular}{lccccc}
\toprule
\textbf{Method} & \textbf{Seq. Rec.(\%)}$_\uparrow$ & \textbf{scRMSD}$_\downarrow$ & \textbf{\%(sc$<$2)}$_\uparrow$ & \textbf{pLDDT}$_\uparrow$ & \textbf{\%(p$>$80)}$_\uparrow$ \\
\midrule
ProteinMPNN & 37.56 & 1.83 & 53.56 & 81.17 & 65.57 \\
DiscreteDiff & 37.46 & 1.85 & 52.29 & 80.62 & 65.48 \\
DiscreteDiff + REED  & 37.54 & 1.84 & 53.67 & 80.74 & 65.58 \\
DiscreteDiff + \textbf{Ours}  & \textbf{38.30} & \textbf{1.81} & \textbf{54.53} & \textbf{81.96} & \textbf{65.81} \\
\bottomrule
\end{tabular}
}
\end{table}

\myparagraph{3D Molecule Generation}
We further evaluate on 3D molecule generation using the GEOM-Drug~\cite{geomdrug} and QM9~\cite{qm9_ramakrishnan2014} datasets with SemlaFlow~\cite{irwin2025semlaflow}, an $E(3)$-equivariant flow matching model.
We evaluate using atom/molecule stability, validity, energy, and strain energy (kcal$\cdot$mol$^{-1}$) on $5,000$ sampled molecules.
All other training and evaluation settings follow~\cite{irwin2025semlaflow}.
As shown in \cref{tab:molecule}, on GEOM-Drug, our method improves molecule stability from $96.04\%$ to $97.75\%$ and reduces energy from $105.76$ to $99.71$ kcal$\cdot$mol$^{-1}$, outperforming REED ($96.84\%$ / $103.49$) by a clear margin.
The consistent improvement on both datasets shows that diversity regularization generalizes to 3D molecule generation.

\begin{table}[!t]
\centering
\scriptsize
\caption{\textbf{Molecule generation results on GEOM-Drug and QM9.}
Atom/molecule stability and validity are in \%, energy and strain in kcal$\cdot$mol$^{-1}$. Results are based on $5,000$ samples.}
\label{tab:molecule}
\setlength{\tabcolsep}{1.5mm}
\begin{tabular}{cl|ccc|cc}
\toprule
\textbf{Dataset}
& \textbf{Method}
& \textbf{Atom}$_\uparrow$
& \textbf{Mol}$_\uparrow$
& \textbf{Valid}$_\uparrow$
& \textbf{Energy}$_\downarrow$
& \textbf{Strain}$_\downarrow$ \\
\midrule

\multirow{3}{*}{GEOM-Drug}
& SemlaFlow & 99.68 & 96.04 & 93.81 & 105.76 & 64.48 \\
& $\quad$ + REED & 99.73 & 96.84 & 94.27 & 103.49 & 63.25 \\
& $\quad$ + \textbf{Ours} & \textbf{99.82} & \textbf{97.75} & \textbf{94.63} & \textbf{99.71} & \textbf{60.57} \\
\midrule

\multirow{3}{*}{QM9}
& SemlaFlow & 98.98 & 98.53 & 99.30 & 41.77 & 18.29 \\
& $\quad$ + REED & 99.21 & 98.83 & 99.56 & 41.42 & 18.07 \\
& $\quad$ + \textbf{Ours} & \textbf{99.97} & \textbf{99.47} & \textbf{99.83} & \textbf{40.85} & \textbf{17.34} \\

\bottomrule
\end{tabular}
\end{table}

The consistent gains across both discrete biological sequences and equivariant geometric graphs suggest that representation diversity serves as a broadly applicable inductive bias for generative modeling, extending beyond the visual domain.

\subsection{Ablation Analysis}
\label{sec:ablation}
We perform ablation studies to testify the efficacy of components and design choices used in our main experiments.

\myparagraph{Ablation on designed components}
In \cref{tab:ablation_merge}~(a), we present an ablation study that analyzes the contribution of different components of our method by removing each component.
The results clearly demonstrate the importance of both the representation diversity loss and the long residual connections for optimal performance.
Removing the diversity loss (w/o diversity) worsens FID scores for both SiT-B (from $27.98$ to $32.74$) and REPA-B (from $17.22$ to $20.62$).
Similarly, removing the long residual connections (w/o residual) also noticeably increases FID for both baseline models.
These results confirm that both components of \ourmethod play a crucial role in promoting diverse representation learning and improving the model performance.

\begin{table}[t]
    \centering
    \scriptsize
    \caption{\textbf{Ablation study of \ourmethod on ImageNet 256$\times$256 (400K iterations, without CFG).} We examine four aspects: \textbf{(a)} the contribution of each core component (diversity loss and long residual connections); \textbf{(b)} the effect of individual loss terms ($\mathcal{L}_{\text{orth}}$, $\mathcal{L}_{\text{MI}}$, $\mathcal{L}_{\text{disp}}$) within the diversity loss; \textbf{(c)} sensitivity to the adaptive weight range $[\epsilon_{\min}, \epsilon_{\max}]$; \textbf{(d)} compatibility with complementary representation learning approaches (DispLoss~\cite{wang2025diffuse}, SRA~\cite{jiang2025no}).}
    \label{tab:ablation_merge}
    \setlength{\tabcolsep}{2.5mm}{
    \begin{tabular}{lccccc}
        \toprule
         \textbf{Configurations} & \textbf{FID}$_\downarrow$ & \textbf{sFID}$_\downarrow$ & \textbf{IS}$_\uparrow$ & \textbf{Prec.}$_\uparrow$ & \textbf{Rec.}$_\uparrow$ \\
         \midrule
         \multicolumn{6}{c}{\textbf{(a) Contribution of Each Component}} \\
         \midrule
        SiT-B + ${\texttt{Ours}}$    & 27.98         & 6.02          & 51.07         & 0.58          & 0.63         \\
        \hspace{0.5em}w/o ${\texttt{diversity}}$   & 32.74         & 6.41          & 45.01         & 0.55          & 0.63         \\
        \hspace{0.5em}w/o ${\texttt{residual}}$    & 33.65         & 6.48          & 44.26         & 0.54          & 0.63         \\
        \hdashline
        REPA-B + ${\texttt{Ours}}$   & 17.22         & 6.52          & 80.04         & 0.63          & 0.66        \\
        \hspace{0.5em}w/o ${\texttt{diversity}}$   & 20.62         & 6.64          & 73.49         & 0.61          & 0.64         \\
        \hspace{0.5em}w/o ${\texttt{residual}}$    & 18.13         & 6.60          & 75.88         & 0.61          & 0.65         \\
        \midrule
        \multicolumn{6}{c}{\textbf{(b) Diversity Loss Variants}} \\
        \midrule
         REPA-B + ${\texttt{Ours}}$  & 17.22   & 6.52         & 80.04          & 0.63          & 0.66         \\
         \hspace{0.5em}${\texttt{only $\mathcal{L}_{\text{orth}}$}}$  & 18.95   & 6.65         & 75.53          & 0.61          & 0.65         \\
         \hspace{0.5em}${\texttt{only $\mathcal{L}_{\text{MI}}$}}$    & 17.71   & 6.58         & 78.42          & 0.62          & 0.64         \\
         \hspace{0.5em}${\texttt{only $\mathcal{L}_{\text{disp}}$}}$   & 20.84   & 6.77         & 68.79          & 0.61          & 0.65         \\
        \midrule
        \multicolumn{6}{c}{\textbf{(c) Adaptive Weight Range $[\epsilon_{\min}, \epsilon_{\max}]$}} \\
        \midrule
         ${\texttt{Adap}}$ $[0.1, 0.5]$    & 27.98         & 6.02          & 51.07         & 0.58          & 0.63        \\
         ${\texttt{Adap}}$ $[0.2, 0.7]$    & 30.57         & 6.36          & 48.67         & 0.55          & 0.63        \\
         ${\texttt{Adap}}$ $[0.3, 0.9]$    & 31.67         & 6.34          & 46.10         & 0.54          & 0.63        \\
         ${\texttt{Constant}}$             & \texttt{diverge} & -             & -             & -             & -        \\
        \midrule
        \multicolumn{6}{c}{\textbf{(d) Combining with Prior Approaches}} \\
        \midrule
         SiT-B                      & 36.80         & 6.77         & 40.09          & 0.51          & 0.63         \\
         \hspace{0.5em}+ ${\texttt{Ours}}$        & 27.98         & 6.02         & 51.07          & 0.58          & 0.63         \\
         \hspace{0.5em}++ ${\texttt{DispLoss}}$~\cite{wang2025diffuse}   & 24.95         & 6.03         & 57.38          & 0.59          & 0.63         \\
         \hspace{0.5em}+++ ${\texttt{SRA}}$~\cite{jiang2025no}        & 21.92         & 5.89         & 64.80          & 0.61          & 0.64         \\
        \bottomrule
    \end{tabular}
    }
\end{table}

\myparagraph{Effect of diversity loss variants}
In the main experiments, we adopt a combination of different discrepancies for the representation diversity loss, and validate the contribution of each component in \cref{tab:ablation_merge}~(b).
The results show that using all components of the diversity loss (REPA-B + full) achieves the best performance (FID $17.22$, IS $80.04$).
Moreover, removing any single loss component degrades performance.
More importantly, adding any component consistently outperforms the REPA-B baseline (\cref{tab:different_size}), demonstrating their effectiveness in encouraging diverse representation learning.

\myparagraph{Effect of the adaptive range of diversity loss}
In \cref{tab:ablation_merge}~(c), we test the impact of different adaptive ranges on the model performance.
A constant weight leads to divergence, possibly due to excessive representation discrepancy as discussed in \cref{sec:method}.
By contrast, adaptively controlling the diversity loss based on the loss value enables stable training.
In particular, $[\epsilon_{min}, \epsilon_{max}] = [0.1, 0.5]$ yields the best performance, suggesting that a narrower adaptive range effectively promotes diversity without sacrificing image quality.

\myparagraph{Combining with existing methods for further improvement}
\cref{tab:ablation_merge}~(d) shows \ourmethod's compatibility with DispLoss~\cite{wang2025diffuse} and SRA~\cite{jiang2025no}, yielding further performance gains.
The results show that our method can be effectively combined with existing approaches for further improvements, reflecting the method's flexibility.
Noticeably, when combining \ourmethod with both DispLoss and SRA, we achieve an FID of $21.92$, which is better than that of REPA ($22.99$ in \cref{tab:different_size}) at the same iterations.
Recall that REPA requires external models for representation alignment, while here we do not rely on any external guidance, demonstrating the potential for representation learning through internal mechanisms.

\myparagraph{Ablation on layer selection $\mathcal{P}$}
In our implementation, we randomly select $|\mathcal{P}|=10$ layers to compute the diversity loss,
and evaluate the impact of this choice by testing different numbers of selected layers on SiT-L ($24$ layers) for $400K$ training steps.
As shown in \cref{tab:ablation_layer_selection}, selecting more layers benefits but increases training time.
Selecting $10$ layers yields a favorable trade-off between performance and efficiency, achieving an FID of $16.10$ with $21.02$ hours of training, compared to $15.77$ FID but $28.50$ hours when using all layers.

\begin{table}[t]
    \centering
    \scriptsize
    \caption{\textbf{Ablation analysis on selecting different numbers of layers for diversity loss.}}
    \label{tab:ablation_layer_selection}
    \setlength{\tabcolsep}{3.2mm}
    {
    \begin{tabular}{l|cccccc}
\hline
\textbf{$\mathcal{P}$}  & \textbf{SiT-L}  & \textbf{5}    & \textbf{10}   & \textbf{15}   & \textbf{20}   & \textbf{all} \\ \hline
FID$_\downarrow$  & 18.77  & 16.85 & 16.10 & 16.01 & 15.84 & \textbf{15.77} \\
IS$_\uparrow$     & 71.44  & 77.62 & 79.47 & 82.05 & 83.95 & \textbf{85.64} \\
Time (h)          & 18.66  & 19.90 & 21.02 & 23.45 & 25.96 & 28.50          \\
    \bottomrule
    \end{tabular}
    }
\end{table}

\section{Conclusions and Limitations}
\label{sec:conc}

\subsubsection{Conclusions}

In this work, we present a comprehensive analysis of representation learning dynamics in Diffusion Transformers, revealing the critical role of inter-block representation diversity.
To quantify this property, we introduce the Weighted Diversity Score (WDS), which exhibits a strong correlation with synthesis quality across $97$ experimental settings (Pearson's $r=-0.869$ with $\log(\mathrm{FID})$).
Guided by this finding, we propose \ourmethod, a novel and efficient framework that explicitly promotes representation diversity with long residual connections to diversify input and a representation diversity loss to encourage distinct features across blocks, without relying on external guidance.
Extensive experiments demonstrate that \ourmethod consistently improves synthesis quality and accelerates convergence across model scales and training settings, including multi-step and one-step ImageNet generation, protein inverse folding, and 3D molecule generation.
These findings contribute to a deeper understanding of representation learning dynamics in DiTs and offer a practical, effective strategy for boosting their performance across diverse domains, paving the way for future research in learning representations for generative models.

\subsubsection{Limitations}
\label{sec:discussion}

\ourmethod demonstrates consistent improvements across tasks and architectures, yet several aspects warrant further discussion.
Methodologically, the long residual connections introduce additional parameters ($+5.5\%$), which may pose constraints in resource-limited or latency-sensitive scenarios.
In addition, the adaptive weight thresholds $\epsilon_{\min}$ and $\epsilon_{\max}$ are determined empirically on ImageNet, and the full performance potential of the diversity loss remains to be unlocked through more systematic hyperparameter exploration on new tasks and architectures.
The current evaluation focuses on class-conditional image generation, whether \ourmethod generalizes effectively to text-to-image synthesis and image editing remains to be explored.
Finally, exploring the theoretical links between diversity and properties such as robustness and generalization, as well as extending this analysis to large language models, remains promising open directions.

\bibliographystyle{IEEEtran}
\bibliography{ref}

\clearpage

\appendices
\newcommand{\AppendixPrefix}{A}
\renewcommand{\thefigure}{\AppendixPrefix\arabic{figure}}
\setcounter{figure}{0}
\renewcommand{\thetable}{\AppendixPrefix\arabic{table}}
\setcounter{table}{0}
\renewcommand{\theequation}{\AppendixPrefix\arabic{equation}}
\setcounter{equation}{0}

\section{Appendix Overview}
This supplementary material is organized as follows:
First, we provide a heuristic analysis of the WDS--FID relationship in~\cref{sec:supp:theory}, motivating the connection between representation diversity, score estimation error, and generation quality.
Next, we present more implementation details in~\cref{sec:supp:implementation_details}, including detailed setups for our extension to protein inverse folding and 3D molecule generation.
Followed by the evaluation details and brief introduction of comparison baseline methods in~\cref{sec:supp:eval} and~\cref{sec:supp:comparison_baselines}, respectively, where domain-specific evaluation metrics and scientific baselines are also described.
Then, \cref{sec:supp:analysis_details} shows the detailed quantitative results of our comprehensive analysis.
In the following, we present more quantitative comparison results under various settings in~\cref{sec:supp:more_quantitative} and more ablation results in~\cref{sec:supp:more_ablation}.
Finally, \cref{sec:supp:more_qualitative_results} illustrates more uncurated images generated by our proposed method.

\section{Heuristic Analysis: Why WDS Correlates with Generation Quality}
\label{sec:supp:theory}

We provide an \emph{intuitive, heuristic motivation} for the empirically observed linear relationship between WDS and $\log(\text{FID})$.
The argument proceeds in the following four steps.

\subsection{Step 1: WDS as a Proxy for Representational Capacity}
\label{sec:supp:theory:step1}

\noindent\textbf{Effective rank.}
Following~\cite{roy2007effective}, the \emph{effective rank} of a matrix $\mathbf{M}$ with singular values $\sigma_1 \geq \cdots \geq \sigma_r > 0$ is defined as:
\begin{equation}
\begin{aligned}
    &\operatorname{erank}(\mathbf{M}) = \exp\!\bigl(H(\mathbf{p})\bigr), \\
    &H(\mathbf{p}) = -\sum_{k=1}^{r} p_k \log p_k,
    \quad p_k = \frac{\sigma_k}{\sum_{j} \sigma_j}.
\end{aligned}
    \label{eq:supp_erank}
\end{equation}
It measures how uniformly $\mathbf{M}$ spreads its energy across directions, satisfying $1 \le \operatorname{erank}(\mathbf{M}) \le \operatorname{rank}(\mathbf{M})$.

\noindent\textbf{From CKA to effective rank.}
For two Frobenius-normalised feature matrices $\mathbf{H}_i, \mathbf{H}_j \in \mathbb{R}^{n \times d}$, linear CKA reduces to~\cite{kornblith2019similarity}:
\begin{equation}
    \mathrm{CKA}(\mathbf{H}_i, \mathbf{H}_j)
    = \frac{\|\mathbf{H}_i^{\!\top}\mathbf{H}_j\|_F^2}
           {\|\mathbf{H}_i^{\!\top}\mathbf{H}_i\|_F\,\|\mathbf{H}_j^{\!\top}\mathbf{H}_j\|_F}.
    \label{eq:supp_linear_cka}
\end{equation}
At $\mathrm{CKA}=1$, both blocks span the same column space and their concatenation gains no new directions; at $\mathrm{CKA}=0$, the blocks are orthogonal and effective rank approximately doubles.
Interpolating between these boundary cases, for a block pair with per-block effective rank $r$ and $c_{ij} = \mathrm{CKA}(\mathbf{H}_i, \mathbf{H}_j)$:
\begin{equation}
    \operatorname{erank}\!\left([\mathbf{H}_i^{\!\top}|\mathbf{H}_j^{\!\top}]^{\!\top}\right)
    \;\approx\; r\,\bigl(1 + (1-c_{ij})\bigr).
    \label{eq:supp_two_block}
\end{equation}
Extending this intuition to all $L$ blocks, we heuristically weight each pair by $|j-i|$, since blocks that are farther apart tend to operate at different abstraction levels~\cite{zeiler2014visualizing} and are therefore expected to contribute more independent directions. This yields the heuristic bound:
\begin{equation}
    \operatorname{erank}(\widetilde{\mathbf{H}})
    \;\gtrsim\; r \cdot \bigl[1 + (L-1)\cdot\mathrm{WDS}\bigr],
    \label{eq:supp_lemma}
\end{equation}
where $\widetilde{\mathbf{H}} = [\mathbf{H}_1^{\!\top}\cdots\mathbf{H}_L^{\!\top}]^{\!\top}$ is the row-stacked aggregated representation.
\emph{Note:} The $L$-block extension assumes additive entropy contributions across pairs (may overestimate gains when pairs are correlated); the argument also uses linear CKA while empirical WDS is computed with RBF-kernel CKA. \Cref{eq:supp_lemma} is therefore a conceptual motivation, not a quantitative bound.

\subsection{Step 2: Effective Rank and Score Estimation Error}
\label{sec:supp:theory:step2}
A Diffusion Transformer with $L$ blocks learns a velocity field $v_\theta(\mathbf{x}_t,t)$ that approximates the true score $\nabla_{\mathbf{x}}\log p_t(\mathbf{x})$.
The quality of this approximation depends on the diversity of intermediate representations. In particular, the utilisation ratio $\operatorname{erank}(\widetilde{\mathbf{H}})/d$ serves as a proxy for the effective dimensionality of the learned feature space: higher diversity distributes energy across more independent directions, providing a richer basis for parameterizing complex score fields.
Under the phenomenological assumption that score error decreases monotonically with utilisation ratio (modelling the prediction head as linear, a deliberate simplification):
\begin{equation}
\begin{aligned}
    \mathcal{E}_\theta
    \;&:=\; \mathbb{E}_{\mathbf{x}_*, \epsilon, t}\!\left[\|v_\theta(\mathbf{x}_t, t) - \nabla_{\mathbf{x}}\log p_t(\mathbf{x}_t)\|^2\right] \\
    &\;\lesssim\; C_s^2\!\left(1 - \frac{\operatorname{erank}(\widetilde{\mathbf{H}})}{d}\right),
\end{aligned}
\label{eq:supp_score_bound}
\end{equation}
where $C_s^2$ encodes the complexity of the data score, independent of model architecture.
This is a \emph{monotone ordering hypothesis} (higher $\operatorname{erank}/d$ implies lower $\mathcal{E}_\theta$), not a tight bound.
Substituting \cref{eq:supp_lemma}:
\begin{equation}
    \mathcal{E}_\theta
    \;\lesssim\; C_s^2\!\left(1 - \frac{r\,\bigl[1 + (L-1)\cdot\mathrm{WDS}\bigr]}{d}\right).
    \label{eq:supp_score_wds}
\end{equation}
This formalises the intuition that \emph{higher WDS is associated with lower score estimation error}.

\subsection{Step 3: Connecting Score Error to FID}
\label{sec:supp:theory:step3}

Chen~\emph{et al.}~\cite{chen2022sampling} proved that under mild moment conditions on $p_\text{data}$:
\begin{equation}
    W_2^2(p_\text{gen},\, p_\text{data})
    \;\leq\; C_T \int_0^T \mathcal{E}_\theta(t)\,\mathrm{d}t,
    \label{eq:supp_w2_score}
\end{equation}
where $C_T$ depends only on the SDE schedule and is bounded for finite $T$.
Treating FID as a proxy for $W_2^2$ in Inception feature space~\cite{fid} and bounding via the $\kappa$-Lipschitz Inception map:
\begin{equation}
    \mathrm{FID}(p_\text{gen}, p_\text{data})
    \;\lesssim\; \kappa^2 \cdot W_2^2(p_\text{gen}, p_\text{data}).
    \label{eq:supp_fid_w2}
\end{equation}
Combining \cref{eq:supp_w2_score,eq:supp_fid_w2}, approximating $\mathcal{E}_\theta(t) \approx \mathcal{E}_\theta$ as roughly constant over $[0,T]$ (a coarse approximation), and substituting \cref{eq:supp_score_wds} with $A := \kappa^2 C_T T C_s^2$:
\begin{equation}
    \mathrm{FID}
    \;\lesssim\; A \cdot \left(1 - \frac{r\,\bigl[1 + (L-1)\cdot\mathrm{WDS}\bigr]}{d}\right),
    \label{eq:supp_fid_wds}
\end{equation}
where $A$ is a positive constant fixed by the data distribution, SDE schedule, and Inception network, independent of model architecture or WDS.

\subsection{Step 4: The Approximate Linear Relationship}
\label{sec:supp:theory:step4}
Across our 97 evaluation points, raw FID spans a wide range (${\approx}1.4$--$43$) and is strongly right-skewed.
Such right-skewed non-negative metrics are routinely log-transformed before linear regression to stabilise variance and symmetrise the distribution.
We verify empirically that the WDS--$\log(\mathrm{FID})$ scatter is substantially more homoscedastic than WDS--FID, confirming $\log(\mathrm{FID})$ as the appropriate response variable.
As shown below, this choice is also theoretically consistent: the bound from Step~3 admits a natural log-linear form.

Let $\alpha = r(L-1)/d$ and $\delta = r/d$; \cref{eq:supp_fid_wds} becomes:
\begin{equation}
\begin{aligned}
    \log(\mathrm{FID}) &\;\lesssim\; A\,(1 - \delta - \alpha\,\mathrm{WDS}) \\
    &\;=\; \underbrace{A(1-\delta)}_{\text{const}} - \underbrace{A\alpha}_{\text{slope}} \cdot \mathrm{WDS}.
\end{aligned}
\label{eq:supp_fid_approx}
\end{equation}
The upper bound on $\log(\mathrm{FID})$ decreases linearly with WDS.
The slope $-Ar(L-1)/d$ grows with depth $L$ and per-block rank $r$, consistent with the stronger correlation for larger models.
Since $A$ and $r$ are absorbed into the intercept and different model sizes converge to similar WDS ranges (similar $\bar{u}$), the slope is approximately stable across settings, explaining why a single linear fit describes all 97 data points.

\noindent\textbf{Empirical corroboration.}
\Cref{eq:supp_fid_approx} is an upper bound, not an equality; the observed Pearson correlation ($r=-0.869$ across 97 data points, $r=-0.944$ for SiT-XL/2 alone) is consistent with WDS being the dominant factor, with other sources of variation (training convergence, architecture details) contributing approximately constant residuals.
The analysis identifies \emph{reasonable mechanisms} through which WDS may influence FID.
The strong empirical correlation is the primary evidence, and this heuristic offers a conceptual framework for why such a relationship is expected.

\section{More Implementation Details}
\label{sec:supp:implementation_details}

\begin{table}[h!]
    \centering
    \small
    \caption{
    \textbf{Architecture configurations for the three model scales used in main experiments.}
    }
    \begin{tabular}{l c c c}
        \toprule
        & \textbf{B} & \textbf{L} & \textbf{XL} \\
        \midrule
        Num. layers & 12 & 24 & 28 \\
        Hidden dim. & 768 & 1,024 & 1,152 \\
        Num. heads  & 12 & 16 & 16 \\
        \bottomrule
    \end{tabular}
    \label{tab:supp_hyperparam}
\end{table}

\paragraph{More implementation details}
We implement our proposed techniques on the original SiT~\cite{sit} and REPA~\cite{repa} implementation, leaving other details unchanged.
Regarding the representation diversity loss, we calculate the corresponding loss items following the definition of each loss in main paper. We randomly select $10$ layers for computing the loss, as we find that computing on more layers gains similar performance improvement.
Such observation aligns with that of DispLoss~\cite{wang2025diffuse}, where the effect of representation diversity loss propagates to other blocks, even though it is not directly applied to them.
The architecture-specific configurations for the three model scales are summarized in \cref{tab:supp_hyperparam}.
For REPA alignment, we set $\lambda = 0.5$, use cosine similarity as the alignment metric, and adopt DINOv2-B~\cite{oquab2023dinov2} as the external encoder; the alignment depth is 5 for B-scale models and 8 for L- and XL-scale models.
To speed up the training process and save pre-processing time, we adopt mixed-precision (fp16) with gradient clipping and pre-compute VAE latents with stable diffusion VAE~\cite{rombach2022high} (sd-vae-ft-mse) following REPA.
For all optimization, we adopt AdamW~\cite{loshchilov2017decoupled} with a learning rate of 1e-4, momentum parameters $(\beta_1, \beta_2) = (0.9, 0.999)$, and a batch size of 256.
Additionally, we implement our method on the official MeanFlow~\cite{geng2025mean} and follow their default configurations for training and evaluation for one-step generation experimental evaluation.

\paragraph{Classifier-free guidance}
We adopt three guidance settings across the main paper.
For experiments reported \emph{without} classifier-free guidance (Table~II, Table~IV, Table~V, Table~IX, Table~X), we set the CFG scale to 1.0 (i.e., no guidance).
We follow the default CFG scale of REPA~\cite{repa} and use a scale of 1.8 (Table~I, Table~VI), except for Table~III with a scale of 4.0.
In the supplementary, we additionally provide results with CFG scale 1.35 in \cref{tab:supp_different_depth_cfg}, \cref{tab:supp_different_size_512_cfg}, \cref{tab:ablation_components_supp}, \cref{tab:ablation_loss_supp}, and \cref{tab:compatibility_supp}   to study the effects under an intermediate guidance strength.

\paragraph{Sampler}
Following the practice of REPA, we employ the Euler-Maruyama sampler with the SDE sampling with a diffusion coefficient $\sigma_t$. The sampling step for generating each image is set to 250.

\paragraph{Discussion on hyperparameter sensitivity}
In our evaluation, we use the same hyperparameter settings when applying our method to various models, namely SiT~\cite{sit}, REPA~\cite{repa}, DispLoss~\cite{wang2025diffuse}, SRA~\cite{jiang2025no} and MeanFlow~\cite{geng2025mean}.
Our method consistently improves performance across different backbones and different sampler settings (multi-step and one-step sampling).
These results indicate that although our diversity loss introduces several hyperparameters, it is robust to hyperparameter choices.

\paragraph{Computing resources}
All experiments were conducted on NVIDIA H100 (80GB) or H200 (141GB) GPUs.
For training, the speed is about 5.8 steps/s for training SiT-XL + Ours, and it takes about 1.38 hours to generate 50,000 images for evaluation (10.04 images/s).
We have uploaded the compute report for detailed GPU hours used for our analysis and evaluation.

\paragraph{Extension to protein inverse folding}
We adopt the ProteinMPNN~\cite{proteinmpnn} architecture as the base model, which consists of a structure encoder and a sequence decoder.
The encoder processes the 3D backbone structure using message passing on a $k$-nearest neighbor graph constructed from C$\alpha$ distances, and the decoder autoregressively generates amino acid sequences conditioned on the encoded structure features.
We train the model using the Multiflow~\cite{multiflow_campbell2024generative} discrete diffusion objective with 500 diffusion timesteps and a temperature of 0.1 during sampling.
Following~\cite{wang2025fine}, we use the PDB training set from ProteinMPNN~\cite{proteinmpnn}, filtering for single-chain proteins with sequences under 256 residues, yielding 8,460 training and 1,145 test instances.
Our representation diversity loss is applied across both the encoder and decoder layers with the same hyperparameter settings as in image generation, including the adaptive weight range $[0.1, 0.5]$ and random selection of 10 paired layers for computing the loss.
The long residual connections are applied within the encoder and decoder separately.

\paragraph{Extension to 3D molecule generation}
Our base model for 3D molecule generation is SemlaFlow~\cite{irwin2025semlaflow}, which employs an $E(3)$-equivariant backbone for processing molecular graphs in 3D space using the ODE flow matching framework.
We adopt the GEOM-Drug~\cite{geomdrug} dataset, which contains large drug-like molecules with an average of 44 atoms. Following SemlaFlow~\cite{irwin2025semlaflow}, we filter out molecules with more than 72 atoms in the training set and use the standard train/validation/test splits.
We train our model for 200 epochs and use 100 NFE for sampling.
Our representation diversity loss is applied across the equivariant backbone layers to encourage diverse feature learning. Importantly, the diversity loss operates on the invariant scalar features within the equivariant framework, thus preserving the $E(3)$-equivariance of the model while still promoting representational diversity.
We use the same adaptive range and layer selection strategy as in other experiments.

\paragraph{Pretrained vision foundation models for representation alignment}
In our systematic analysis and experimental evaluation, we use three pretrained visual encoders as external representation guidance, namely DINOv2-B~\cite{oquab2023dinov2}, MAE~\cite{he2022masked}, and MoCov3~\cite{chen2021empirical}.
For DINOv2-B\footnote{\url{https://github.com/facebookresearch/dino}} and MAE\footnote{\url{https://github.com/facebookresearch/mae}} models, we download the pretrained model officially released by the original authors.
Regarding the MoCov3 model, we download the -L version from the implementation of RCG\footnote{\url{https://github.com/LTH14/rcg}}~\cite{li2023return}; following REPA.
For representation alignment, we perform projection with three MLP layers with SiLU activations following the exact configuration of REPA.
\begin{itemize}
\item \textbf{DINOv2}~\cite{oquab2023dinov2}: DINOv2 employs a vision transformer (ViT) architecture and learns self-supervised representations by enforcing consistency between different views of an image. It measures the feature distance between the representations of real and generated images, capturing high-level semantic information.
\item \textbf{MAE}~\cite{he2022masked}: MAE trains an encoder and a lightweight decoder with a reconstruction objective. It learns to reconstruct masked patches of an image, learning robust representations.
\item \textbf{MoCov3}~\cite{chen2021empirical}: based on the philosophy of contrastive learning, MoCov3 empirically revisits prior MoCo series~\cite{he2020momentum, chenxinlei2020improved} and scales to larger model sizes to learn representations by maximizing the similarity between different views of the same image while minimizing the similarity between views of different images.
\end{itemize}

\section{Evaluation Details}
\label{sec:supp:eval}

\paragraph{Implementation details for evaluation}
We strictly follow the setup and use the same reference batches of ADM~\cite{dhariwal2021diffusion} for evaluation, following their official implementation.
Specifically, for 256$\times$256 evaluation, we generate 50,000 images and convert them into a .npz file and compute the quantitative metrics from the reference batch (VIRTUAL\_imagenet256\_labeled.npz) of ADM\footnote{\url{https://github.com/openai/guided-diffusion/tree/main/evaluations}}.
Similarly, we quantify the result of 512$\times$512 evaluation via computing the metrics between our generated images and the reference batch (VIRTUAL\_imagenet512.npz).

\paragraph{Evaluation metrics}
We adopt several popular metrics for evaluation: Fréchet Inception Distance (FID)~\cite{fid}, structural FID (sFID)~\cite{sfid}, Inception Score (IS)~\cite{is}, Precision (Prec.) and Recall (Rec.)~\cite{precrecall}. Their main concepts are:
\begin{itemize}
\item \textbf{FID}~\cite{fid} computes the Fr\'{e}chet Distance between two observed data distributions, which represent the feature distributions of synthesized and real images extracted by the pre-trained Inception-V3~\cite{szegedy2016rethinking}.
Formally,  FID is calculated by
\begin{align}
\mathrm{FID(X,Y)}=\left\|\mu_s-\mu_r\right\|^2+\operatorname{Tr}\left(\Sigma_s+\Sigma_r-2\left(\Sigma_s \Sigma_r\right)^{\frac{1}{2}}\right),
\end{align}
where $\mathrm{X}$ and $\mathrm{Y}$ represent the synthesized distribution and real  distribution, respectively.
$\mu$ and $\Sigma$ correspond to the mean and covariance matrix of the distribution, and $\operatorname{Tr(\cdot)}$ is the trace operation.
\item \textbf{sFID}~\cite{sfid} is a variant of FID that aims to be more robust to structural differences between real and generated images. Instead of using the standard Inception-V3 features, sFID uses features extracted from different layers of the network, focusing on structural information. This makes it more sensitive to the arrangement of objects and their parts, and less sensitive to color or texture differences.
\item \textbf{IS}~\cite{is} measures the quality and diversity of generated images. It uses the Inception-V3 model to predict the class of each generated image. A good Inception Score means that the generated images are clear and belong to a specific class (high confidence), and that the generated images cover a wide range of classes (high diversity).
Formally, Inception Score is calculated by:
\begin{align}
\mathrm{IS} = \exp(\mathbb{E}_{\mathrm{x} \sim \mathrm{X}}[D_{KL}(p(\mathrm{y}|\mathrm{x}) || p(\mathrm{y}))])
\end{align}
where $\mathrm{x}$ represents the generated images, $\mathrm{X}$ is the distribution of generated images.
$p(\mathrm{y}|\mathrm{x})$ is the conditional probability distribution of the class $\mathrm{y}$ given the image $\mathrm{x}$, predicted by the Inception model, $p(\mathrm{y})$ is the marginal probability of class $\mathrm{y}$.
$D_{KL}$ is the Kullback-Leibler divergence.
\item \textbf{Precision and Recall}~\cite{precrecall} are used to evaluate the quality of generated images by comparing them to real images. Precision measures how much the generated images resemble real images, while Recall measures how much of the real image distribution is captured by the generated images.
\item \textbf{Centered Kernel Alignment} (CKA) is a widely adopted metric for quantifying neural network representations~\cite{davari2022reliability, kornblith2019similarity}, which has been demonstrated with several advantages:
1) CKA is invariant to orthogonal transformation and isotropic scaling, thus it is stable under various image transformations;
2) CKA can capture the non-linear correspondence between representations benefiting from its kernel mapping in the kernel space;
and 3) CKA can quantify the correspondence between different features across different widths, whereas previous metrics fail~\cite{kornblith2019similarity}.
Formally, CKA is normalized from Hilbert-Schmidt Independence Criterion (HSIC)~\cite{HSIC} to be invariant to orthogonal
transformation and isotropic scaling:
\begin{align}
    \label{eq:supp:cka}
    \mathrm{CKA(X,Y)}=\frac{\mathrm{HSIC}(\mathrm{x},\mathrm{y})}{\sqrt{\mathrm{HSIC}(\mathrm{x},\mathrm{x}) \mathrm{HSIC}(\mathrm{y},\mathrm{y})}}.
\end{align}
HSIC identifies whether two distributions ($\mathrm{X,Y}$) are independent: $\mathrm{HSIC}(K,L) \!=\! \frac{1}{(n-1)^2}\operatorname{Tr}(K H L H)$, where $K_{i j}\!=\!k\left(\mathrm{x}_i, \mathrm{x}_j\right)$ { and } $L_{i j}\!=\!l\left(\mathrm{y}_i, \mathrm{y}_j\right)$, where $k$ and $l$ are kernels.
Note that for kernel selections of $k$ and $l$ in \cref{eq:supp:cka}, we find that different kernels (RBF, polynomial, and linear) reflect similar discrepancies across various representations of DiTs, while the RBF kernel contributes to the distinguishability of quantitative results.
\end{itemize}

\paragraph{Evaluation metrics for protein inverse folding}
All metrics are reported as the mean across 3 random seeds.
We assess inverse folding performance using:
\begin{itemize}
\item \textbf{Sequence Recovery Rate}: the proportion of correctly recovered amino acids compared to the native sequence.
\item \textbf{Self-consistency RMSD (scRMSD)}: measures how closely the generated sequence folds back to the target structure. We use ESMFold~\cite{esmfold} to predict the structure from the generated sequence and compute the RMSD against the target backbone.
\item \textbf{pLDDT}: the predicted Local Distance Difference Test score from ESMFold, indicating the confidence of structure predictions. The higher values suggest more designable sequences.
\item \textbf{\%(scRMSD $<$ 2\AA)}: the proportion of cases where scRMSD is below 2\AA, a standard threshold for successful inverse folding~\cite{multiflow_campbell2024generative}.
\item \textbf{\%(pLDDT $>$ 80)}: the proportion of cases where pLDDT exceeds 80, indicating high-confidence structures.
\end{itemize}

\paragraph{Evaluation metrics for 3D molecule generation}
All metrics are computed from 5,000 sampled molecules.
We adopt standard benchmark metrics for 3D molecule generation:
\begin{itemize}
\item \textbf{Atom stability}: the fraction of atoms with correct valency (number of bonds matches the expected valency for the atom type).
\item \textbf{Molecule stability}: the fraction of molecules where all atoms are stable.
\item \textbf{Validity}: the fraction of generated molecules that can be converted to valid SMILES strings.
\item \textbf{Energy} (kcal$\cdot$mol$^{-1}$): the potential energy of the generated conformation computed using the Universal Force Field (UFF) in RDKit, indicating physical plausibility.
\item \textbf{Strain energy} (kcal$\cdot$mol$^{-1}$): the difference between the energy of the generated conformation and the energy-minimized conformation, measuring geometric strain.
\end{itemize}

\section{Comparison Baselines}
\label{sec:supp:comparison_baselines}

In this part, we briefly introduce the main concept of baseline methods that are used for our evaluation.

\subsection{Multi-step baseline models}
\begin{itemize}
\item \textbf{ADM}~\cite{dhariwal2021diffusion} achieved improved synthesis performance with architectural improvement on traditional Unet-based diffusion models and developed classifier guidance to improve the synthesis fidelity for class-conditional tasks.
\item \textbf{VDM++}~\cite{kingma2023understanding} demonstrated that commonly used diffusion model objectives equate to a weighted integral of ELBOs over different noise levels, where the weighting depends on the specific objective used. Based on this, a sample adaptive noise schedule was introduced for improved training efficiency.
\item \textbf{CDM}~\cite{ho2022cascaded} proposed a cascaded architecture that trains multiple models across different resolutions, starting from the lowest resolution to higher resolution.
\item \textbf{LDM}~\cite{rombach2022high} developed latent diffusion models that train diffusion in a low-dimensional compressed latent space to improve the training efficiency. Specifically, the images are first encoded into latent codes and then added noise for training, and the denoised latents are decoded back to pixel space for sampling.
\item \textbf{MDTv2}~\cite{gao2023mdtv2} introduced an asymmetric encoder-decoder paradigm for efficient training of diffusion transformer. To stabilize the training and improve model performance, they further employ U-Net-like long-shortcuts in the encoder and dense input-shortcuts in the decoder.
\item \textbf{MaskDiT}~\cite{zheng2023fast} used a similar encoder-decoder architecture with MDTv2, while the model was trained with an auxiliary reconstruction objective like \cite{he2022masked} to reconstruct masked inputs.
\item \textbf{SD-DiT}~\cite{zhu2024sd} extended the reconstruction-based MaskDiT architecture, while introducing a self-supervised discrimination objective with a momentum encoder for improved training.
\item \textbf{DiT}~\cite{dit} proposed to replace the conventional Unet-based architectures with transformers and further explored different condition injection mechanisms for conditional generation.
\item \textbf{SiT}~\cite{sit} systematically investigated the connections between discrete diffusion to continuous flow matching and developed practical training configurations for achieving strong synthesis performance.
\item \textbf{REPA}~\cite{repa} connected diffusion training dynamics and representation learning, revealing that pretrained external guidance could facilitate the representation learning of diffusion transformers.
\item \textbf{REG}~\cite{wu2025representation} further advanced REPA with a decoupled representation alignment technique, which entangled image latents and class tokens to improve the conditional discrimination capability.
\item \textbf{E2E-REPA}~\cite{leng2025repa} unlocked an end-to-end training paradigm for joint tuning both the VAE and diffusion models throughout the training process, improving the VAE itself and downstream generation performance simultaneously.
\item \textbf{SRA}~\cite{jiang2025no} leveraged representations from later layers with lower noise of the EMA teacher to guide representations of earlier layers with higher noise, enabling a scheme of self-alignment.
\item \textbf{DispLoss}~\cite{wang2025diffuse} introduced a regularized dispersive loss to encourage internal features to spread out in the embedding space, thus facilitating the model to learn informative representations.
\end{itemize}

\subsection{One-step baseline models}
\begin{itemize}
\item \textbf{MeanFlow}~\cite{geng2025mean} introduced average velocity that was defined as the ratio of displacement to a time interval, with displacement given by the time integral of the instantaneous velocity. An intrinsic relation between the average and instantaneous velocities was then derived to guide efficient and effective one-step generative training.
\item \textbf{Shortcut}~\cite{shortcut} enhanced the few-step flow matching by adding a self-consistency loss, designed to learn the relationships between flow behaviors observed at different discrete time points.
\item \textbf{IMM}~\cite{inductive} learned a model that enforces self-consistency among stochastic interpolants evaluated at different points in time.
\item \textbf{iCT}~\cite{ict} leveraged consistency constraints across network outputs at different time steps to ensure that they predict the same endpoints along the trajectory.
\end{itemize}

\subsection{Protein inverse folding baselines}
\begin{itemize}
\item \textbf{ProteinMPNN}~\cite{proteinmpnn} is the foundational model for protein inverse folding, which processes the 3D backbone structure via message passing on a $k$-nearest neighbor graph of C$\alpha$ distances and autoregressively decodes amino acid sequences.
\item \textbf{Multiflow}~\cite{multiflow_campbell2024generative} introduced a discrete diffusion objective for protein sequence generation conditioned on backbone structures, providing the training framework we adopt.
\item \textbf{REED}~\cite{reed} extended the REPA representation alignment paradigm to scientific domains. For protein inverse folding, AlphaFold3~\cite{af3} is used to generate auxiliary structures from target sequences, and representations are extracted at multiple levels: latents from the AlphaFold3 diffusion head as \textit{structure representations}, single embeddings from the Pairformer module as \textit{amino acid representations}, and pair embeddings from the Pairformer module as \textit{pairwise interaction representations}.
\end{itemize}

\subsection{3D molecule generation baselines}
\begin{itemize}
\item \textbf{SemlaFlow}~\cite{irwin2025semlaflow} is our base model for 3D molecule generation, which employs an $E(3)$-equivariant backbone and the ODE flow matching framework to learn continuous trajectories from noise to molecular conformations. The equivariant architecture ensures the generated molecules are invariant to rotations and translations.
\item \textbf{REED}~\cite{reed} applied representation alignment to 3D molecule generation by using Unimol~\cite{zhou2023unimol} finetuned on GEOM-Drug as the pretrained encoder. Features extracted from the final layer of Unimol are aligned with intermediate representations of the SemlaFlow backbone via cosine similarity, using the same three-layer MLP projector with SiLU activations as in image generation.
\end{itemize}

\section{Detailed Results of Our Analysis}
\label{sec:supp:analysis_details}

\begin{figure*}[t]
    \vskip -0.2in
    \begin{center}  \centerline{\includegraphics[width=\linewidth]{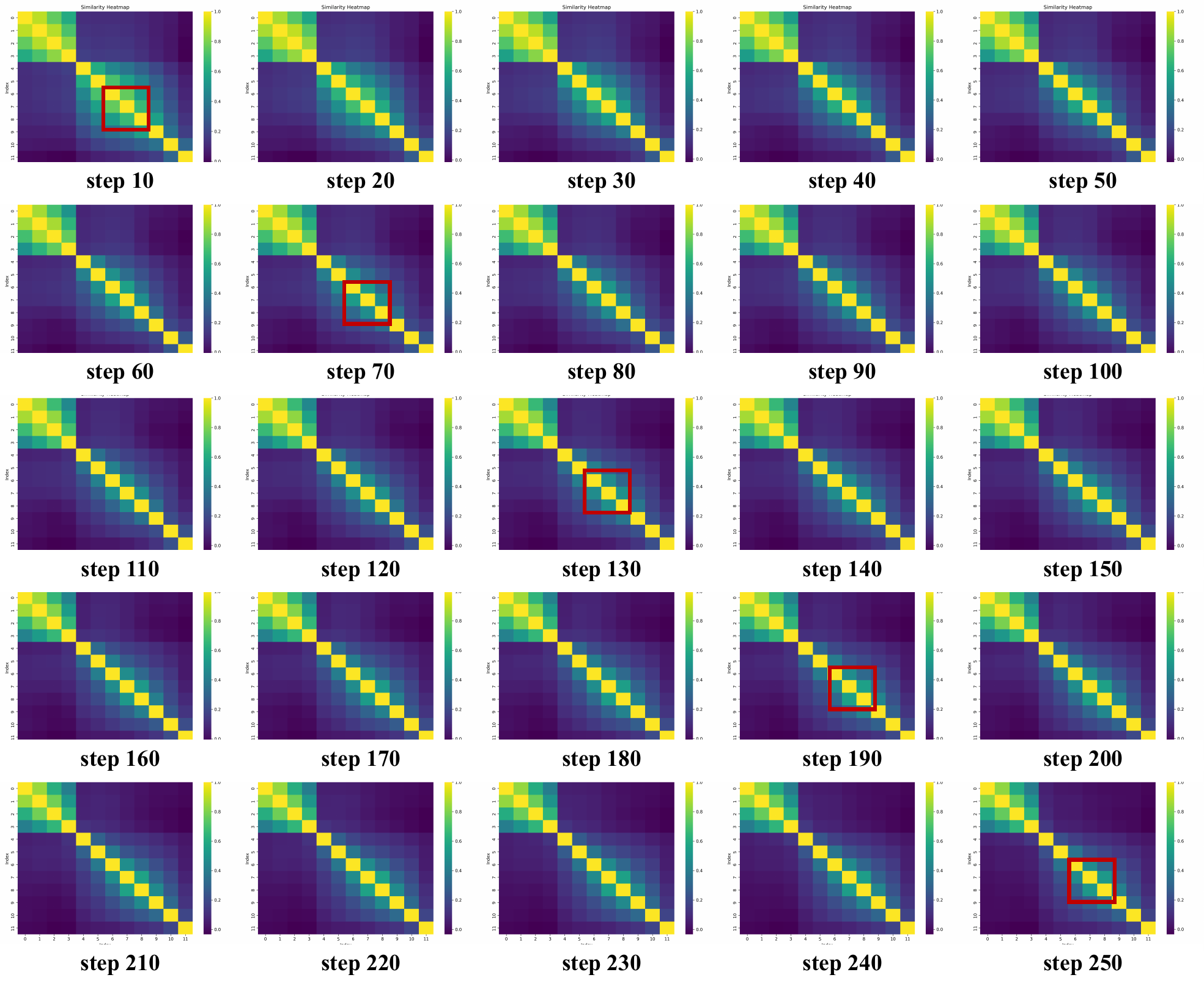}}
    \vskip -0.1in
    \caption{\textbf{CKA representation similarities across different timesteps.}
    The representational discrepancies across different timesteps show similar correlations.
    }
    \label{fig:supp_cka_different_time}
    \end{center}
\end{figure*}

\paragraph{CKA similarity across various timesteps}
In the main paper Fig.~2, we present the CKA similarity heatmaps between representations of different blocks at the final denoising timestep.
To investigate the difference of block representations across different timesteps, we calculate their representational discrepancies of different timesteps in~\cref{fig:supp_cka_different_time}.
We could observe that the representation similarities between different blocks across different timesteps show a very similar pattern.
That is, the representational discrepancy across diffusion transformer blocks originates from the internal representation instead of different denoising timesteps.
Moreover, we can see that as the inference step increases, the representational discrepancy between different blocks at different timesteps tends to slightly increase as well.
Such observation is reasonable because the noisy hidden states become less noisy throughout the sampling process.

\begin{table*}[t]
    \centering
    \small
    \caption{\textbf{Variation in alignment depth on ImageNet 256$\times$256 with different CFG scales}.
    CFG=1.0 means no classifier-free guidance is applied.
    Our proposed method brings consistent performance gains across all model scales when applied to REPA with different alignment depths and evaluated with different CFG scales.
    }
    \vskip -0.1in
    \setlength{\tabcolsep}{1.8mm}{
    \begin{tabular}{lccccccccc}
        \toprule
        \textbf{Model} & \textbf{Iter}.  & \textbf{Encoder}. & \textbf{Align Depth}. &\textbf{CFG}. & \textbf{FID}$_\downarrow$ & \textbf{sFID}$_\downarrow$ & \textbf{IS}$_\uparrow$ & \textbf{Prec.}$_\uparrow$ & \textbf{Rec.}$_\uparrow$ \\
        \midrule
        \textbf{REPA-B} \\
           & 400k & DINOv2-B  & 5 & 1.0  & 22.99   & 6.70  & 64.73    & 0.59    & 0.65 \\
           & 400k & DINOv2-B  & 5 & 1.35 & 12.47   & 5.95  & 107.38   & 0.67    & 0.61 \\
         +\textbf{Ours} \\
           & 400k & DINOv2-B  & 5 & 1.0  & 17.22   & 6.52  & 80.04    & 0.63    & 0.66 \\
           & 400k & DINOv2-B  & 5 & 1.35 & 8.33    & 5.84  & 134.16   & 0.70    & 0.63 \\
        \hdashline
        \textbf{REPA-B} \\
           & 400k & DINOv2-B  & 8 & 1.0  & 27.94   & 7.19  & 54.32   & 0.56    & 0.64 \\
           & 400k & DINOv2-B  & 8 & 1.35 & 16.46   & 6.38  & 90.97   & 0.64    & 0.62 \\
         +\textbf{Ours} \\
           & 400k & DINOv2-B  & 8 & 1.0  & 23.24   & 6.83  & 62.57   & 0.59    & 0.65 \\
           & 400k & DINOv2-B  & 8 & 1.35 & 12.92   & 6.08  & 104.46  & 0.66    & 0.63 \\
        \midrule
        \textbf{REPA-L} \\
           & 400k & DINOv2-B  & 5 & 1.0  & 12.02   & 6.77  & 40.09   & 0.51    & 0.63 \\
           & 400k & DINOv2-B  & 5 & 1.35 & 5.14    & 4.74  & 157.71  & 0.75    & 0.61 \\
         +\textbf{Ours} \\
           & 400k & DINOv2-B  & 5 & 1.0  & 10.03   & 5.49  & 107.31   & 0.68    & 0.64 \\
           & 400k & DINOv2-B  & 5 & 1.35 & 4.18    & 4.61  & 172.26   & 0.76    & 0.61 \\
        \hdashline
        \textbf{REPA-L} \\
           & 400k & DINOv2-B  & 8 & 1.0  & 9.57    & 5.34  & 113.42   & 0.69    & 0.66 \\
           & 400k & DINOv2-B  & 8 & 1.35 & 3.86    & 4.82  & 183.18   & 0.75    & 0.63 \\
         +\textbf{Ours} \\
           & 400k & DINOv2-B  & 8 & 1.0  & 8.42    & 5.32  & 124.21   & 0.69    & 0.68 \\
           & 400k & DINOv2-B  & 8 & 1.35 & 3.39    & 4.80  & 196.08   & 0.76    & 0.63 \\
        \midrule
        \textbf{REPA-XL} \\
           & 400k & DINOv2-B  & 5 & 1.0  & 8.27    & 5.19  & 123.85   & 0.69    & 0.66 \\
           & 400k & DINOv2-B  & 5 & 1.35 & 3.33    & 4.73  & 196.52   & 0.75    & 0.64 \\
         +\textbf{Ours} \\
           & 400k & DINOv2-B  & 5 & 1.0  & 8.19    & 5.03  & 128.12   & 0.70    & 0.66 \\
           & 400k & DINOv2-B  & 5 & 1.35 & 3.17    & 4.71  & 198.30   & 0.77    & 0.62 \\
        \hdashline
        \textbf{REPA-XL} \\
           & 400k & DINOv2-B  & 8 & 1.0  & 8.73    & 5.21  & 118.68   & 0.69    & 0.65 \\
           & 400k & DINOv2-B  & 8 & 1.35 & 3.50    & 4.72  & 188.96   & 0.76    & 0.63 \\
         +\textbf{Ours} \\
           & 400k & DINOv2-B  & 8 & 1.0  & 8.06    & 5.01  & 123.46   & 0.70    & 0.66 \\
           & 400k & DINOv2-B  & 8 & 1.35 & 3.16    & 5.60  & 194.36   & 0.77    & 0.62 \\
        \bottomrule
    \end{tabular}
    }
    \vskip -0.15in
    \label{tab:supp_different_depth_cfg}
\end{table*}

\section{More Quantitative Results}
\label{sec:supp:more_quantitative}

\begin{table}[t]
    \centering
    \small
    \vskip -0.1in
    \caption{\textbf{Variation in model-scale on ImageNet 512$\times$512 with CFG=1.35}.
    Our proposed method brings consistent performance gains across all model-scales when applied to both SiT and REPA.
    }
    \vskip -0.1in
    \setlength{\tabcolsep}{1.8mm}{
    \begin{tabular}{lcccccc}
        \toprule
        \textbf{Model}  & \textbf{Iter}. & \textbf{FID}$_\downarrow$ & \textbf{sFID}$_\downarrow$ & \textbf{IS}$_\uparrow$ & \textbf{Prec.}$_\uparrow$ & \textbf{Rec.}$_\uparrow$ \\
        \midrule
    SiT-B             & 400k  & 32.77          & 6.95          & 50.85          & 0.67          & \textbf{0.62}           \\
    \textbf{+ (Ours)} & 400k  & \textbf{23.96} & \textbf{6.45} & \textbf{63.07} & \textbf{0.73} & \textbf{0.62}  \\
    \hdashline
    REPA-B            & 400k  & 21.27          & 7.24          & 78.25          & 0.73          & 0.62  \\
    \textbf{+ (Ours)} & 400k  & \textbf{16.44} & \textbf{7.30} & \textbf{92.11} & \textbf{0.75} & \textbf{0.63}           \\
    \midrule
    SiT-L             & 400k  & 14.85          & 5.41          & 91.49          & 0.78          & \textbf{0.60}  \\
    \textbf{+ (Ours)} & 400k  & \textbf{12.44} & \textbf{5.40} & \textbf{101.08} & \textbf{0.79} & 0.58           \\
    \hdashline
    REPA-L            & 400k  & 5.57          & \textbf{5.35}          & 158.39         & \textbf{0.80}          & 0.62           \\
    \textbf{+ (Ours)} & 400k  & \textbf{4.66}  & 5.58 & \textbf{173.94} & \textbf{0.80} & \textbf{0.64}  \\
    \midrule
    SiT-XL            & 400k  & 12.50          & \textbf{5.17}          & 102.38          & 0.79          & 0.57  \\
    \textbf{+ (Ours)} & 400k  & \textbf{11.24} & 5.27 & \textbf{107.73} & \textbf{0.80} & \textbf{0.58}  \\
    \hdashline
    REPA-XL           & 400k  & 4.30           & \textbf{5.09}          & 174.70         & \textbf{0.81} & \textbf{0.62}  \\
    \textbf{+ (Ours)} & 400k  & \textbf{3.98}  & 5.48 & \textbf{184.73}& 0.80       & \textbf{0.62}           \\
    \bottomrule
    \end{tabular}
    }
    \vskip -0.15in
    \label{tab:supp_different_size_512_cfg}
\end{table}

\paragraph{Comparison results across different model scales with different CFG scales}
We mainly present comparison results without using classifier-free guidance (CFG)~\cite{nichol2021improved} in the main paper.
In this part, we present comparison results across different model scales with CFG enabled to further investigate its impact and our performance.
Specifically, we conduct experiments on ImageNet 256x256 on REPA using the DINOv2-B encoder for 400k training iterations across different model sizes (REPA-B, REPA-L, and REPA-XL).
We systematically evaluated the performance with different CFG scales (1.0, representing no classifier-free guidance, and 1.35).
\cref{tab:supp_different_depth_cfg} presents a detailed analysis of the impact of Classifier-Free Guidance (CFG) scale on the performance of our proposed method when applied to REPA models of varying sizes (REPA-B, REPA-L, and REPA-XL).

First, the results consistently demonstrate that increasing the CFG scale from 1.0 to 1.35 leads to significant improvements in image quality and diversity across all model scales.
This is evidenced by the substantial increase in IS scores and the decrease in FID scores observed across all REPA model sizes when CFG is enabled.
Second, our proposed method also gains consistent performance improvement across different model scales when CFG is enabled.
For instance, our model advances the FID score of REPA-B from 12.47 to 8.33 and IS score from 107.38 to 134.16 with CFG=1.35, attaining a $>$32\% performance improvement on FID.
Similarly, our model advances the FID score of REPA-XL from 3.50 to 3.16 and the IS score from 188.96 to 194.36 with CFG=1.35.
Furthermore, \cref{tab:supp_different_size_512_cfg} presents the comparison results on ImageNet 512$\times$512 with CFG=1.35.
Across all model scales, our method consistently improves the FID and sFID scores when CFG is used, indicating enhanced image quality and fidelity.
For example, when applied to SiT-B, our method reduces the FID from 32.77 to 23.96 and the sFID from 6.95 to 6.45. Similarly, for REPA-B, the FID decreases from 21.27 to 16.44.
Together with the results that were tested without using CFG, these results demonstrate the scalability and effectiveness of our method to higher resolutions and different model sizes.

\begin{table}[t]
    \centering
    \small
    \caption{\textbf{Ablation analysis on different components with CFG=1.35}.}
    \label{tab:ablation_components_supp}
    \vskip -0.1in
    \setlength{\tabcolsep}{2mm}{
    \begin{tabular}{lccccc}
        \toprule
         \textbf{Component} & \textbf{FID}$_\downarrow$ & \textbf{sFID}$_\downarrow$ & \textbf{IS}$_\uparrow$ & \textbf{Prec.}$_\uparrow$ & \textbf{Rec.}$_\uparrow$ \\
         \midrule
        SiT-B Baseline               & 23.28         & 6.00          & 65.23         & 0.61          & 0.60         \\
        SiT-B + ${\texttt{full}}$    & 16.21         & 5.45          & 84.00         & 0.66          & 0.60         \\
        w/o ${\texttt{diversity}}$   & 20.07         & 5.72          & 73.65         & 0.63          & 0.60         \\
        w/o ${\texttt{residual}}$    & 20.76         & 5.77          & 69.23         & 0.61          & 0.61         \\ \midrule
        REPA                         & 12.47         & 5.85          & 107.38        & 0.61          & 0.62         \\
        REPA-B + ${\texttt{full}}$   & 8.34          & 5.64          & 134.16        & 0.70          & 0.63        \\
        w/o ${\texttt{diversity}}$   & 10.75         & 5.75          & 115.42        & 0.68          & 0.62         \\
        w/o ${\texttt{residual}}$    & 11.02         & 5.77          & 112.93        & 0.64          & 0.62         \\
        \bottomrule
    \end{tabular}
    }
\end{table}

\paragraph{Quantitative results of applying our method on REPA with alignment on different blocks}
\cref{tab:supp_different_depth_cfg} also presents insights into the impact of alignment depth on the performance of our proposed method.
We evaluated the models with alignment depths of 5 and 8, while keeping other parameters constant.
Despite these variations, our method consistently improves upon the baseline REPA models when performing alignment on different blocks, with or without CFG.
For example, REPA-XL with our method and an alignment depth of 5 achieves an FID score of 3.17 at CFG 1.35, compared to 3.33 for the baseline.
Similarly, the IS score improves from 196.52 to 198.30.
This consistent trend of improvement, regardless of alignment depth, demonstrates the effectiveness of our approach in enhancing image generation.
The consistent improvements observed across different alignment depths and model sizes further demonstrate the robustness and generalizability of our approach.

\section{More Ablation and Analysis Results}
\label{sec:supp:more_ablation}

\paragraph{Ablation on different loss variants}
Here, we further perform ablation on each loss component of the proposed diversity loss on SiT-B baseline.
The results in Tab.~\ref{tab:ablation_loss_sit} show the effectiveness of each loss, consistent with the findings of REPA results in the main paper.
Specifically, using all components of the diversity loss achieves the best performance and removing any single loss component degrades performance.

\begin{table}[t]
    \centering
    \small
    \caption{\textbf{Ablation analysis on different loss variants on SiT-B baseline.}}
    \label{tab:ablation_loss_sit}
    \vskip -0.1in
    \setlength{\tabcolsep}{2mm}{
    \begin{tabular}{lccccc}
        \toprule
         \textbf{Component} & \textbf{FID}$_\downarrow$ & \textbf{sFID}$_\downarrow$ & \textbf{IS}$_\uparrow$ & \textbf{Prec.}$_\uparrow$ & \textbf{Rec.}$_\uparrow$ \\
         \midrule
     SiT-B + ${\texttt{full}}$                        & 28.05   & 6.04  & 50.66  & 0.57  & 0.63 \\
     ${\texttt{only $\mathcal{L}_{\text{orth}}$}}$  & 31.32   & 6.45  & 47.09  & 0.56  & 0.63 \\
     ${\texttt{only $\mathcal{L}_{\text{MI}}$}}$    & 29.97   & 6.21  & 48.23  & 0.57  & 0.63 \\
     ${\texttt{only $\mathcal{L}_{\text{div}}$}}$   & 36.12   & 6.64  & 45.04  & 0.55  & 0.62 \\
        \bottomrule
    \end{tabular}
    }
\end{table}

\paragraph{Ablation on designed components with CFG.}
\cref{tab:ablation_components_supp} presents the ablative results on the designed components of our \ourmethod{} with CFG.
We could see that applying the CFG consistently improves the overall scores.
Similar to the results in the main paper, the results clearly demonstrate the importance of both the representation diversity loss and the long residual connections for optimal performance.
Removing the diversity loss (w/o diversity) worsens the FID scores for both SiT-B (from 23.28 to 20.07) and REPA-B (from 12.47 to 10.75).
Similarly, removing the long residual connections (w/o residual) also noticeably increases FID for both baseline models.
Despite some performance degradation, we can observe that applying any of our proposed techniques to the baseline methods, \emph{i.e.,} REPA-B and SiT-B, brings substantial performance improvements.
For instance, with only long residual connections (w/o diversity), we achieve an FID of 20.07 on SiT-B and an FID of 10.75 on REPA-B, which are better than the original baseline results (23.28 for SiT-B and 12.47 for REPA-B).
Similar conclusions could be observed from the results of only diversity loss (w/o residual) as well.
These results confirm that both components of \ourmethod{} play a crucial role in promoting diverse representation learning and improving the performance.

\paragraph{Effect of diversity loss variant with CFG.}
\cref{tab:ablation_loss_supp} presents an ablation analysis on different loss variants with CFG=1.35.
Similar to the previous results, the table demonstrates the importance of each loss component for optimal performance. Removing any of the loss components, namely $\mathcal{L}_{orth}$, $\mathcal{L}_{MI}$, or $\mathcal{L}_{div}$, degrades the FID score compared to the REPA-B + full configuration (8.34).
While using only $\mathcal{L}_{orth}$ results in an FID of 10.98, using only $\mathcal{L}_{MI}$ gives an FID of 10.78, and using only $\mathcal{L}_{div}$ improves the FID to 8.59.
These results confirm that each loss component plays a role in improving the model's performance, which is also reflected by the better results compared with the REPA baseline when each loss is used in isolation.

\paragraph{Combining with existing methods for further improvement with CFG.}
\cref{tab:compatibility_supp} further explores the effect of combining our method with existing approaches, specifically DispLoss~\cite{wang2025diffuse} and SRA~\cite{jiang2025no}, on the SiT-B baseline with CFG=1.35.
Adding our method to the SiT-B baseline improves the FID from 23.28 to 16.21.
Further combining with DispLoss results in an even lower FID of 13.73.
This demonstrates that our method is complementary to existing techniques and can be combined with them to achieve further improvements in image generation quality.
Note that SRA and DispLoss require no additional external models for representation alignment, and combining our proposed method with them achieves a better performance than that of REPA, which needs pretrained models as guidance, demonstrating the potential for representation learning through internal mechanisms.

\begin{table}[t]
    \centering
    \small
    \caption{\textbf{Ablation analysis on different loss variants with CFG=1.35.}}
    \label{tab:ablation_loss_supp}
    \vskip -0.1in
    \setlength{\tabcolsep}{2mm}{
    \begin{tabular}{lccccc}
    \toprule
    \textbf{Component} & \textbf{FID}$_\downarrow$ & \textbf{sFID}$_\downarrow$ & \textbf{IS}$_\uparrow$ & \textbf{Prec.}$_\uparrow$ & \textbf{Rec.}$_\uparrow$ \\
    \midrule
REPA Baseline                                  & 12.47   & 5.85         & 107.38        & 0.61          & 0.62         \\
REPA-B + ${\texttt{full}}$                     & 8.34    & 5.64         & 134.16        & 0.70          & 0.63         \\
${\texttt{only $\mathcal{L}_{\text{orth}}$}}$  & 10.98   & 5.78         & 115.03        & 0.68          & 0.62         \\
${\texttt{only $\mathcal{L}_{\text{MI}}$}}$    & 10.78   & 5.76         & 115.95        & 0.69          & 0.63         \\
${\texttt{only $\mathcal{L}_{\text{div}}$}}$   & 8.59    & 5.77         & 131.69        & 0.70          & 0.63         \\
    \bottomrule
    \end{tabular}
    }
    \vskip -0.15in
\end{table}

\begin{table}[t]
    \centering
    \small
    \caption{
    \textbf{Combining our method with prior approaches with CFG=1.35.}
    }
    \label{tab:compatibility_supp}
    \vskip -0.1in
    \setlength{\tabcolsep}{1.8mm}{
    \begin{tabular}{lccccc}
        \toprule
         \textbf{Component} & \textbf{FID}$_\downarrow$ & \textbf{sFID}$_\downarrow$ & \textbf{IS}$_\uparrow$ & \textbf{Prec.}$_\uparrow$ & \textbf{Rec.}$_\uparrow$ \\
         \midrule
REPA Baseline                                  & 12.47   & 5.85         & 107.38        & 0.61          & 0.62         \\
\midrule
SiT-B Baseline                                  & 23.28         & 6.00         & 65.23      & 0.61     & 0.60         \\
+ ${\texttt{Ours}}$                             & 16.21         & 5.45         & 84.00      & 0.66     & 0.60         \\
++ ${\texttt{DispLoss}}$~\cite{wang2025diffuse} & 13.73         & 5.76         & 95.31      & 0.68     & 0.60         \\
+++ ${\texttt{SRA}}$~\cite{jiang2025no}         & 11.25         & 5.37         & 108.15     & 0.69     & 0.61         \\
        \bottomrule
    \end{tabular}
    }
    \vskip -0.2in
\end{table}

\section{More Qualitative Results}
\label{sec:supp:more_qualitative_results}

\begin{figure*}[t]
    \centering
    \includegraphics[width=0.82\linewidth]{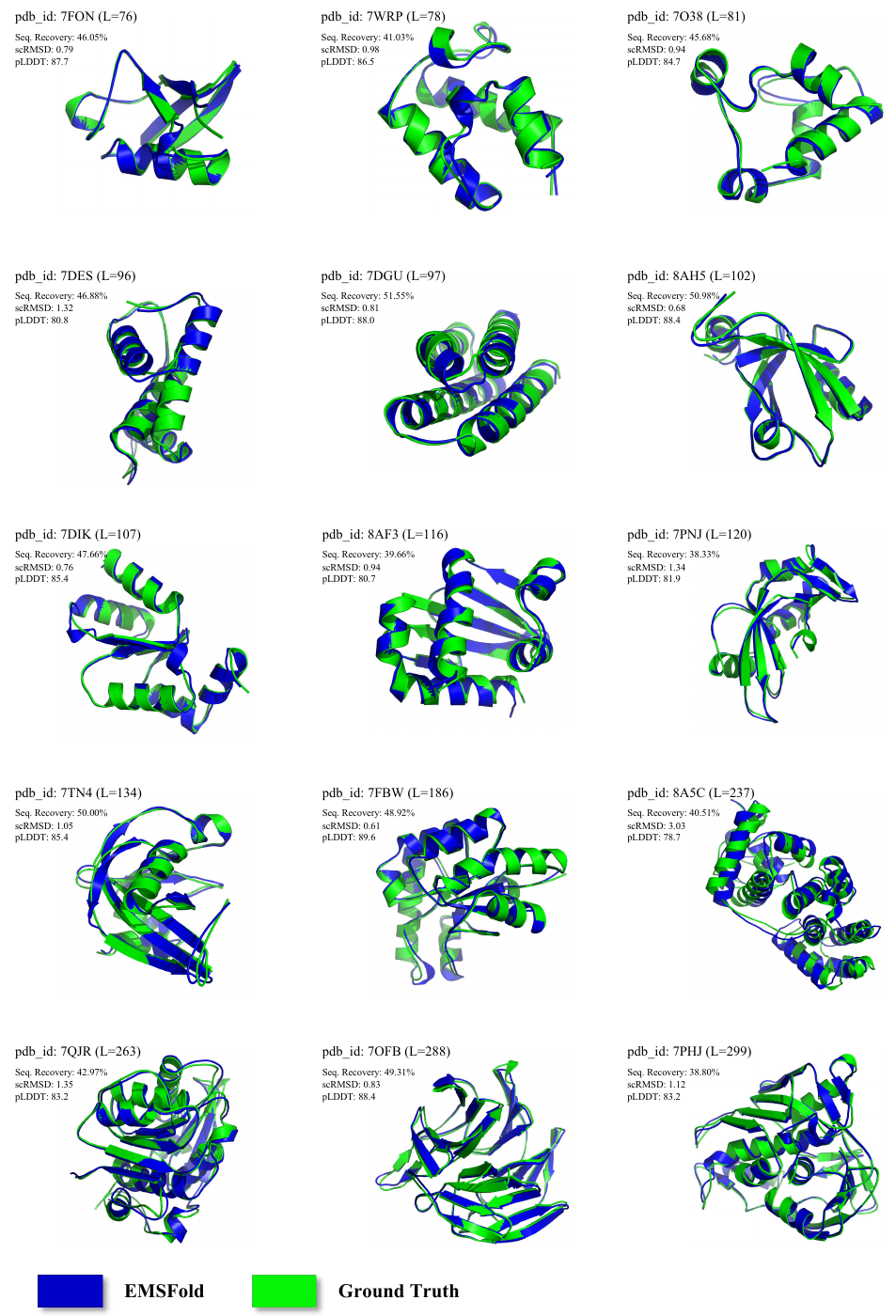}
    \caption{\textbf{Selected samples of our method on protein inverse folding.} Blue color denotes our generated sequence folded by ESMFold~\cite{esmfold} and green color denotes the ground truth structure.}
    \label{fig:supp_more_visualization_protein}
\end{figure*}

\begin{figure*}[t]
    \centering
    \includegraphics[width=0.9\linewidth]{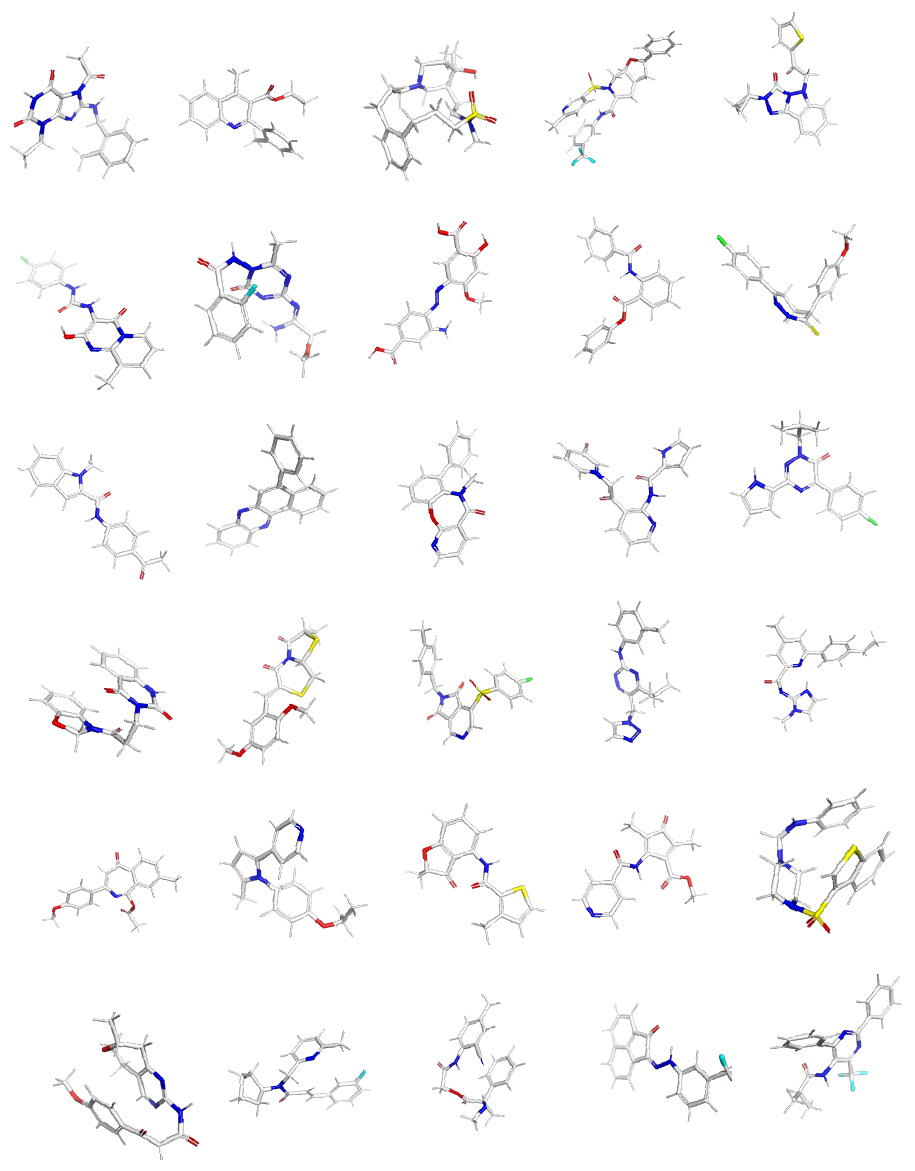}
    \caption{\textbf{Selected samples of our method on molecule generation on GEOM-Drug dataset~\cite{geomdrug}.} Atoms are colored according to element type: C (white), O (red), N (blue), S (yellow), F (cyan), Cl (green), P (orange), and I (purple)}
    \label{fig:supp_more_visualization_molecule_geomdrug}
\end{figure*}

\begin{figure*}[t]
    \centering
    \includegraphics[width=0.9\linewidth]{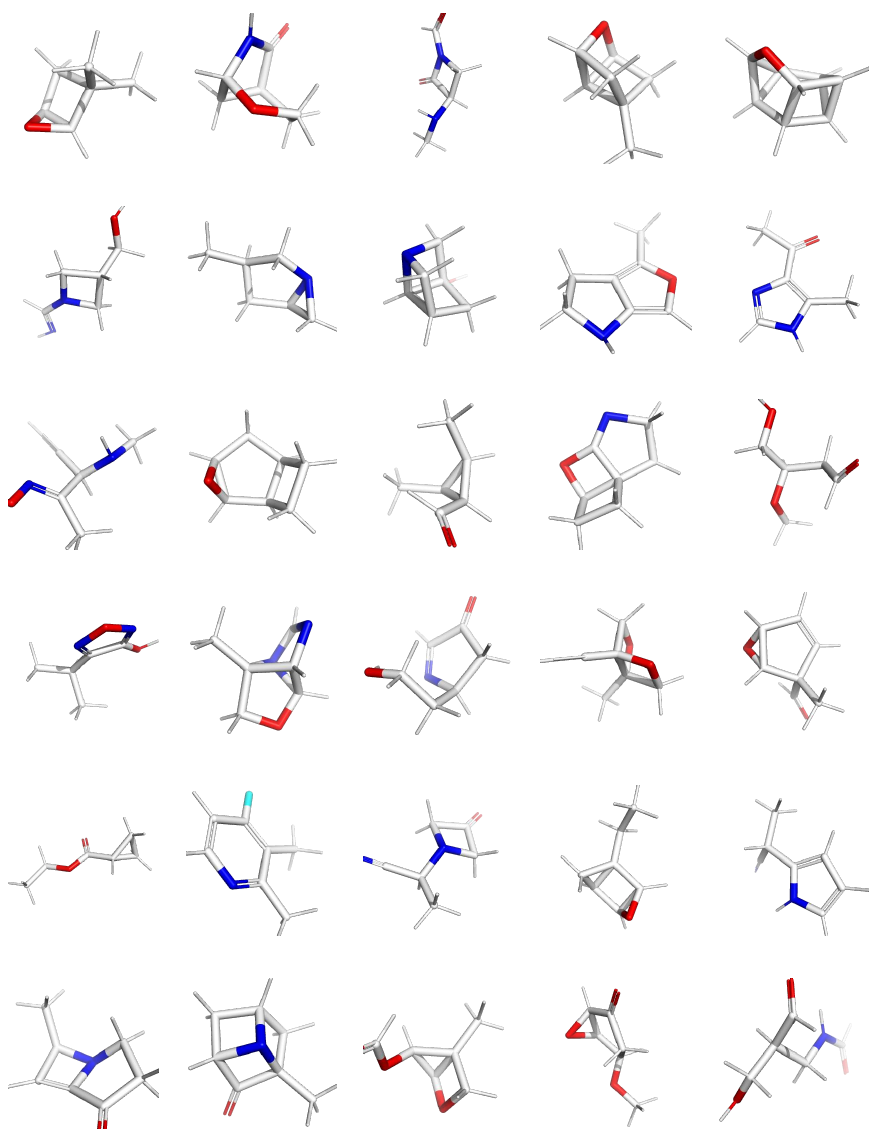}
    \caption{\textbf{Selected samples of our method on molecule generation on QM9 dataset~\cite{qm9_ramakrishnan2014}.} Atoms are colored according to element type: C (white), O (red), N (blue), S (yellow), F (cyan), Cl (green), P (orange), and I (purple)}
    \label{fig:supp_more_visualization_molecule_qm9}
\end{figure*}

\begin{figure*}[t]
    \centering
    \includegraphics[width=0.82\linewidth,page=1]{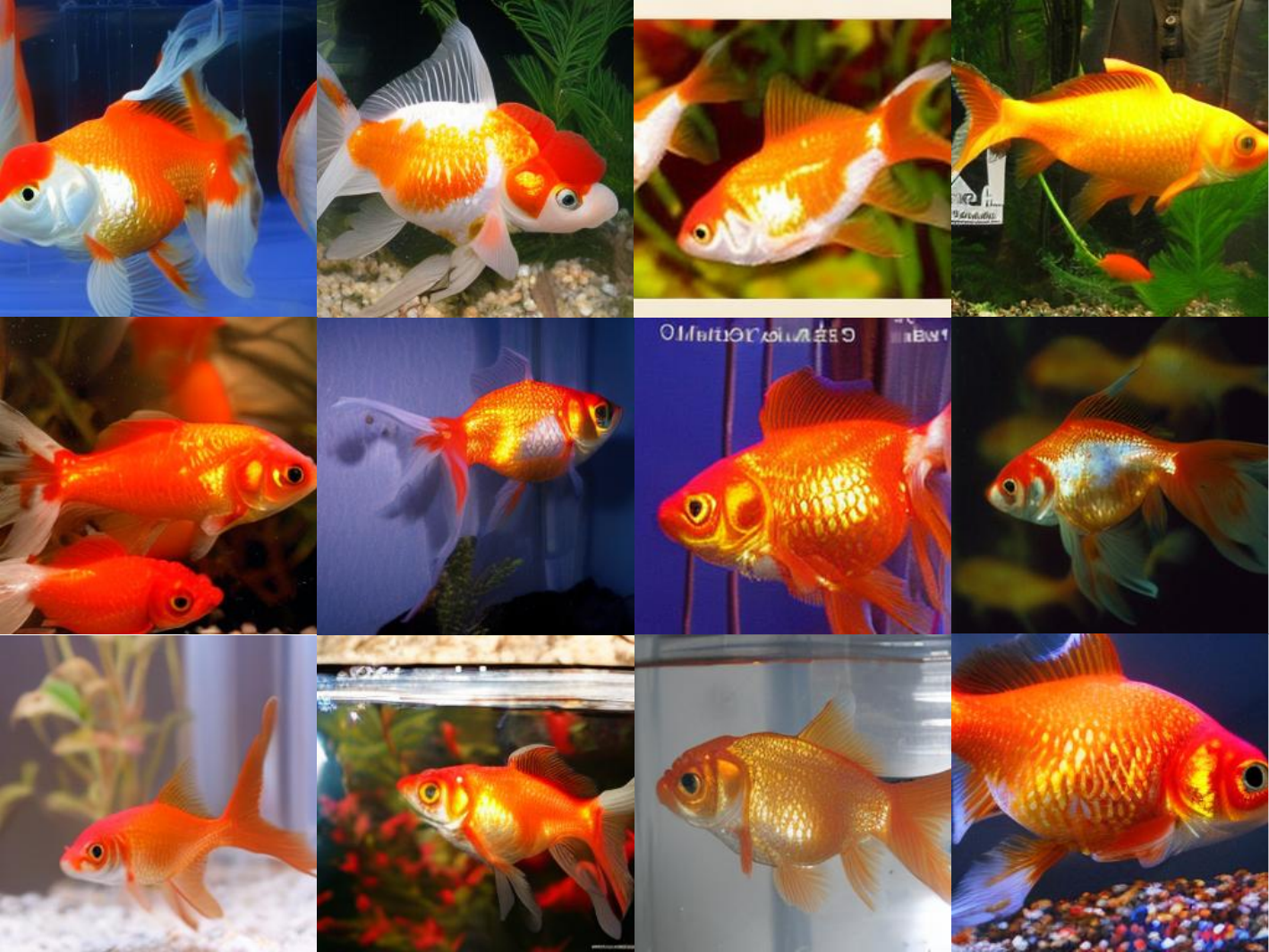}
    \caption{\textbf{Uncurated generation results of our \ourmethod on SIT-XL on ImageNet 256$\times$256.} The class label is ``Goldfish'' (1).}
    \label{fig:supp_more_visualization_1}
\end{figure*}

\begin{figure*}[ht!]
    \centering
    \includegraphics[width=0.82\linewidth,page=2]{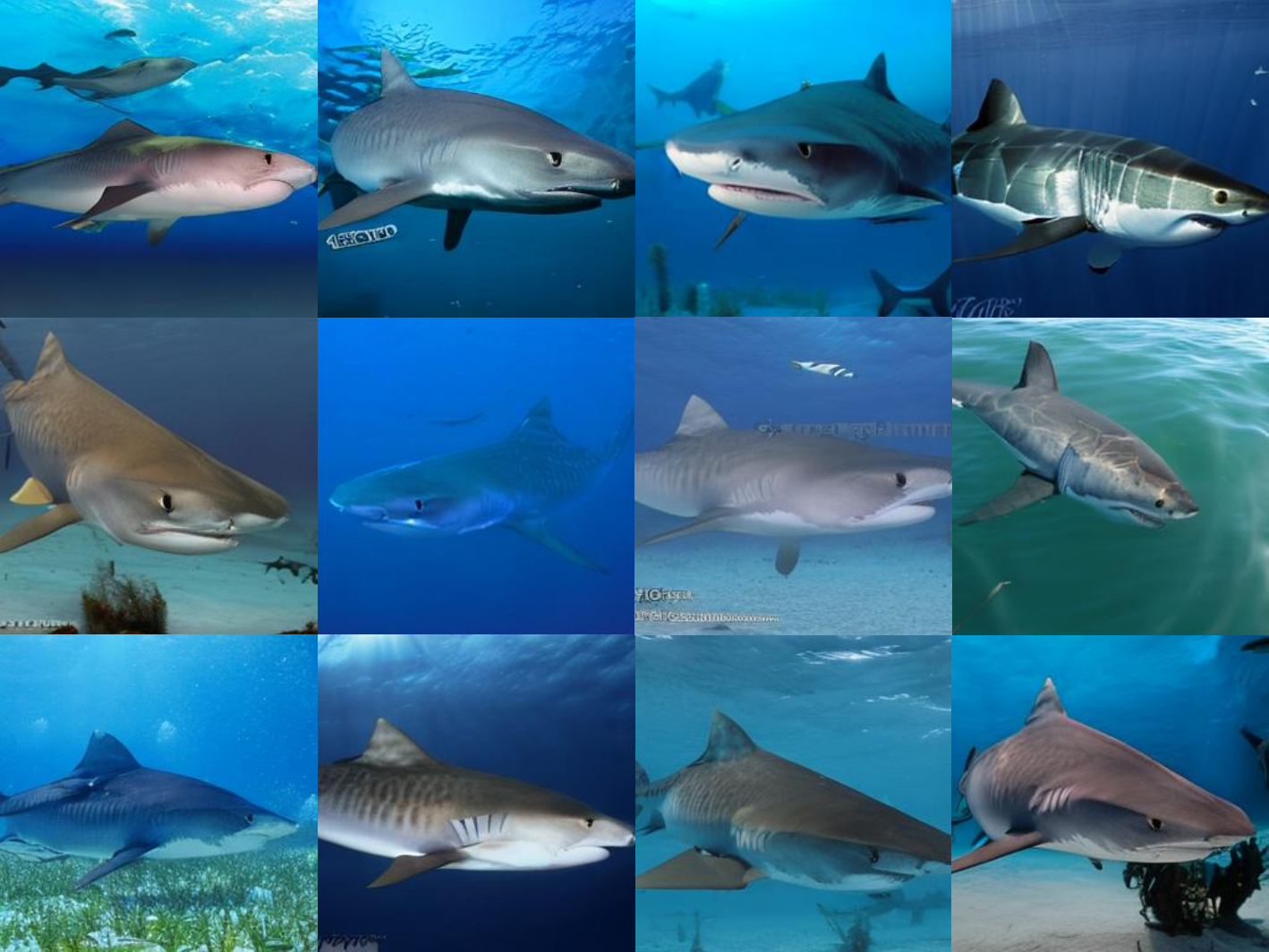}
    \caption{\textbf{Uncurated generation results of our \ourmethod on SIT-XL on ImageNet 256$\times$256.} The class label is ``Tiger shark'' (3).}
    \label{fig:supp_more_visualization_2}
\end{figure*}

\begin{figure*}[ht!]
    \centering
    \includegraphics[width=0.82\linewidth,page=3]{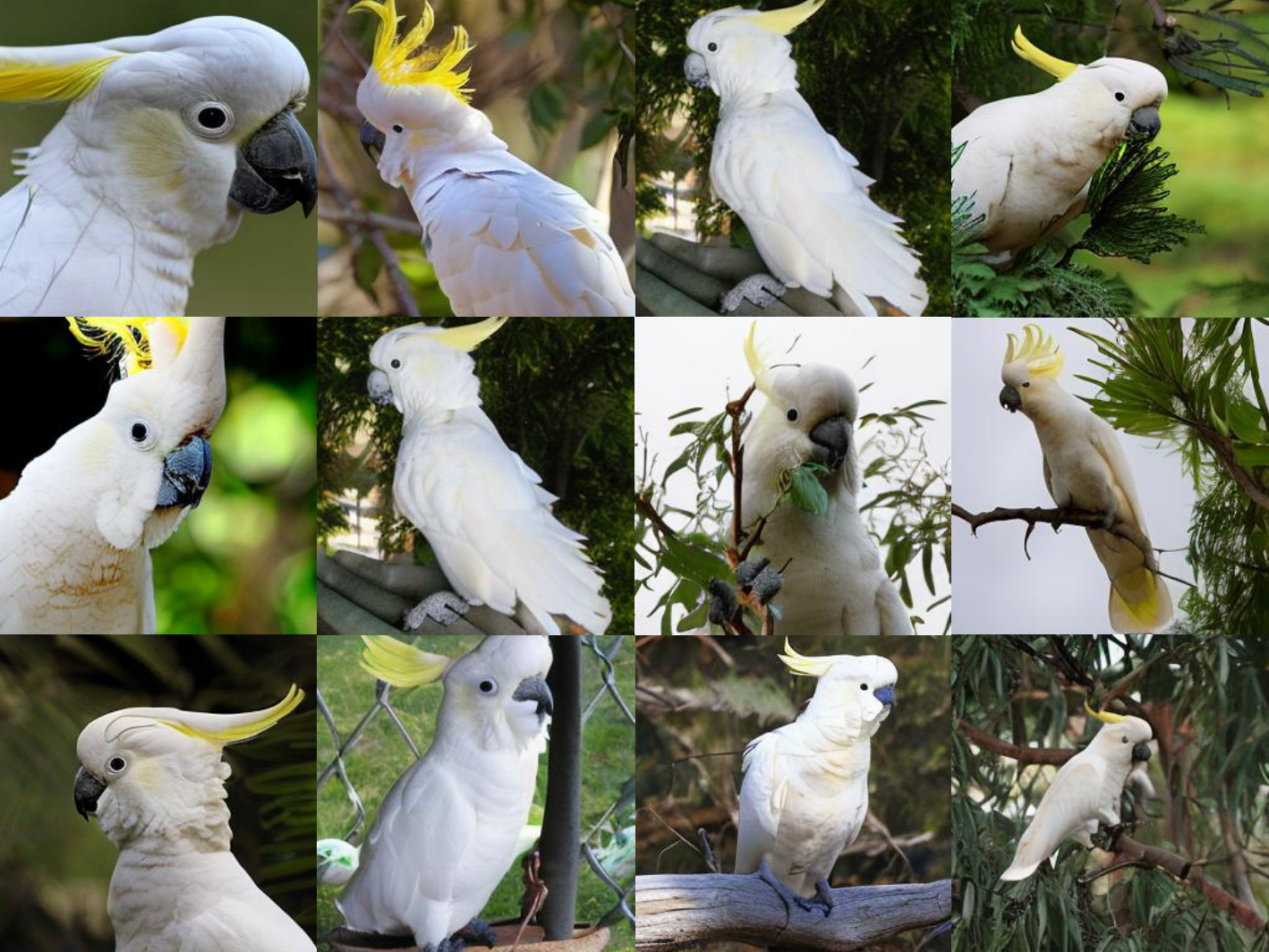}
    \caption{\textbf{Uncurated generation results of our \ourmethod on SIT-XL on ImageNet 256$\times$256.} The class label is ``Kakatoe galerita'' (89).}
    \label{fig:supp_more_visualization_3}
\end{figure*}

\begin{figure*}[ht!]
    \centering
    \includegraphics[width=0.82\linewidth,page=4]{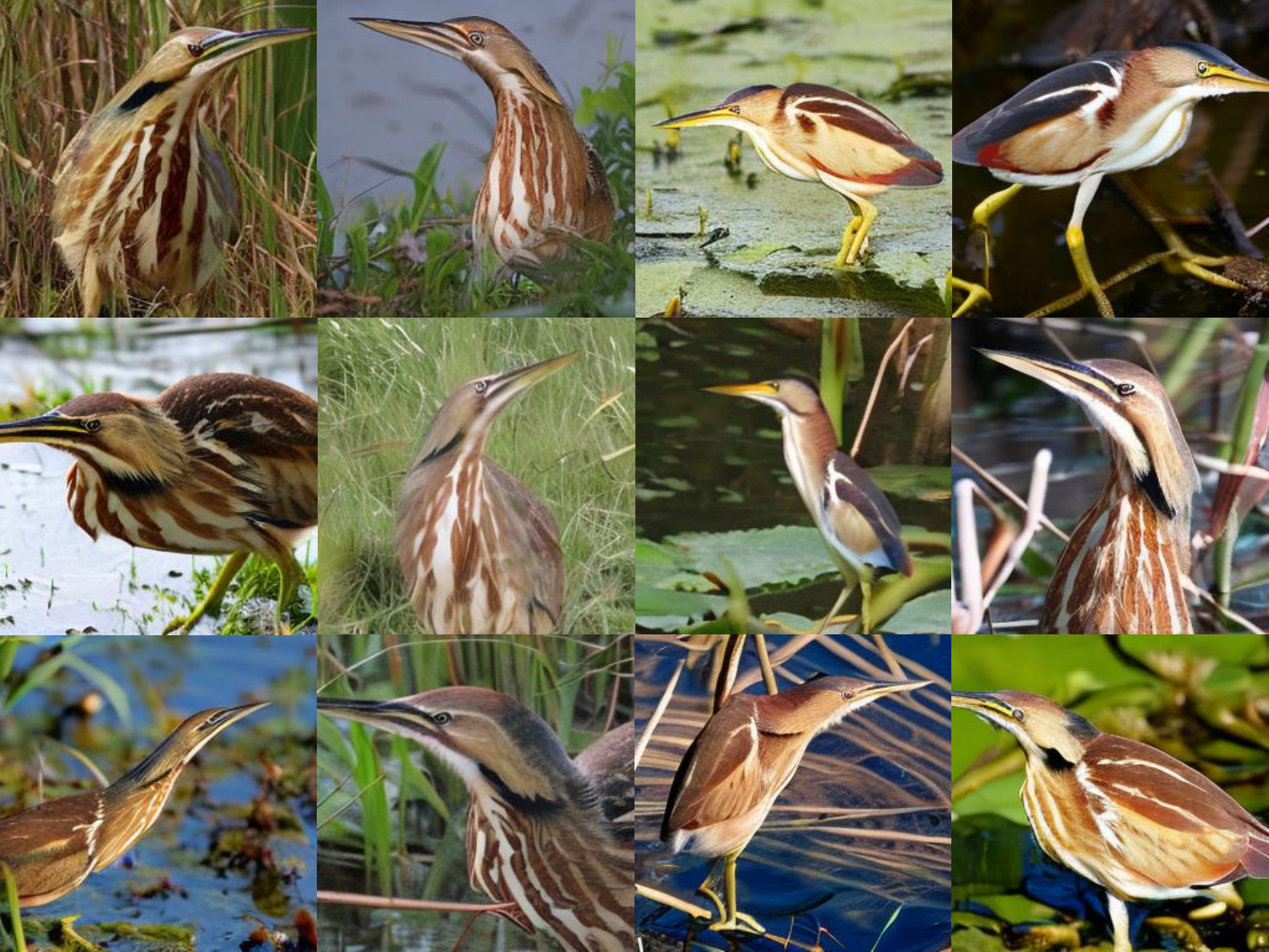}
    \caption{\textbf{Uncurated generation results of our \ourmethod on SIT-XL on ImageNet 256$\times$256.} The class label is ``Bittern'' (133).}
    \label{fig:supp_more_visualization_4}
\end{figure*}

\begin{figure*}[ht!]
    \centering
    \includegraphics[width=0.82\linewidth,page=5]{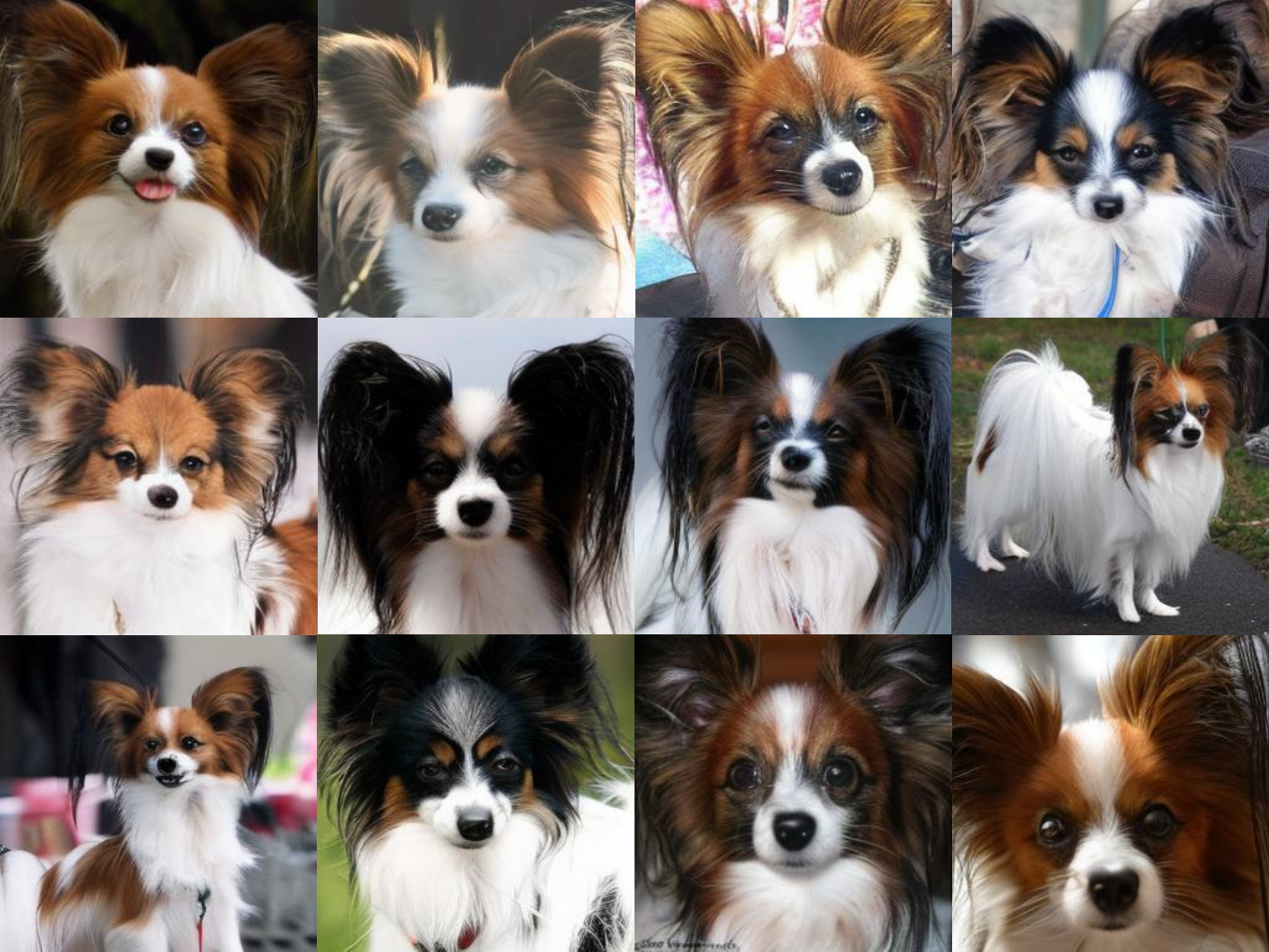}
    \caption{\textbf{Uncurated generation results of our \ourmethod on SIT-XL on ImageNet 256$\times$256.} The class label is ``Papillon'' (157).}
    \label{fig:supp_more_visualization_5}
\end{figure*}

\begin{figure*}[ht!]
    \centering
    \includegraphics[width=0.82\linewidth,page=6]{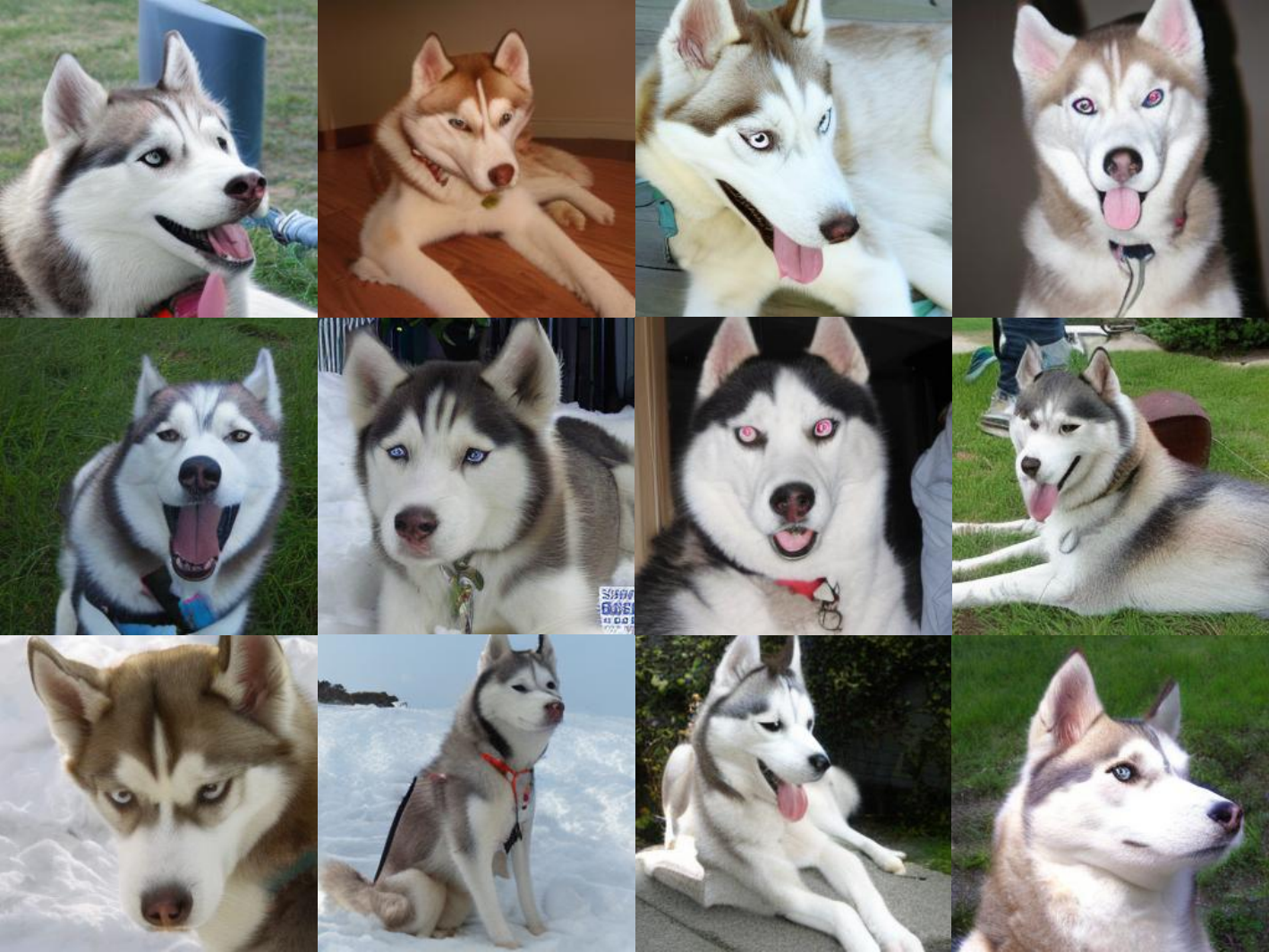}
    \caption{\textbf{Uncurated generation results of our \ourmethod on SIT-XL on ImageNet 256$\times$256.} The class label is ``Siberian husky'' (250).}
    \label{fig:supp_more_visualization_6}
\end{figure*}

\begin{figure*}[ht!]
    \centering
    \includegraphics[width=0.82\linewidth,page=7]{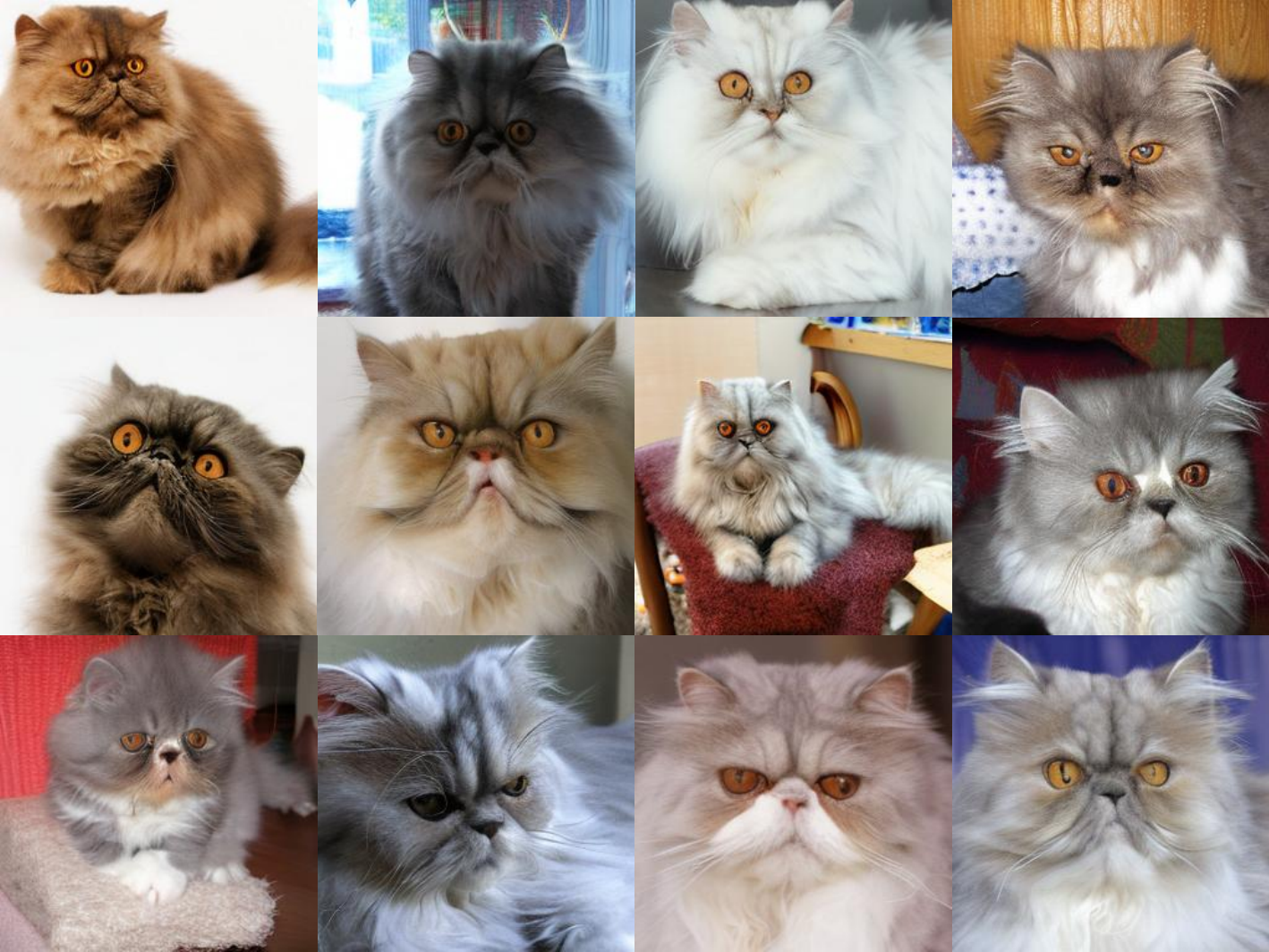}
    \caption{\textbf{Uncurated generation results of our \ourmethod on SIT-XL on ImageNet 256$\times$256.} The class label is ``Persian cat'' (283).}
    \label{fig:supp_more_visualization_7}
\end{figure*}

\begin{figure*}[ht!]
    \centering
    \includegraphics[width=0.82\linewidth,page=8]{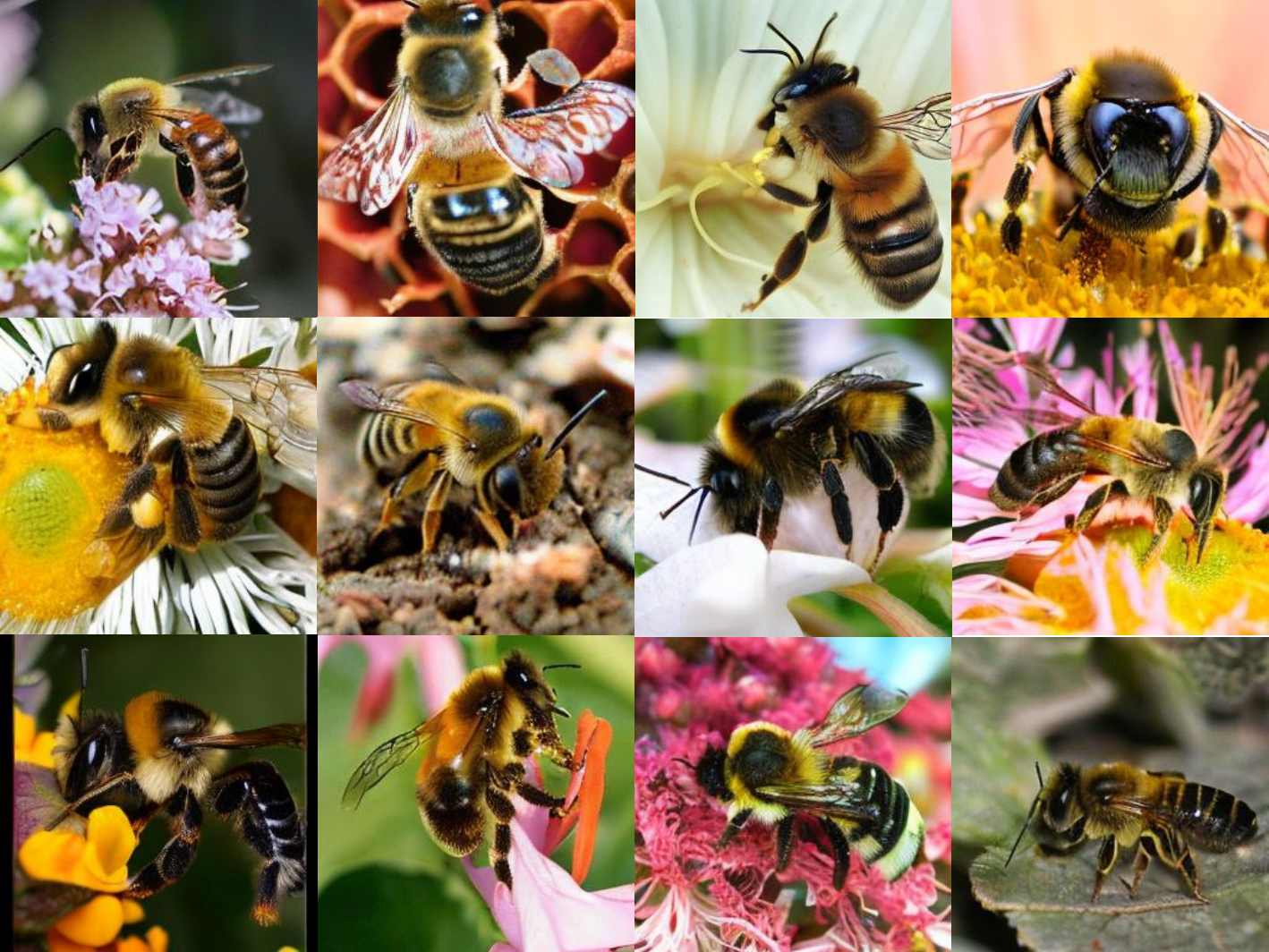}
    \caption{\textbf{Uncurated generation results of our \ourmethod on SIT-XL on ImageNet 256$\times$256.} The class label is ``Bee'' (309).}
    \label{fig:supp_more_visualization_8}
\end{figure*}

\begin{figure*}[ht!]
    \centering
    \includegraphics[width=0.82\linewidth,page=9]{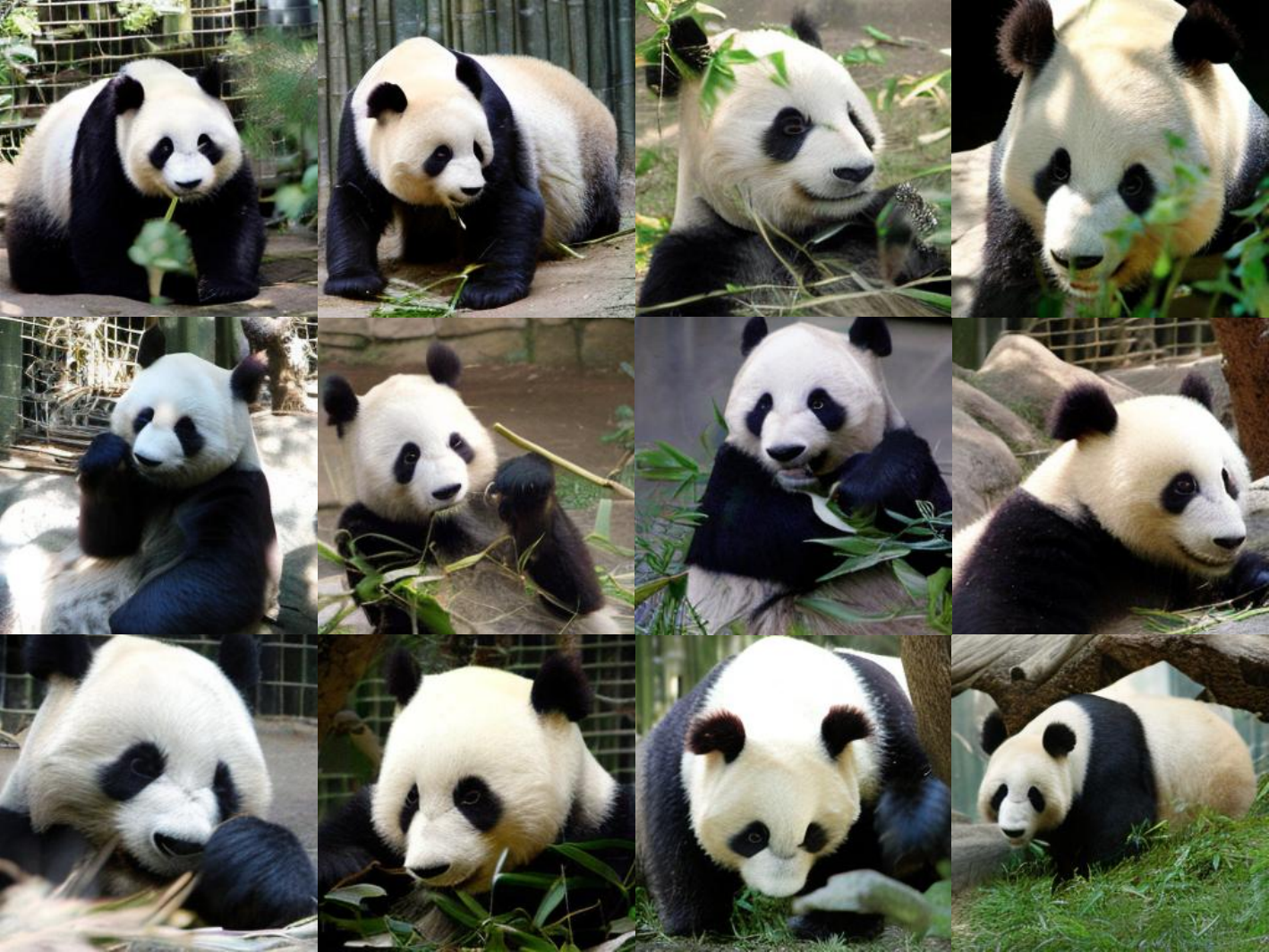}
    \caption{\textbf{Uncurated generation results of our \ourmethod on SIT-XL on ImageNet 256$\times$256.} The class label is ``Giant panda'' (388).}
    \label{fig:supp_more_visualization_9}
\end{figure*}

\begin{figure*}[ht!]
    \centering
    \includegraphics[width=0.82\linewidth,page=10]{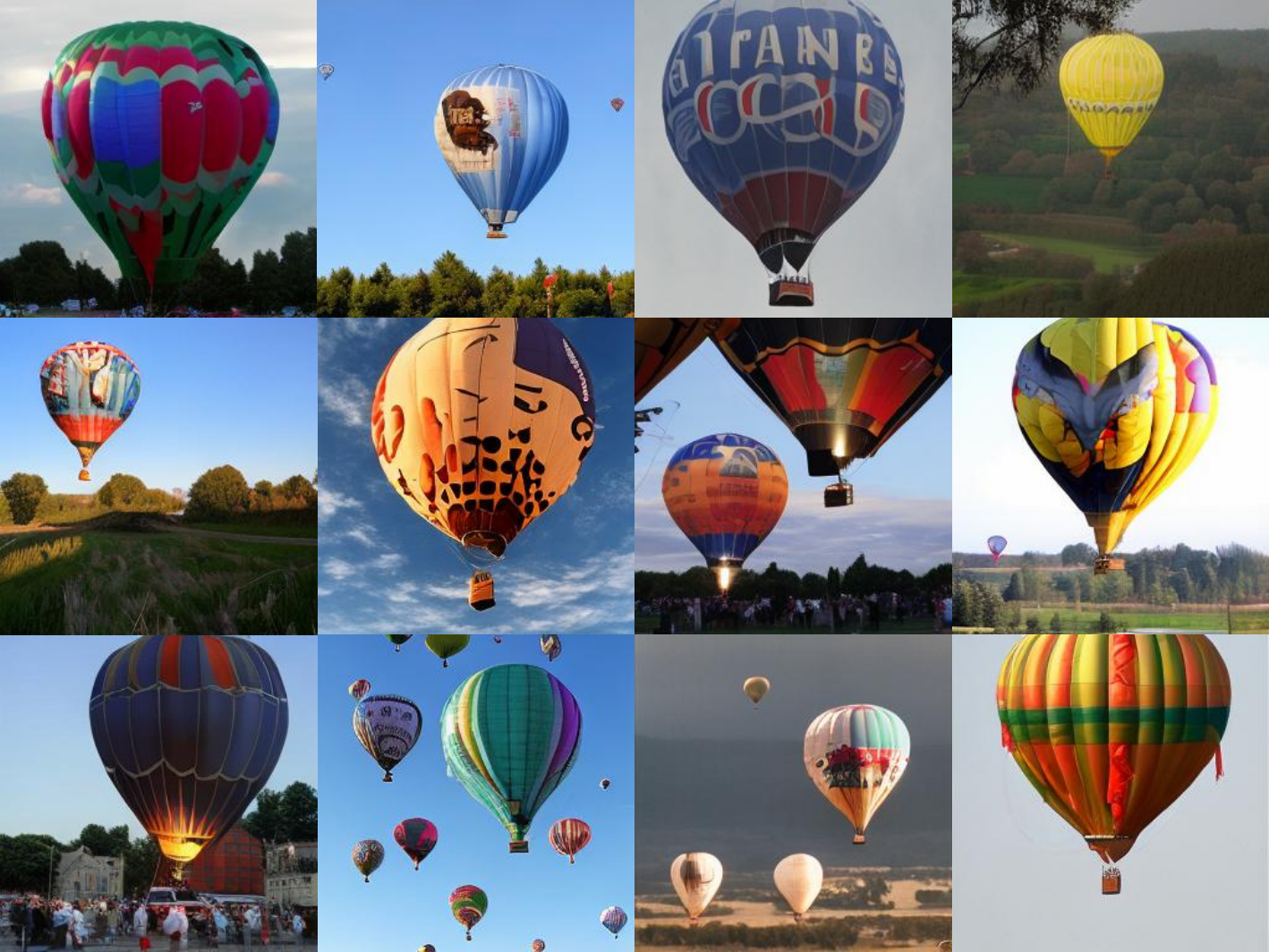}
    \caption{\textbf{Uncurated generation results of our \ourmethod on SIT-XL on ImageNet 256$\times$256.} The class label is ``Ballon'' (417).}
    \label{fig:supp_more_visualization_10}
\end{figure*}

\begin{figure*}[ht!]
    \centering
    \includegraphics[width=0.82\linewidth,page=11]{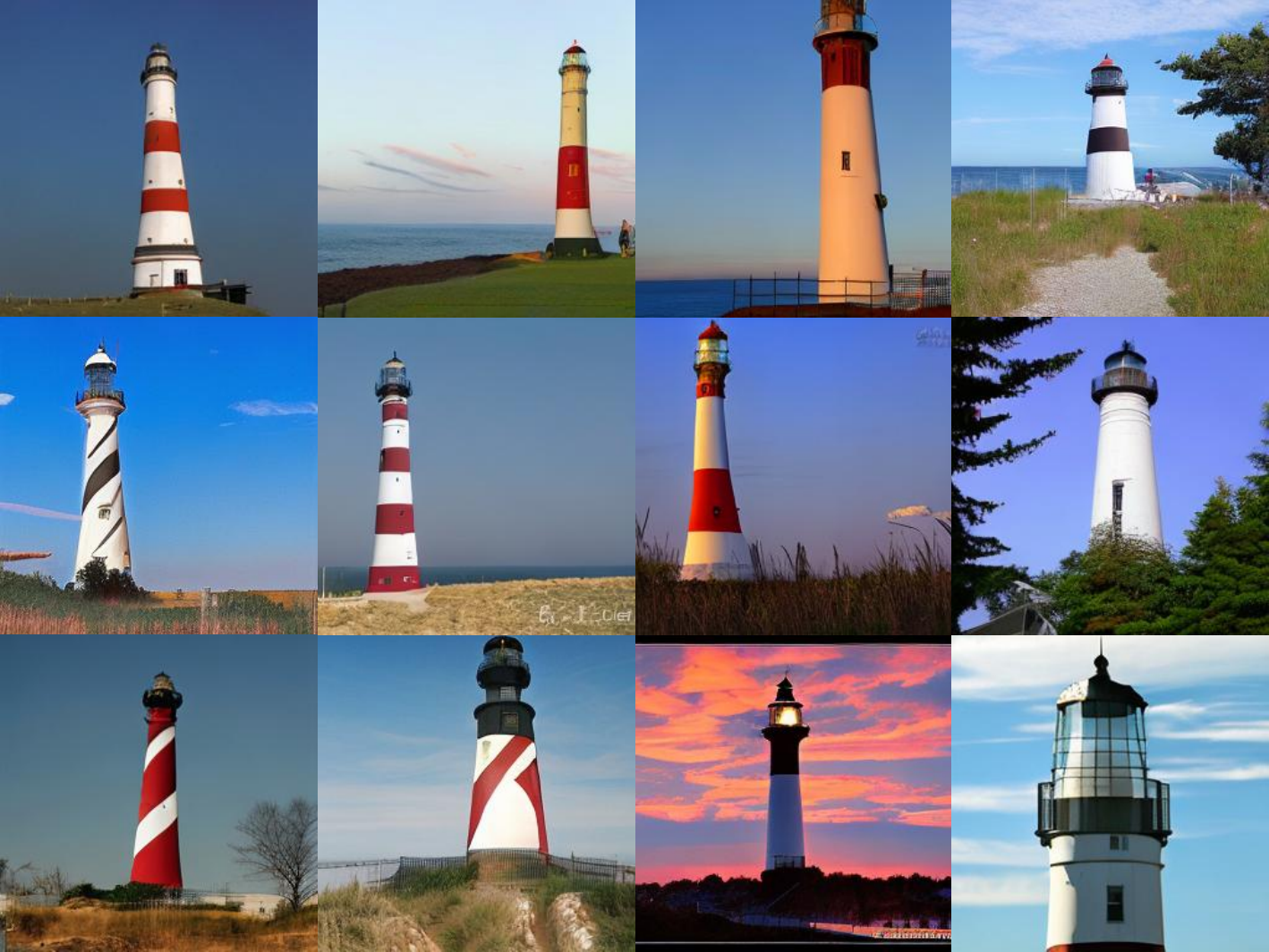}
    \caption{\textbf{Uncurated generation results of our \ourmethod on SIT-XL on ImageNet 256$\times$256.} The class label is ``Beacon, Lighthouse'' (437).}
    \label{fig:supp_more_visualization_11}
\end{figure*}

\begin{figure*}[ht!]
    \centering
    \includegraphics[width=0.82\linewidth,page=12]{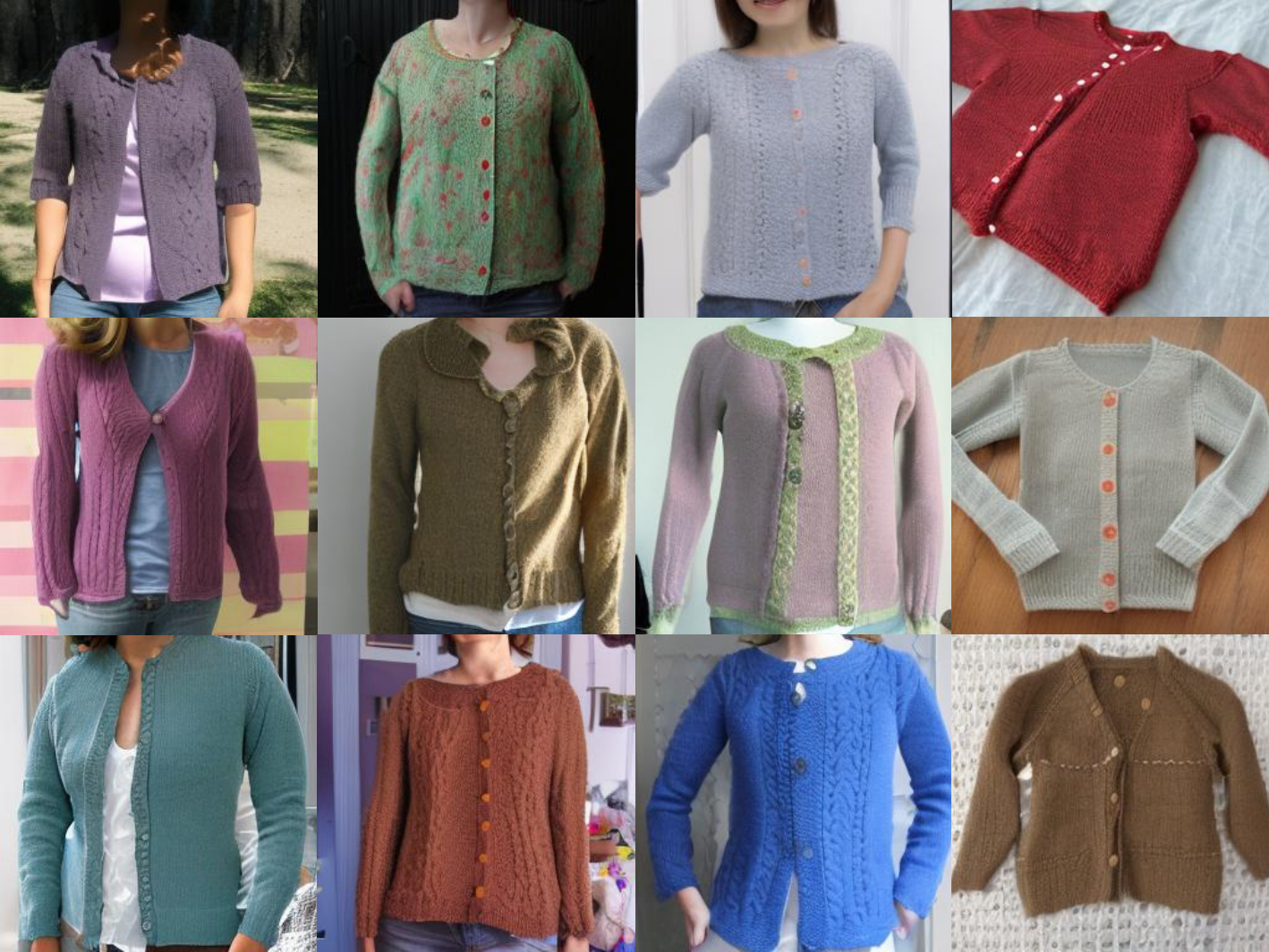}
    \caption{\textbf{Uncurated generation results of our \ourmethod on SIT-XL on ImageNet 256$\times$256.} The class label is ``Cardigan'' (474).}
    \label{fig:supp_more_visualization_12}
\end{figure*}

\begin{figure*}[ht!]
    \centering
    \includegraphics[width=0.82\linewidth,page=13]{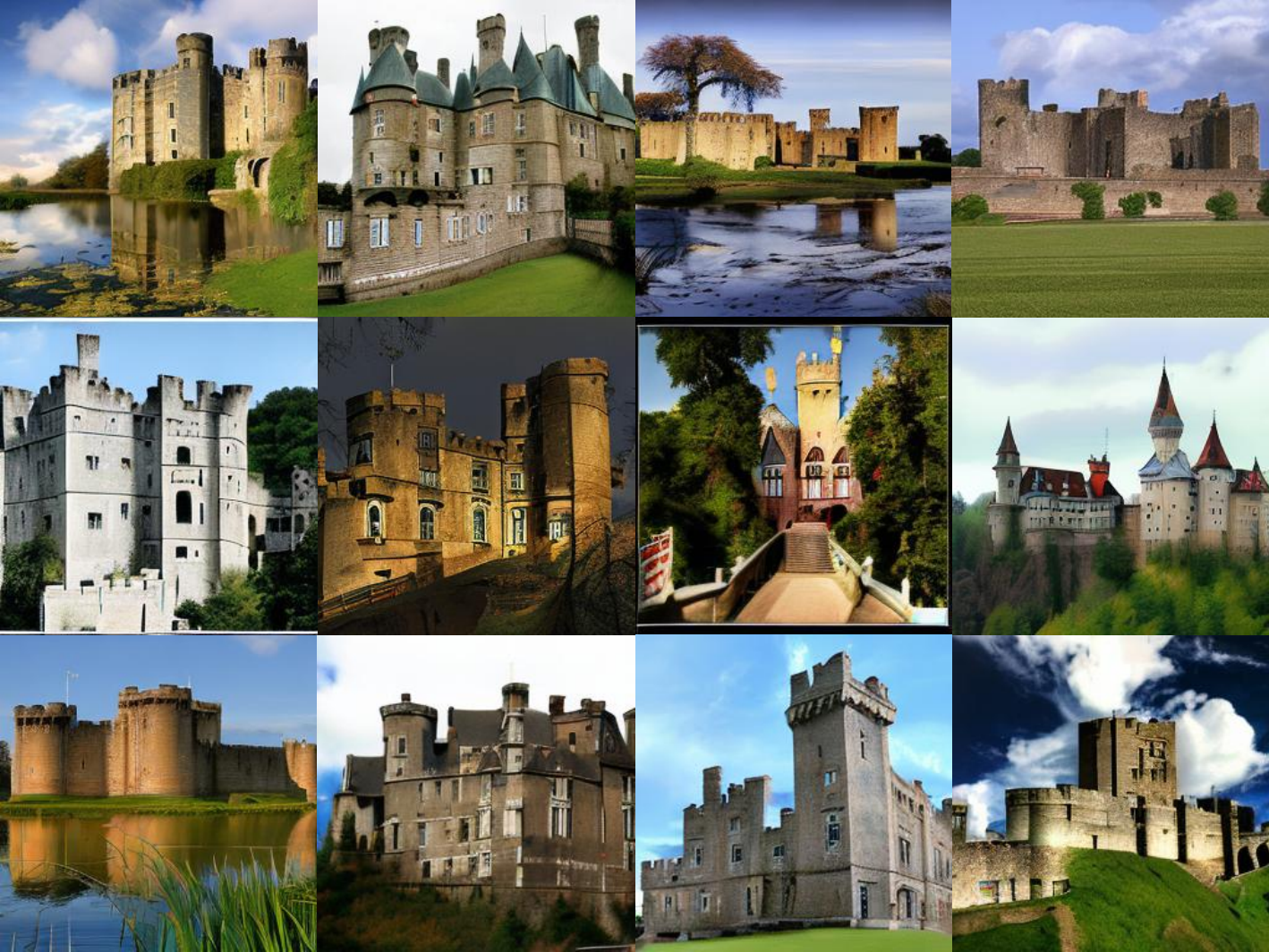}
    \caption{\textbf{Uncurated generation results of our \ourmethod on SIT-XL on ImageNet 256$\times$256.} The class label is ``Castle'' (483).}
    \label{fig:supp_more_visualization_13}
\end{figure*}

\begin{figure*}[ht!]
    \centering
    \includegraphics[width=0.82\linewidth,page=14]{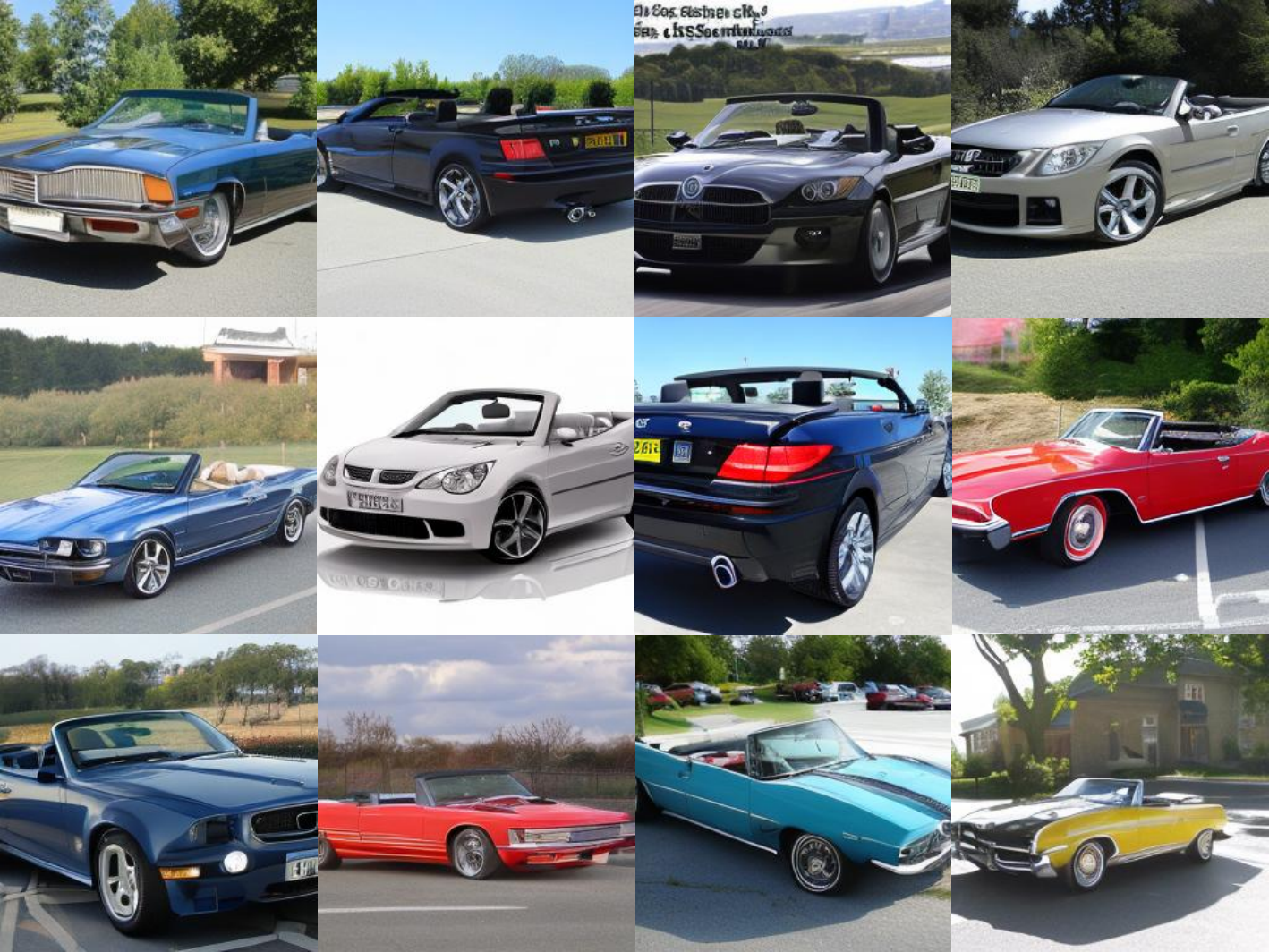}
    \caption{\textbf{Uncurated generation results of our \ourmethod on SIT-XL on ImageNet 256$\times$256.} The class label is ``Check, convertible'' (511).}
    \label{fig:supp_more_visualization_14}
\end{figure*}

\begin{figure*}[ht!]
    \centering
    \includegraphics[width=0.82\linewidth,page=15]{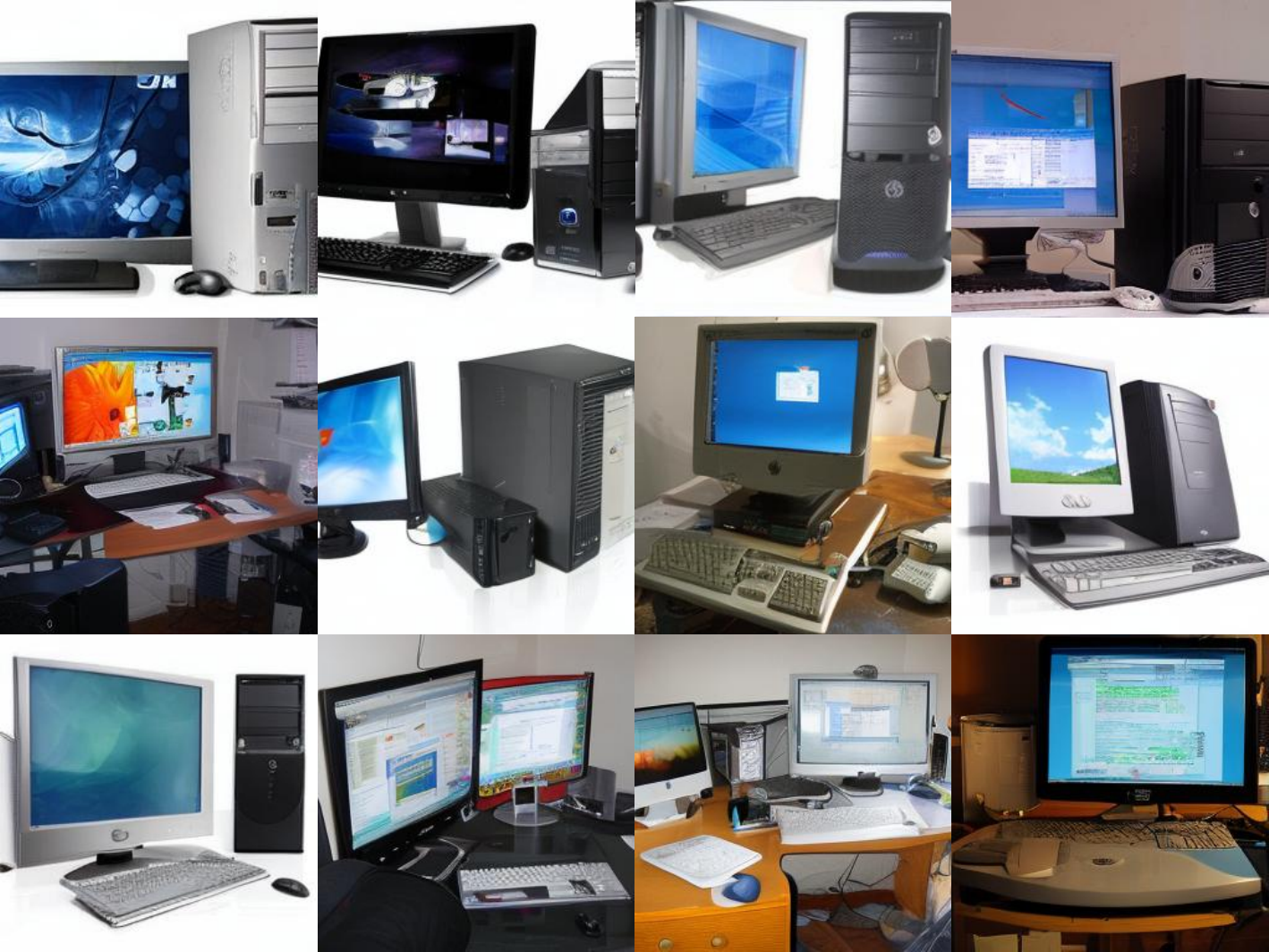}
    \caption{\textbf{Uncurated generation results of our \ourmethod on SIT-XL on ImageNet 256$\times$256.} The class label is ``Desktop computer'' (527).}
    \label{fig:supp_more_visualization_15}
\end{figure*}

\begin{figure*}[ht!]
    \centering
    \includegraphics[width=0.82\linewidth,page=16]{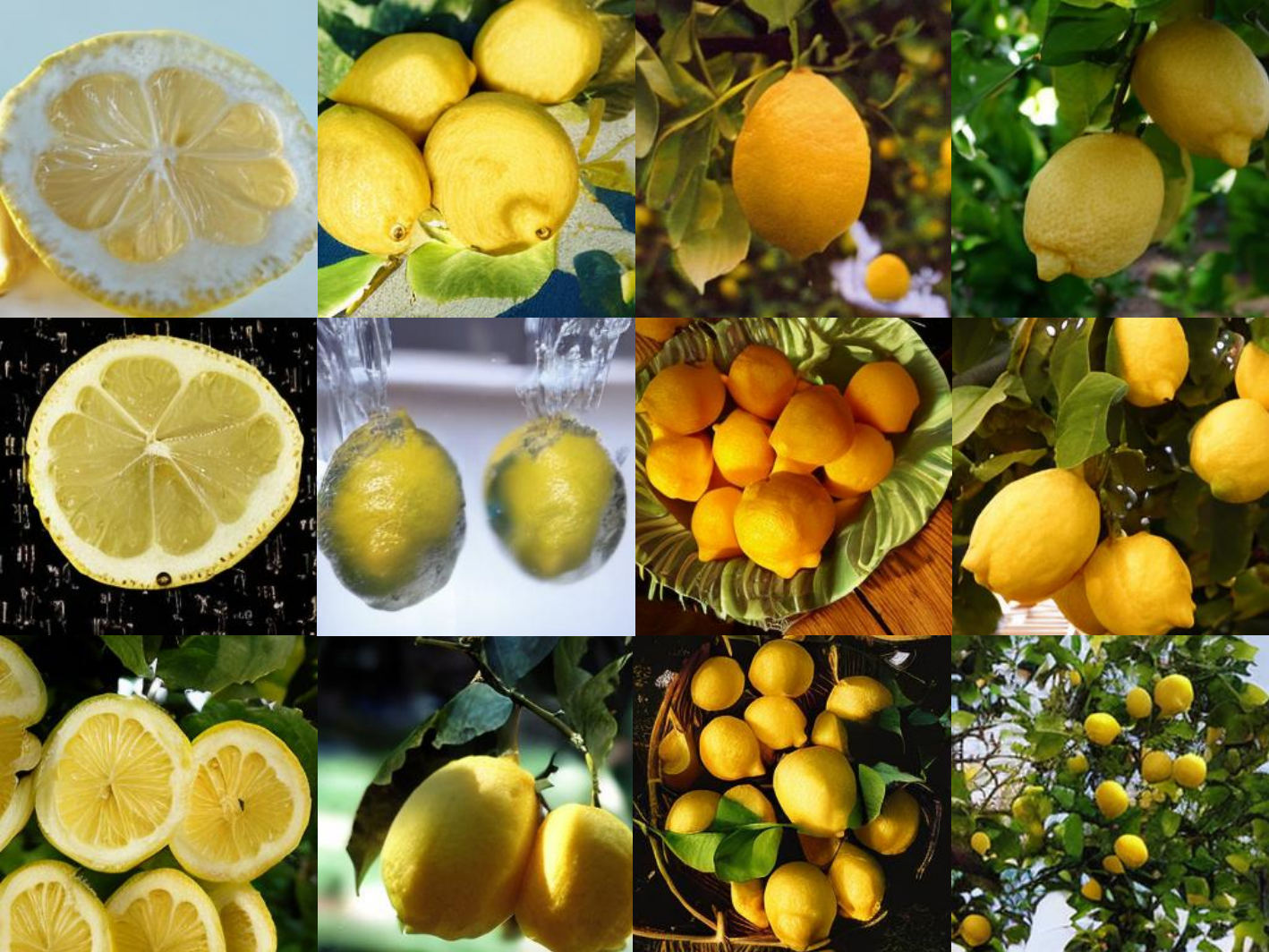}
    \caption{\textbf{Uncurated generation results of our \ourmethod on SIT-XL on ImageNet 256$\times$256.} The class label is ``Lemon'' (951).}
    \label{fig:supp_more_visualization_16}
\end{figure*}

\begin{figure*}[ht!]
    \centering
    \includegraphics[width=0.82\linewidth,page=17]{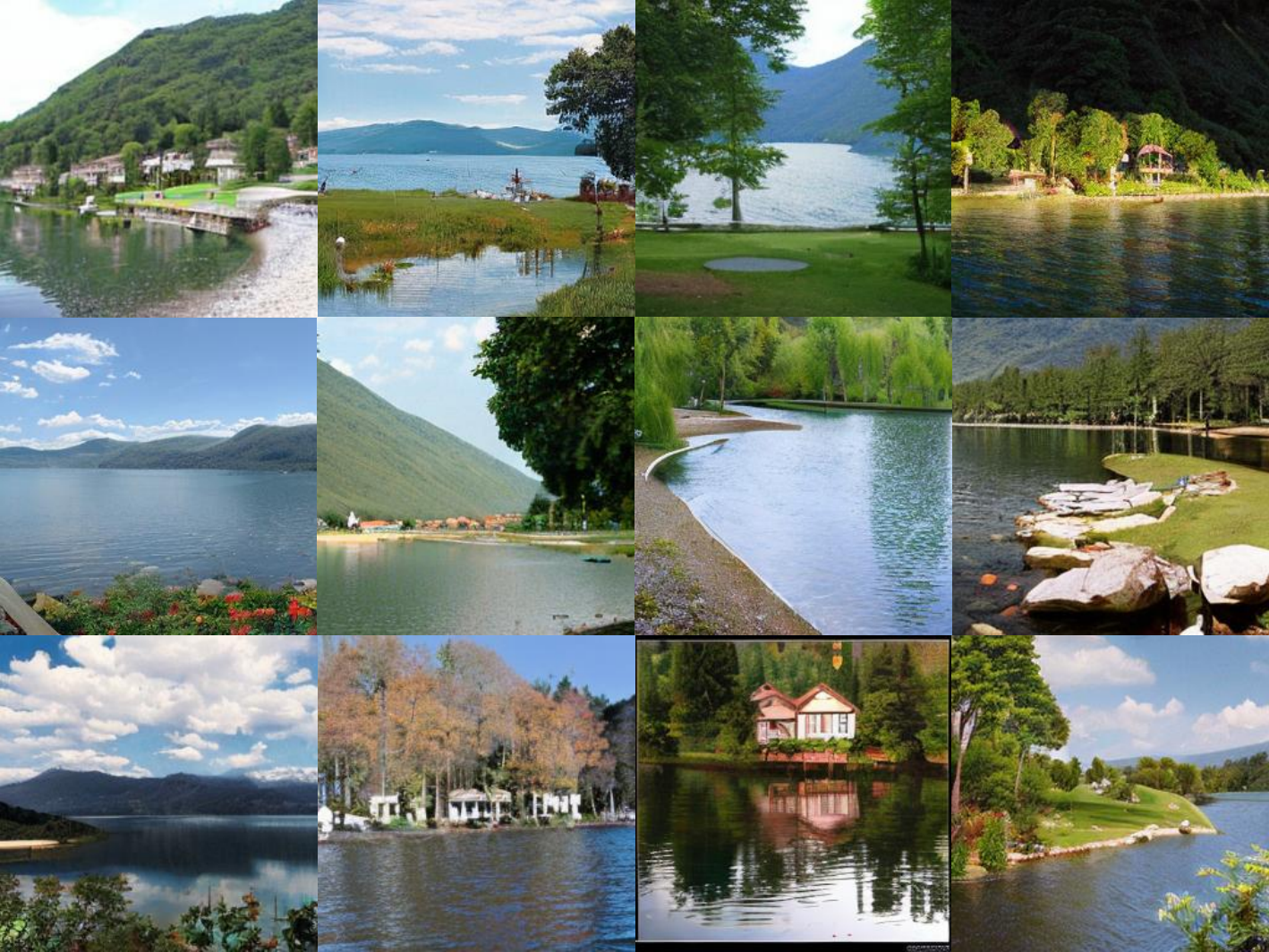}
    \caption{\textbf{Uncurated generation results of our \ourmethod on SIT-XL on ImageNet 256$\times$256.} The class label is ``Lakeside'' (975).}
    \label{fig:supp_more_visualization_17}
\end{figure*}

\begin{figure*}[ht!]
    \centering
    \includegraphics[width=0.82\linewidth,page=18]{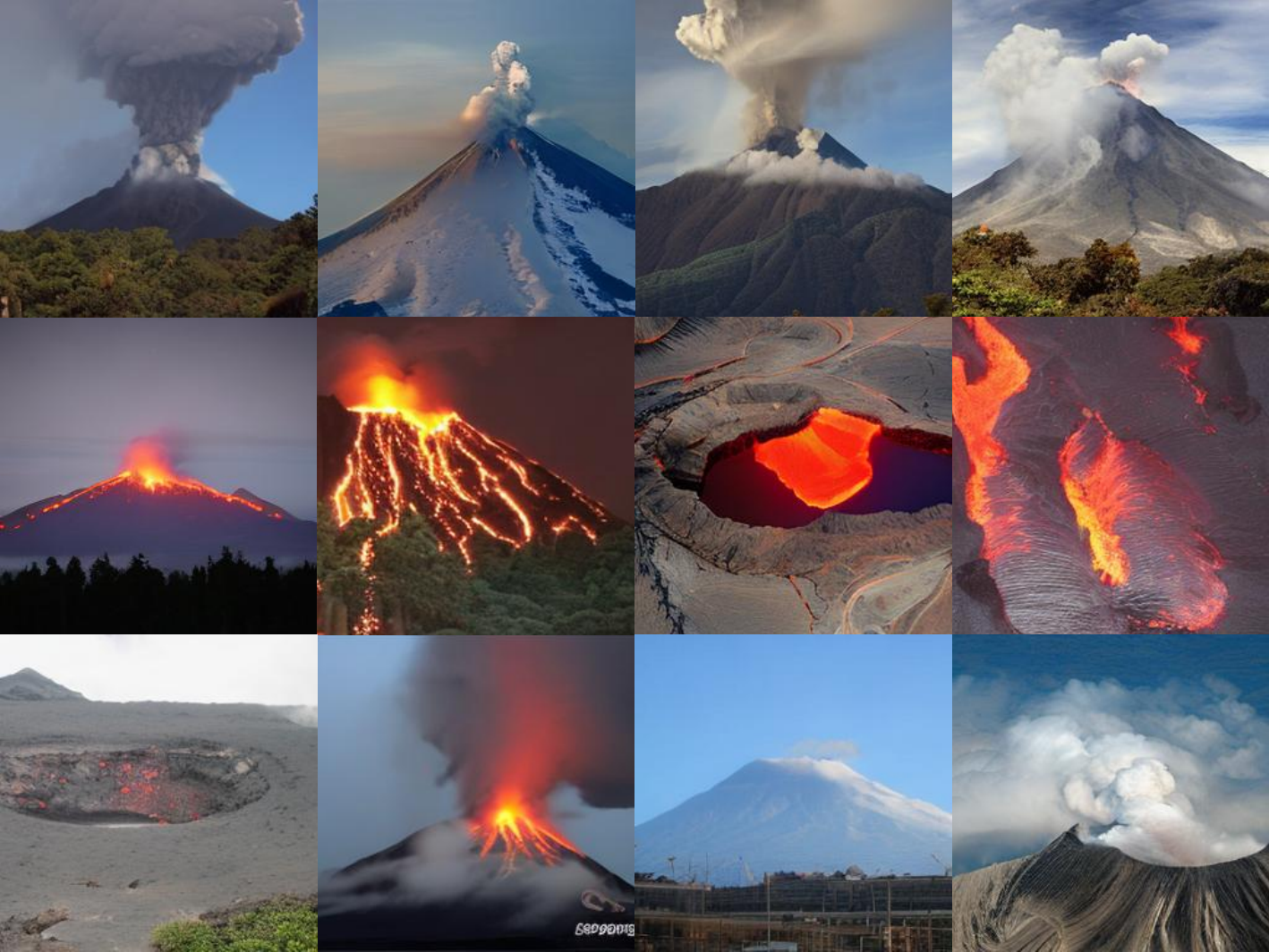}
    \caption{\textbf{Uncurated generation results of our \ourmethod on SIT-XL on ImageNet 256$\times$256.} The class label is ``Volcano' (980).}
    \label{fig:supp_more_visualization_18}
\end{figure*}

\begin{figure*}[ht!]
    \centering
    \includegraphics[width=0.82\linewidth,page=19]{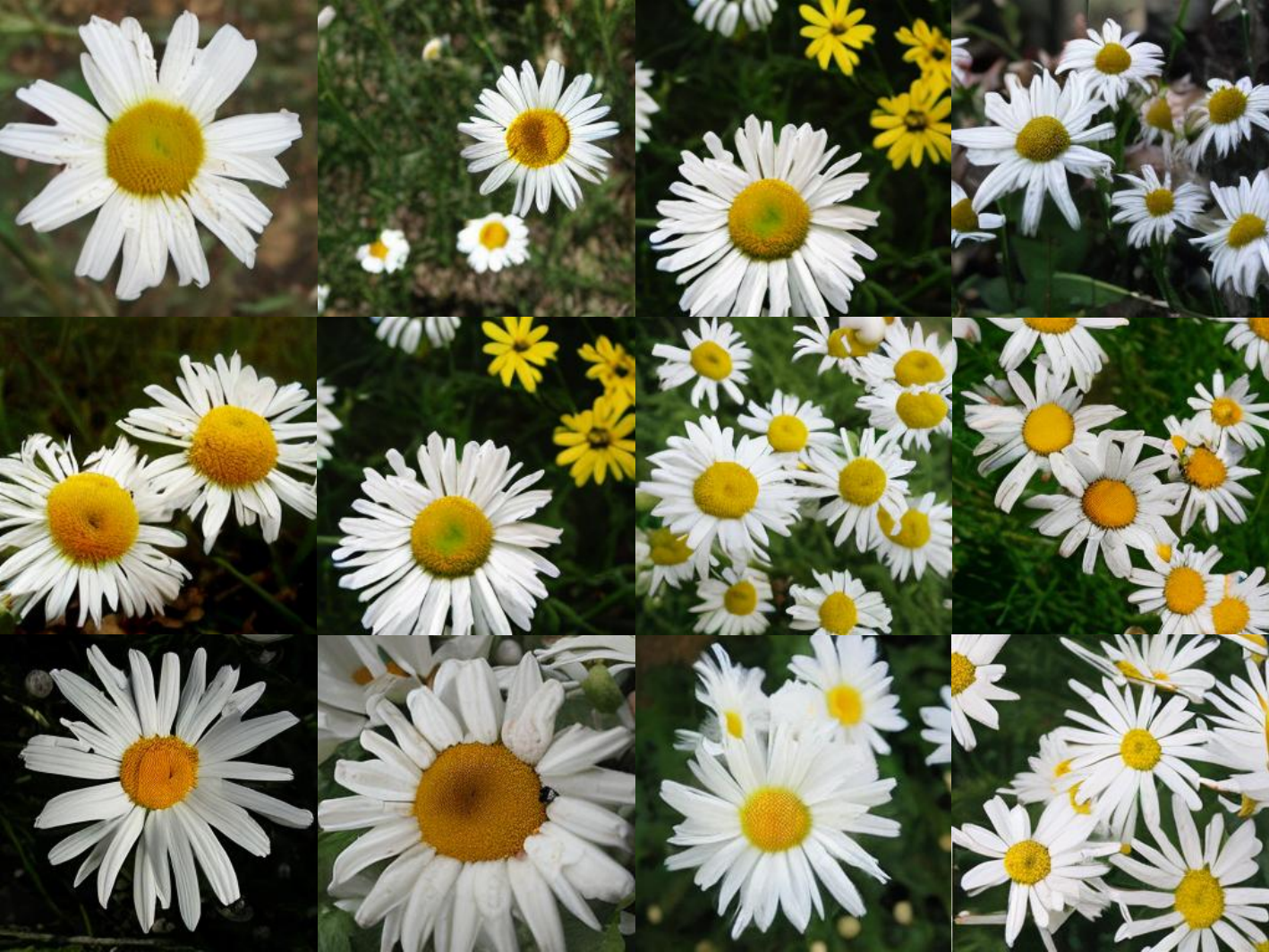}
    \caption{\textbf{Uncurated generation results of our \ourmethod on SIT-XL on ImageNet 256$\times$256.} The class label is ``Daisy'' (985).}
    \label{fig:supp_more_visualization_19}
\end{figure*}

\begin{figure*}[ht!]
    \centering
    \includegraphics[width=0.82\linewidth,page=20]{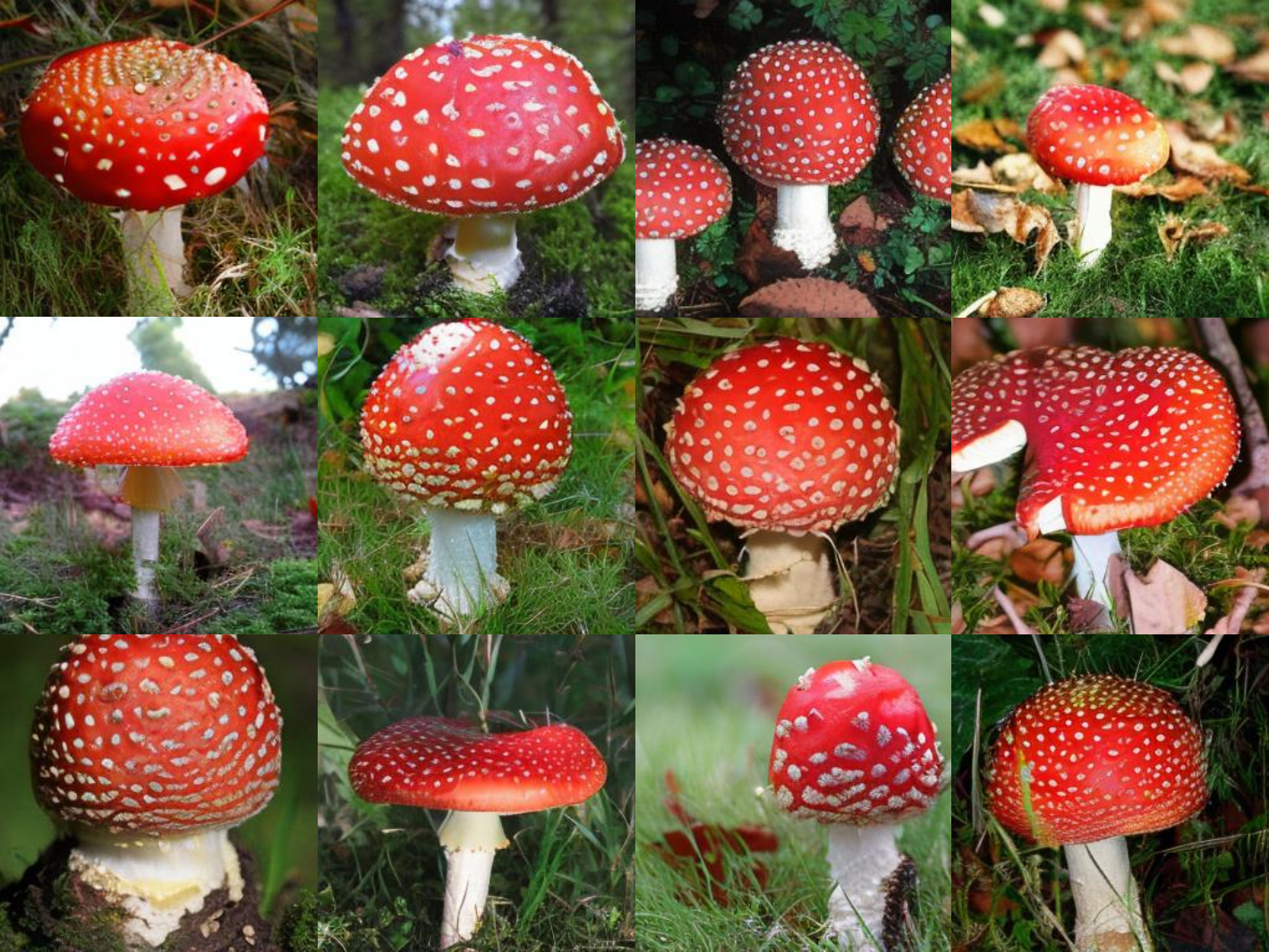}
    \caption{\textbf{Uncurated generation results of our \ourmethod on SIT-XL on ImageNet 256$\times$256.} The class label is ``Agaric'' (992).}
    \label{fig:supp_more_visualization_20}
\end{figure*}

\cref{fig:supp_more_visualization_protein} shows selected samples from protein inverse folding. \cref{fig:supp_more_visualization_molecule_geomdrug} and \cref{fig:supp_more_visualization_molecule_qm9} present molecular generation results on the GEOM-Drug and QM9 datasets.
Additionally, we present more uncurated generation results of our DiverseDiT-XL on ImageNet 256$\times$256 in \cref{fig:supp_more_visualization_1} - \cref{fig:supp_more_visualization_20} with CFG (w = 4.0).
These visualizations complement the quantitative results in the main paper and confirm that the representation diversity promoted by \ourmethod.



\end{document}